\documentclass[pdflatex,sn-apa,iicol]{sn-jnl}

\usepackage{graphicx}%
\usepackage{multirow}%
\usepackage{amsmath,amssymb,amsfonts}%
\usepackage{amsthm}%
\usepackage{mathrsfs}%
\usepackage[title]{appendix}%
\usepackage{xcolor}%
\usepackage{textcomp}%
\usepackage{manyfoot}%
\usepackage{booktabs}%
\usepackage{algorithm}%
\usepackage{algorithmicx}%
\usepackage{algpseudocode}%
\usepackage{listings}%
\usepackage{url}%

\usepackage{makecell}%

\usepackage[utf8]{inputenc} 

\theoremstyle{thmstyleone}%
\theoremstyle{thmstyletwo}%

\theoremstyle{thmstylethree}%

\begin{document}

\title[Article Title]{3DGen-Bench: Comprehensive Benchmark Suite for 3D Generative Models}

\author{
    \textbf{
    Yuhan Zhang\textsuperscript{$1,3*$} \quad
    Mengchen Zhang\textsuperscript{$2,3*$} \quad    
    Tong Wu\textsuperscript{$4\dag$} \quad
    Tengfei Wang\textsuperscript{$3$} \\
    Gordon Wetzstein\textsuperscript{$4$} \quad
    Dahua Lin\textsuperscript{$5$} \quad
    Ziwei Liu\textsuperscript{$6\dag$}
    }
}
        
\affil{
    \textsuperscript{$1$}Fudan University \quad
    \textsuperscript{$2$}Zhejiang University \quad
    \textsuperscript{$3$}Shanghai Artificial Intelligence Laboratory \\
    \textsuperscript{$4$}Stanford University \quad
    \textsuperscript{$5$}The Chinese University of Hong Kong \\
    \textsuperscript{$6$}S-Lab, Nanyang Technological University
}

\affil{
\textsuperscript{$*$}These authors contributed equally to this work. \quad \textsuperscript{$\dag$}Corresponding author}



\abstract{
  3D generation is experiencing rapid advancements, while the development of 3D evaluation has not kept pace. How to keep automatic evaluation equitably aligned with human perception has become a well-recognized challenge. Recent advances in the field of language and image generation have explored human preferences and showcased respectable fitting ability. However, the 3D domain still lacks such a comprehensive preference dataset over generative models. To mitigate this absence, we develop \emph{3DGen-Arena}, an integrated platform in a battle manner. Then, we carefully design diverse text and image prompts and leverage the arena platform to gather human preferences from both public users and expert annotators, resulting in a large-scale multi-dimension human preference dataset \emph{3DGen-Bench}. Using this dataset, we further train a CLIP-based scoring model, \textit{3DGen-Score}, and a MLLM-based automatic evaluator, \textit{3DGen-Eval}. These two models innovatively unify the quality evaluation of text-to-3D and image-to-3D generation, and jointly form our automated evaluation system with their respective strengths. Extensive experiments demonstrate the efficacy of our scoring model in predicting human preferences, exhibiting a superior correlation with human ranks compared to existing metrics. We believe that our \textit{3DGen-Bench} dataset and automated evaluation system will foster a more equitable evaluation in the field of 3D generation, further promoting the development of 3D generative models and their downstream applications.}




\maketitle

\begin{figure*}[ht]
  \centering
   \includegraphics[width=\linewidth]{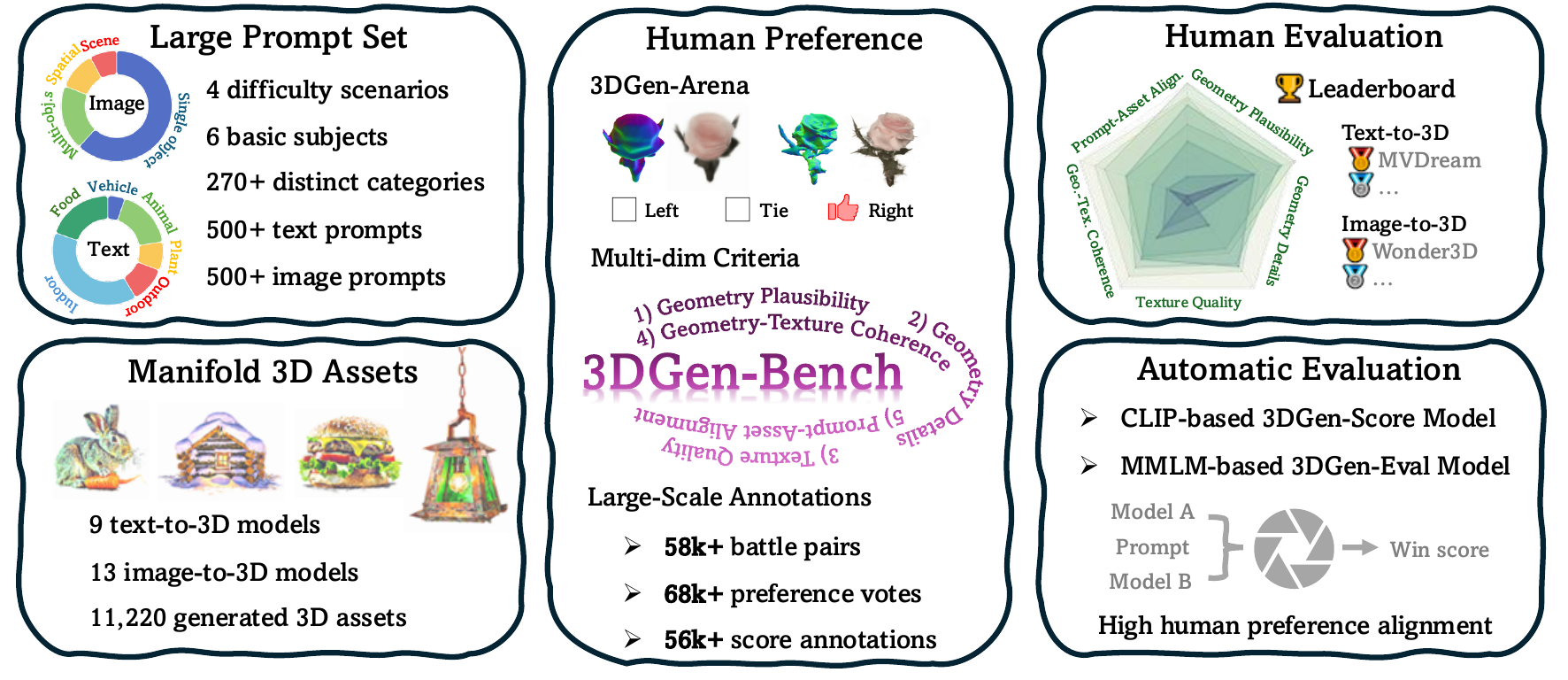}
    \vspace{-10pt}
   \caption{\textbf{Overview of 3DGen-Bench,} which is the first comprehensive human preference dataset for 3D models. For efficient data collection, we build the \textbf{3DGen-Arena} platform in a pairwise battle manner. Based on the annotated data, we perform a comprehensive evaluation for state-of-the-art 3D generative models, and propose two scoring model, \textbf{3DGen-Score} and \textbf{3DGen-Eval} models, regarded as automated 3D evaluators that align well with human judgments.
   }
   \label{fig:teaser}
\end{figure*}


\section{Introduction}\label{sec1: intro}
3D content generation has emerged as a pivotal research area with diverse applications across virtual reality, video games, films, and robotics. In recent years, the field has undergone a transformative revolution fueled by the extraordinary generative capabilities of diffusion models and the availability of large-scale 3D datasets. This has facilitated an explosion of 3D generative models~\citep{Dreamfusion, Score_Jacobian_Chaining, Fantasia3D, Latent-NeRF, Magic3D,tang2023_makeit3d, Magic123, Mvdream, GeoDream, Sherpa3D, chen2024comboverse,SweetDreamer, HIFA, ProlificDreamer, raj2023dreambooth3d,DreamCraft3D,wang2024themestation, GSGEN, Hyper-3DG, dreamgaussian, Ponit-e,Shap•E, wu2022voxurf, gupta20233dgen, cheng2023sdfusion, hong20243dtopia, Zero-1-to-3, gao2024cat3d, EscherNet, One-2-3-45++, One-2-3-45, SyncDreamer, Zero123++, Free3D, SV3D, Wonder3D, IM-3D, Triplane, LGM, GRM} that can craft high-fidelity 3D assets from specific textual or visual prompts.

However, the advancement in 3D evaluation techniques has not kept pace with the increase of 3D generative models. 
While CLIP similarity~\citep{Clip, CLIPScore} has been widely adopted for measuring text-image alignment, its application in 3D often proves inadequate and inaccurate. This is primarily due to the lack of 3D-specific prior knowledge and its inability to accommodate the versatile evaluation requirements of 3D assets.User-Study provides more reliable assessments, but the absence of a standardized prompt set makes fair comparisons across different works challenging.
Recognizing this, $T^3$Bench~\citep{T3_Bench} attempted to establish a more suitable benchmark by manually curating a prompt set designed for text-to-3D generation. However, its metrics for the \textit{Quality Assessment} dimension still rely on the original CLIP~\citep{Clip}, leaving its inherent limitations unaddressed. Meanwhile, using BLIP~\citep{BLIP} as the caption model introduces unnecessary errors in the \textit{Alignment Assessment} dimension.
GPTEval3D~\citep{Gpt4v_evaluation} explored the potential of prompting GPT-4V(ision), a powerful multimodal large language model, as a versatile and human-aligned 3D evaluator. While it demonstrated notable robustness, the closed-source and black-box nature of GPT-4V raises concerns about potential biases, which remain challenging to interpret.

Existing metrics for 3D assessment lack grounding in real-world human preference data, a principle well-established in 2D image generation benchmarks. Numerous works in image evaluation~\citep{Pick-a-Pic, ImageReward, ImagenHub, HPSv2, Laion-5B} have developed metrics that closely align with human preferences. In contrast, a similar, transparent, user-focused metric for 3D assets is still absent. This highlights the pressing need for a large-scale dataset that captures human preferences for 3D generation, a crucial step toward aligning 3D assessments with human judgment.

In this paper, we propose the first comprehensive benchmark that incorporates large-scale human preferences and uniforms the evaluation of text-to-3D and image-to-3D generation, named \textbf{\emph{3DGen-Bench}}. Our main contributions can be summarized in three aspects:
\textbf{1) Comprehensive model zoo for 3D generation.} 
To build this dataset, we carefully curate over \emph{1,000} text and image prompts across various categories, covering single-object, multi-object, and micro-scene, which is approximately tenfold the number of prompts designed in previous works~\cite{T3_Bench, Gpt4v_evaluation}. With the proposed prompt set, we evaluate 19 representative open-sourced generative models, including 9 for text-to-3D and 13 for image-to-3D, yielding over \emph{10,000} 3D assets and more than \emph{58,000} battle pairs. 
\textbf{2) Large-scale multi-criteria human preference collection.} 
To gather extensive human feedback, we build a voting platform open to community, dubbed as \emph{3DGen-Arena}, which adopt the anonymous pairwise battle manner, enabling users to express their preferences between pairs of 3D generative models.
Considering the complexity of 3D contents, we identified five key dimensions for assessment~\cite{Gpt4v_evaluation}: \emph{Geometry Plausibility}, \emph{Geometry Details}, \emph{Texture Quality}, \emph{Geometry-Texture Coherency}, and \emph{Prompt-Asset Alignment}.
In practice, we collect feedback from two user groups: the general public and specialized experts. Although votes from public users are valuable for their volume, they may introduce noises that can affect subsequent tasks. To mitigate this, we also engage experts whose annotations offer a more reliable data source for human preferences.
In addition to the pairwise comparisons from \emph{3DGen-Arena}, we also ask our experts to assign absolute scores for each 3D model, which is more suitable for large-scale evaluation.
\textbf{3) Human-aligned automatic scoring models.}
The collected human preference dataset enables the training of a scoring model. Inspired by \cite{InstructGPT, Pick-a-Pic}, we finetune a pre-trained CLIP model~\cite{Clip} on our proposed dataset in a Reward Modeling(RM) style, named \textbf{{\emph{3DGen-Score}}}, which is designed to predict the human preference between two 3D models generated from the same prompt. Specifically, the model takes a prompt along with four-view normal and RGB renderings from two models as input and returns a win-rate tuple for five dimensions. Furthermore, leveraging the success of Multi-modal Large Language Models(MLLMs), we fine-tune LLaVA~\cite{LLaVa} to evaluate 3D models in a Question\&Answer(QA) style, termed as \textbf{\emph{3DGen-Eval}}, which is encouraged to provide explanations to enhance credibility.

We have carried out comparative experiments on test data, evaluating our models against commonly used metrics such as CLIP-Score~\cite{CLIPScore}, Aesthetic score~\cite{Laion-5B}, GPTEval3D~\cite{Gpt4v_evaluation}, and others. Experimental results demonstrate that our proposed models exhibit strong alignment with human references, comprehensively outperforming all other metrics in both pairwise alignment and ranking correlation. Meanwhile, each model offers distinct advantages. 3DGen-Score excels in human alignment, efficiency, and scalability, while 3DGen-Eval offers superior interpretability and robustness, matching or surpassing GPT-4V without additional costs.
Furthermore, we also experimentally validate the potential of our scoring model in optimizing the quality of generated models served as a loss function or reward model. 
We believe that our work will play a significant role in advancing 3D evaluation, paving the way for high-quality 3D generation.

\begin{table*}[ht]
    \centering
    \small
    \caption{\textbf{Comparison between 3DGen-Bench dataset and others.} Our dataset is the first to incorporate both text and image prompts, as well as the first to implement absolute value scoring annotations, showcasing a substantial advantage in model scale. Note that dimension "public" is for annotations only, and symbol "$\aleph$" represents "partly public avaliable" as 3DRewardDB \citep{DreamReward} only release partial annotations of 1000 prompts.}
    \setlength{\tabcolsep}{7.5pt}
    \renewcommand{\arraystretch}{1.0}
    \begin{tabular}{l|cc|cccc|ccc}
    \toprule
        \multirow{2}{*}{Dataset} & \multicolumn{2}{c|}{Condition} & \multirow{2}{*}{Prompt} & \multirow{2}{*}{Dim} & \multirow{2}{*}{Model} & \multirow{2}{*}{Assert} & 
        \multicolumn{3}{c}{Annotation} \\
                                 & text & image & & & & & pair & score & public\\
        \midrule
        $T^3$ Bench & $\checkmark$ & - & 300 & 2 & 6 & - & - & - & - \\
        GPTEval3D & $\checkmark$ & - & 110 & 5 & 13 & 1,430& 234  & - & - \\
        3DRewardDB & $\checkmark$ & - & 2,530 & 4 & 5 & - & 25.3k & - & $\aleph$ \\
        3DGen-Bench(Ours) & $\checkmark$ & $\checkmark$ & 1,020 & 5 & 22 & 11,220 & 13.8k & 11.2k & \checkmark \\
    \bottomrule
    \end{tabular}
    \label{tab:dataset_compare}
\end{table*}

\section{Related Work}\label{sec2:realted_work}

\subsection{3D Generative Models.}
\label{sec2.1:3d_generative_models}
Conditional 3D generation creates 3D models from text or image prompts. Empowered by advances in text-to-image diffusion models \citep{Stable_diffusion, Imagen}, optimization-based methods \citep{Dreamfusion, Score_Jacobian_Chaining, Fantasia3D, Latent-NeRF, Magic3D,tang2023_makeit3d, Magic123, Mvdream, GeoDream, Sherpa3D, chen2024comboverse,SweetDreamer, HIFA} have applied Score Distillation Sampling (SDS) \citep{Dreamfusion} for conditional 3D generation. \citep{ProlificDreamer} further presented Variational Score Distillation (VSD), which optimized the distribution of 3D scenes instead. Some works~\citep{raj2023dreambooth3d, DreamCraft3D, wang2024themestation} employ a second-stage DreamBooth-like model fine-tuning for enhanced texture modeling.  Recent works \citep{GSGEN, Hyper-3DG, dreamgaussian} have incorporated SDS with 3D Gaussian Splatting~\citep{kerbl20233d}. 
In the meantime, feed-forward diffusion models \citep{Ponit-e, Shap•E, wang2022_rodin, gupta20233dgen, cheng2023sdfusion, hong20243dtopia} also show respectable performance with fast inference. Transformer-based regression models~\citep{Lrm, OpenLRM, Instant3d} train a regression model to predict triplanes directly from images. Another line of work~\citep{Zero-1-to-3, gao2024cat3d, EscherNet, One-2-3-45++, One-2-3-45, SyncDreamer, Zero123++, Free3D, SV3D, Wonder3D} fine-tunes image or video diffusion models to generate novel views for stronger 3D guidance. Some works~\citep{ IM-3D, Triplane, LGM, GRM} use 3D Gaussian Splatting~\citep{kerbl20233d} for multi-view reconstruction. 
With rapid advancements in 3D generative technologies, the need for an objective and comprehensive evaluation system has become increasingly urgent. In this paper, we propose a comprehensive evaluation framework that encompasses five distinct dimensions informed by the collection of human preference data. Building on this foundation, we further train automated evaluators, achieving exceptional performance in simulating human judgments.

\subsection{3D Evaluation Metrics.}
\label{sec2.2:3d_evaluation_metrics}
Evaluating conditional 3D generative models is challenging due to the need to understand both 3D worlds and given conditions. 
For text-conditioned generative tasks, CLIP~\citep{Clip} has been widely used to assess text-to-3D alignment.
For image-conditioned generative tasks, multi-view datasets such as GSO~\citep{GSO} and RTMV~\citep{RTMV} are usually used to evaluate single-view reconstruction quality.
However, these metrics are constrained by the challenge of acquiring a comprehensive reference set. 
User studies are adopted by many works~\citep{ProlificDreamer, Mvdream, SV3D, GRM, Magic3D} but are costly and difficult to scale. Recent efforts have focused on automated text-to-3D evaluation. $T^3$Bench~\citep{T3_Bench} introduced two automatic metrics based on multi-view images, but they rely on a hand-designed regional convolution mechanism, raising concerns about their credibility and validity. Wu et al.~\citep{Gpt4v_evaluation} proposed leveraging GPT-4V's visual question-answering capability to evaluate 3D models, but this approach may introduce uninterpretable biases in answer selection.
Our framework stands out by incorporating human preference data into a comprehensive five-dimensional evaluation scheme, effectively capturing the subtleties of generated 3D content missed by prior metrics. Leveraging these insights to train automated evaluators, we enhance alignment with human preferences. This hybrid approach bridges the gap between subjective evaluations and automated metrics, providing a more reliable and scalable solution for 3D model assessment.

\subsection{Human Preference.}
\label{sec2.3:human_preference}
Recent advances in language and image generation have placed significant emphasis on human preference. Learning from human preference not only enables adaptation to complex tasks that may challenge automatic evaluation but also guides models to align with human expectations. Specifically, there are two major downstream scenarios: human evaluation for model ranking~\citep{Chatbot-Arena, Pick-a-Pic, tts-arena, ImagenHub, WildWision-Arena} and human feedback for model tuning~\citep{Webgpt, Openai_RLHF, HH-RLHF, OASST, SHP, alpaca, Self_Instruct}. LMSYS~\citep{Chatbot-Arena} introduced a new preference collection method to mitigate individual bias, making it accessible to the public through an open arena platform.
In this paper, we extend human preference learning to the field of 3D evaluation by establishing the first 3D arena. Additionally, we integrate the two key application aspects: on one hand, we establish a 3D leaderboard driven by human evaluations; on the other hand, we develop a reward model for automatic evaluation and generative model optimization.

\begin{figure}[t]
    \centering
    \includegraphics[width=0.35\textwidth]{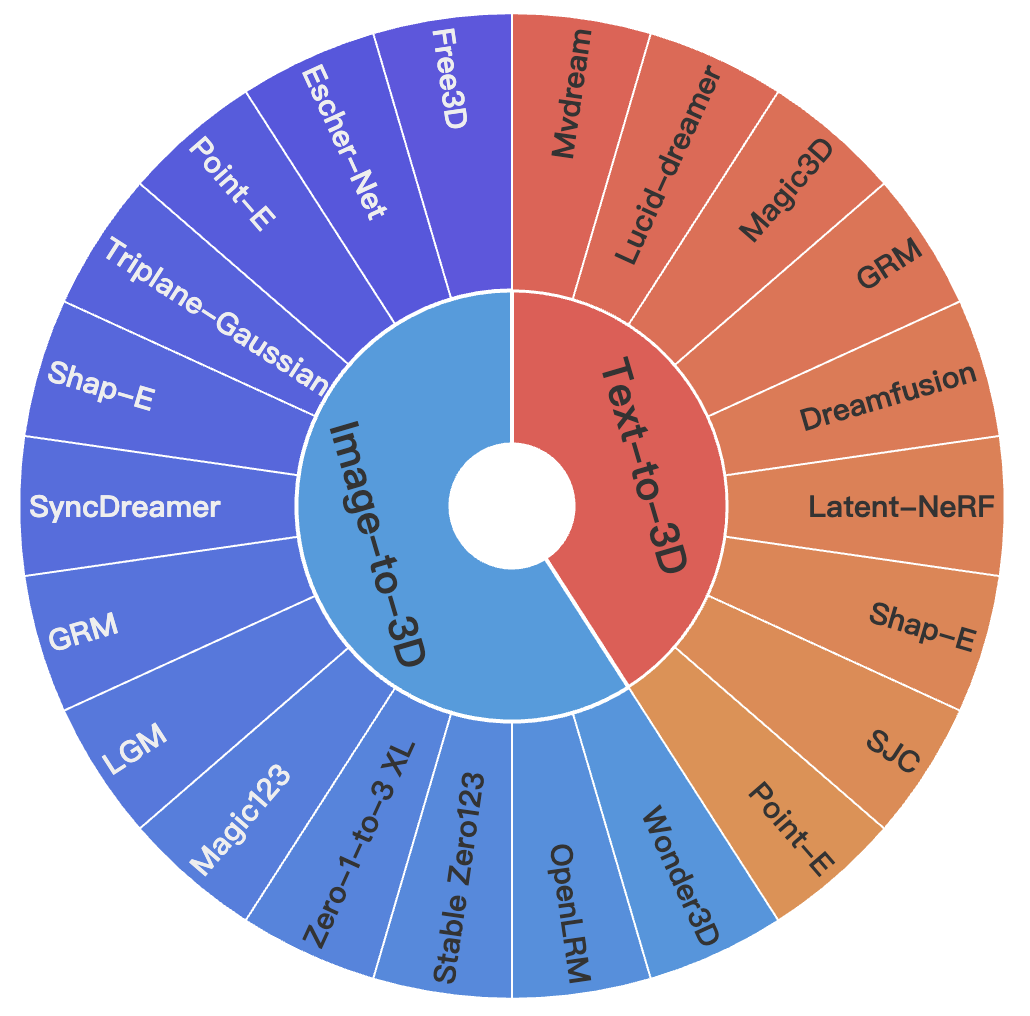}
    \caption{\textbf{3D Generative Models.} The specific list of generative models we selected for our benchmark. }
    \label{fig:models}
\end{figure}

\begin{figure}[t]
    \centering
    \includegraphics[width=\linewidth]{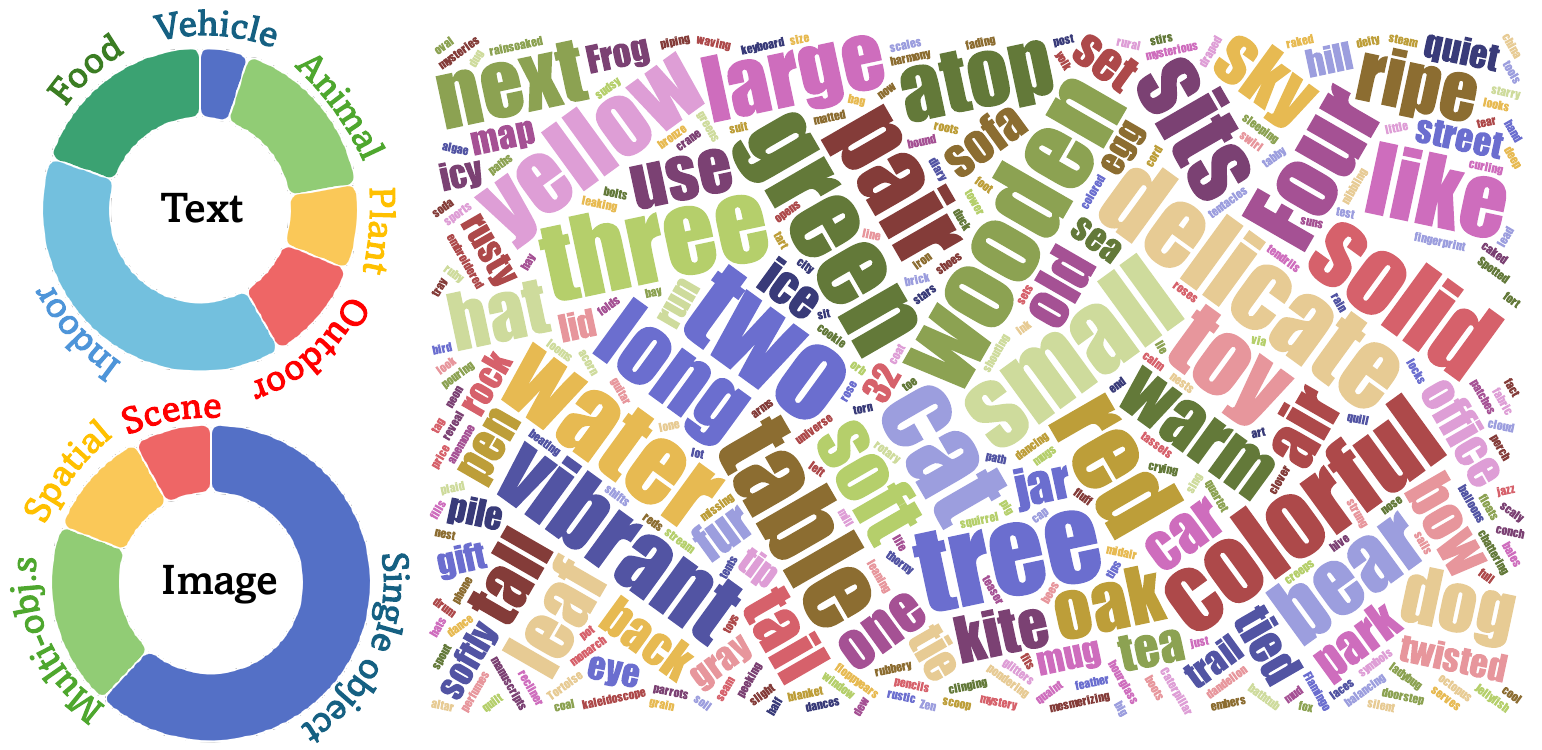}
    \vspace{-10px}
    \caption{\textbf{Visualization of prompts}. \textbf{Left} are two fan diagrams respectively show the distribution of the 6 basic category subjects and 4 main difficulty scenarios (where the "implicit count" and "explicit count" are both grouped into the "Multi-obj.s"). \textbf{Right} is the visualization of word cloud of text prompts.}
    \label{fig:prompt_analysis}
\end{figure}

\section{3DGen-Bench Dataset}
\label{sec3:Dataset}

\subsection{Comprehensive Model Zoo for 3D Generation}
\label{sec3.1:model_zoo}

\subsubsection{3D Generative Models Involved} 
\label{sec3.1.1:3d_generative_models} 
\noindent{\textbf{Model Selection.}} To ensure comprehensive benchmark coverage, we collect a variety of 3D generative models, including both optimization-based and feed-forward approaches, and encompassing various representations such as point clouds, triplanes, NeRF, and 3D Gaussian models. Additionally, we include pioneering works that leverage multi-view diffusion models for novel sparse reconstruction. 
Ultimately, our benchmark includes 9 text-to-3D generative models~\citep{Mvdream, Luciddreamer, Magic3D, GRM, Dreamfusion, Latent-NeRF, Shap•E, Score_Jacobian_Chaining, Ponit-e} and 13 image-to-3D generative models~\citep{Wonder3D, OpenLRM, Zero-1-to-3, Magic123, LGM, GRM, SyncDreamer, Shap•E, Triplane, Ponit-e, EscherNet, Free3D}, as shown in Figure~\ref{fig:models}. The complete list of the included models is provided in Appendix~\ref{secA2.2:appendix_3dmodel}.
This diverse selection ensures that our benchmark encompasses a broad spectrum of 3D generative models, catering to varied research requirements and offering a comprehensive evaluation platform.

\noindent{\textbf{Model Implementation.}} We prioritize using the official code of each model when available; otherwise, we rely on Threestudio's~\citep{threestudio2023} implementation for consistency. For LRM~\citep{Lrm}, we employ the open-sourced implementation OpenLRM~\citep{OpenLRM}. All experiments were conducted using the default hyperparameters specified in each code.
The entire generation process took place over 4 weeks, utilizing 8 NVIDIA A100-SXM4-80GB GPUs.

\noindent{\textbf{Render Implementation.}}For each 3D model, we attempt to obtain surrounding videos in three formats: RGB, normal, and geomerty(texture-free geometry). As the 3D models are generated in PLY, OBJ or GLB formats, for models that can directly generate RGB and normal renderings, we capture these directly. Otherwise, we uniformly convert it into mesh representation, and then render it into required videos. 

\subsubsection{Prompt Generation} 
\label{sec3.1.2:prompt_generation}
We carefully design 1,020 prompts, including 510 texts and 510 images, which is approximately $10\times$ the previous prompt suites~\citep{T3_Bench, Gpt4v_evaluation}. Our prompt set spans more than 270 distinct categories, derived from 6 basic subjects: \emph{``Vehicle'', ``Animal'', ``Plant'', ``Food'', ``Indoor objects'', ``Outdoor objects''}. Furthermore, to evaluate the models' robustness for complex prompts, we expanded the prompt set to include scenarios with multiple objects and micro-scenes. Ensuring diversity and a balanced distribution of prompts across both text and image generation is crucial. During the prompt generation process, we use predefined proportions to sample categories, attributes, and scenario complexities. The raw prompts are then manually screened to ensure diversity and representativeness. The visualization of prompts can be found in Figure~\ref{fig:prompt_analysis}. Subsequently, we conducted stratified sampling on the cleaned prompts, splitting them into training, validation, and test sets in a $9:1:2$ ratio. 
More details on this process can be found in Appendix~\ref{secA2.1:appendix_prompt}.

\noindent{\textbf{Text Prompt.}}
We employ GPT-4 to generate text prompts heuristically. This process involves deconstructing sentences into four fundamental components: count, category, attribution, and spatial relation. For example, a sentence can be structured to include an explicit count ("three"), a category ("apples"), an attribution ("green"), and a spatial relation ("on the table"). Based on this, we provide GPT-4 with various sentence templates and manually filter the final prompts. More details can be found in Appendix~\ref{secA2.1:appendix_prompt}.
Ultimately, our text set includes exactly 510 prompts, comprising 43 with explicit counts, 104 with implicit counts (e.g., ``a pile of''), 73 featuring spatial relationships, and 290 individual objects.

\noindent{\textbf{Image Prompt.}}
We manually collect and curate a set of 510 images, ensuring their quality and diversity through a careful selection process. To prepare them as prompts, we use \texttt{rembg}~\citep{rembg} to remove backgrounds. As a result, the final image set includes 47 multi-object compositions featuring more than six objects, 49 compositions with a modest number of 2-6 objects, 61 instances of spatial relational compositions, 40 complex scenes, and 313 individual objects.

\begin{figure*}[t]
    \centering
    \includegraphics[width=0.49\textwidth]{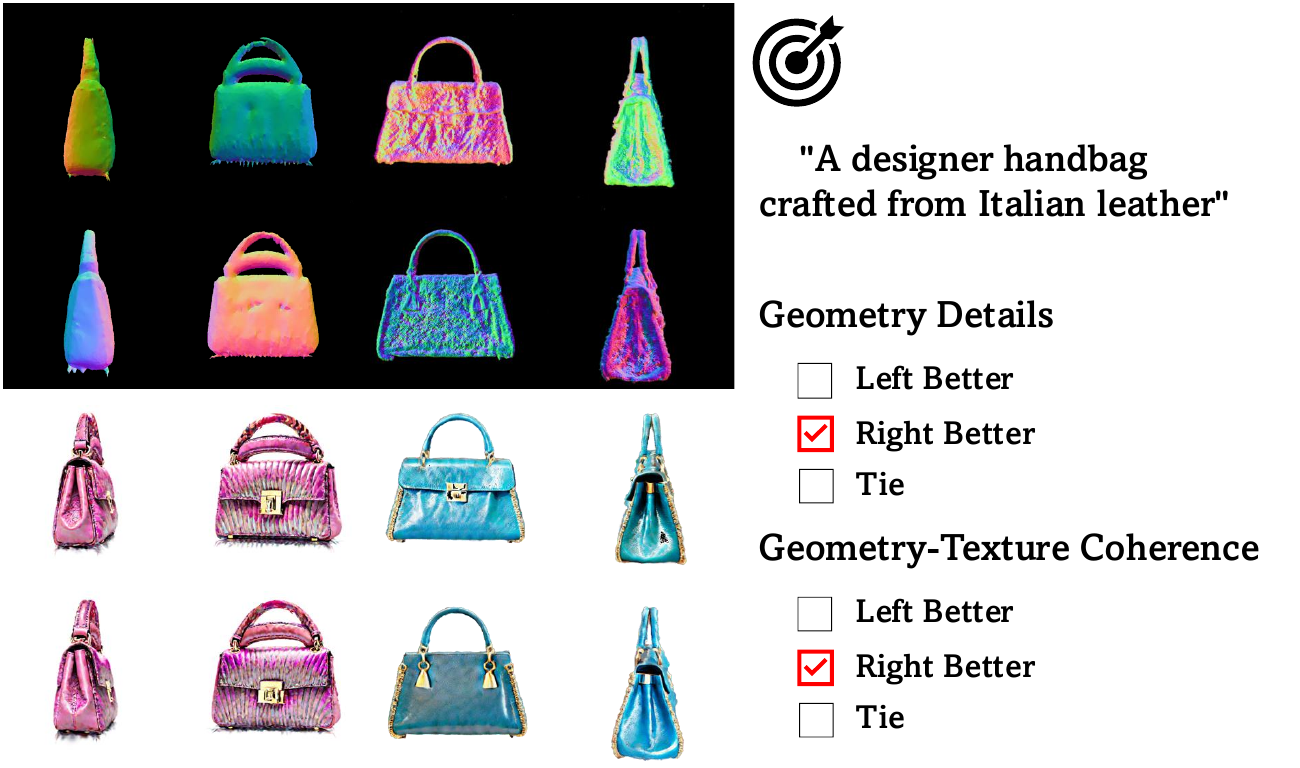}
    \includegraphics[width=0.49\textwidth]{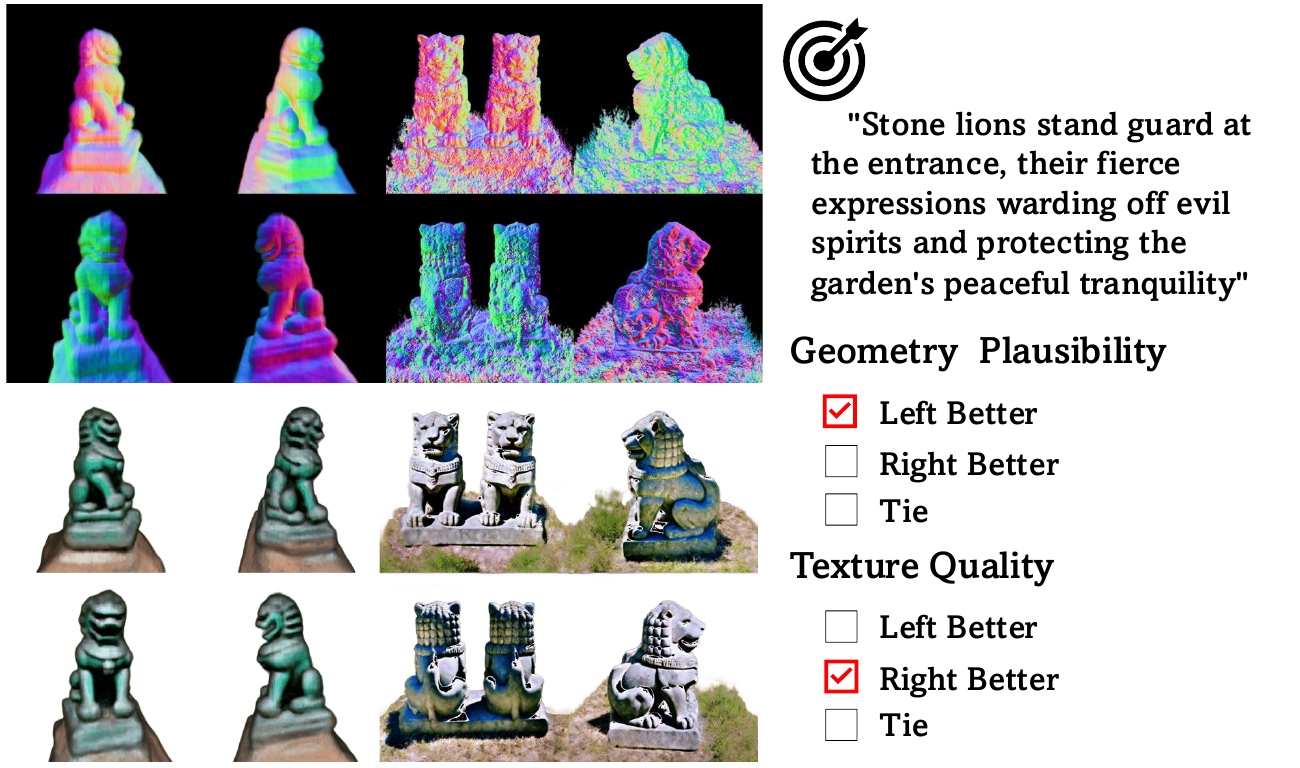}
    \includegraphics[width=0.49\textwidth]{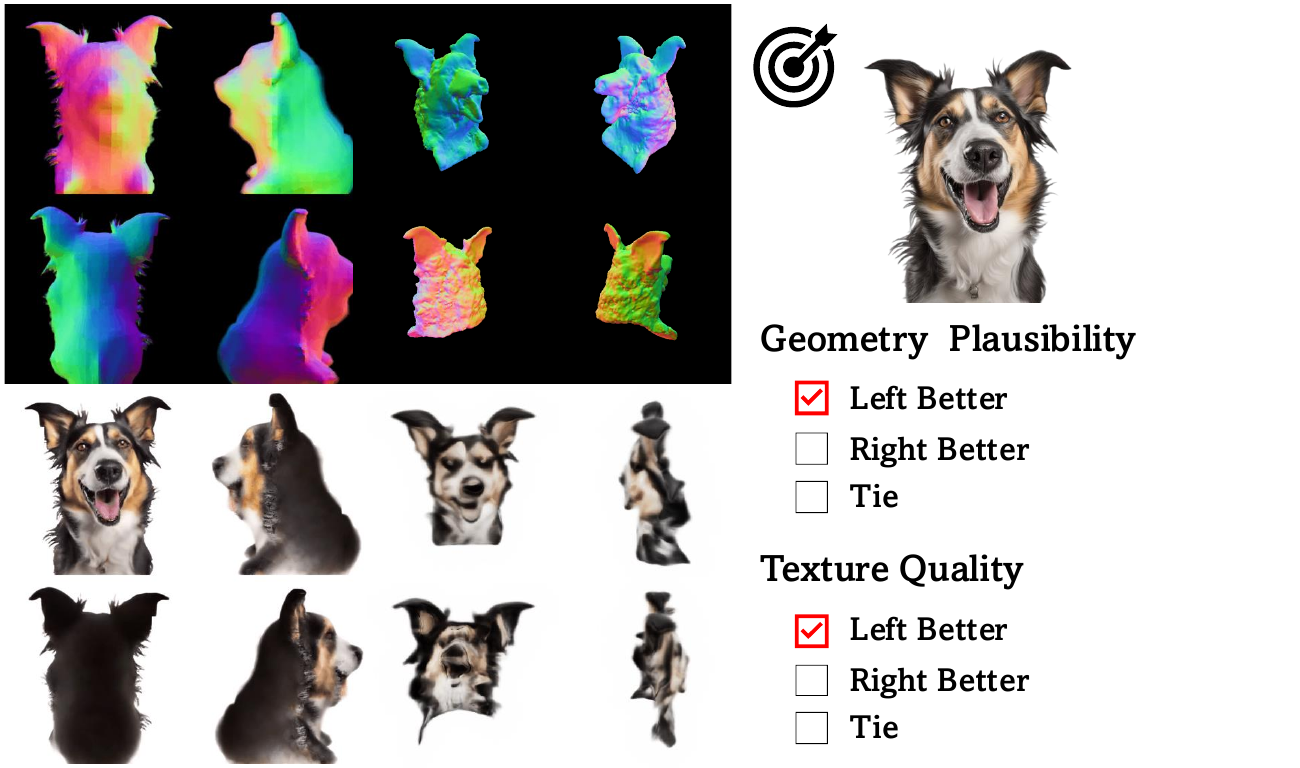}
    \includegraphics[width=0.49\textwidth]{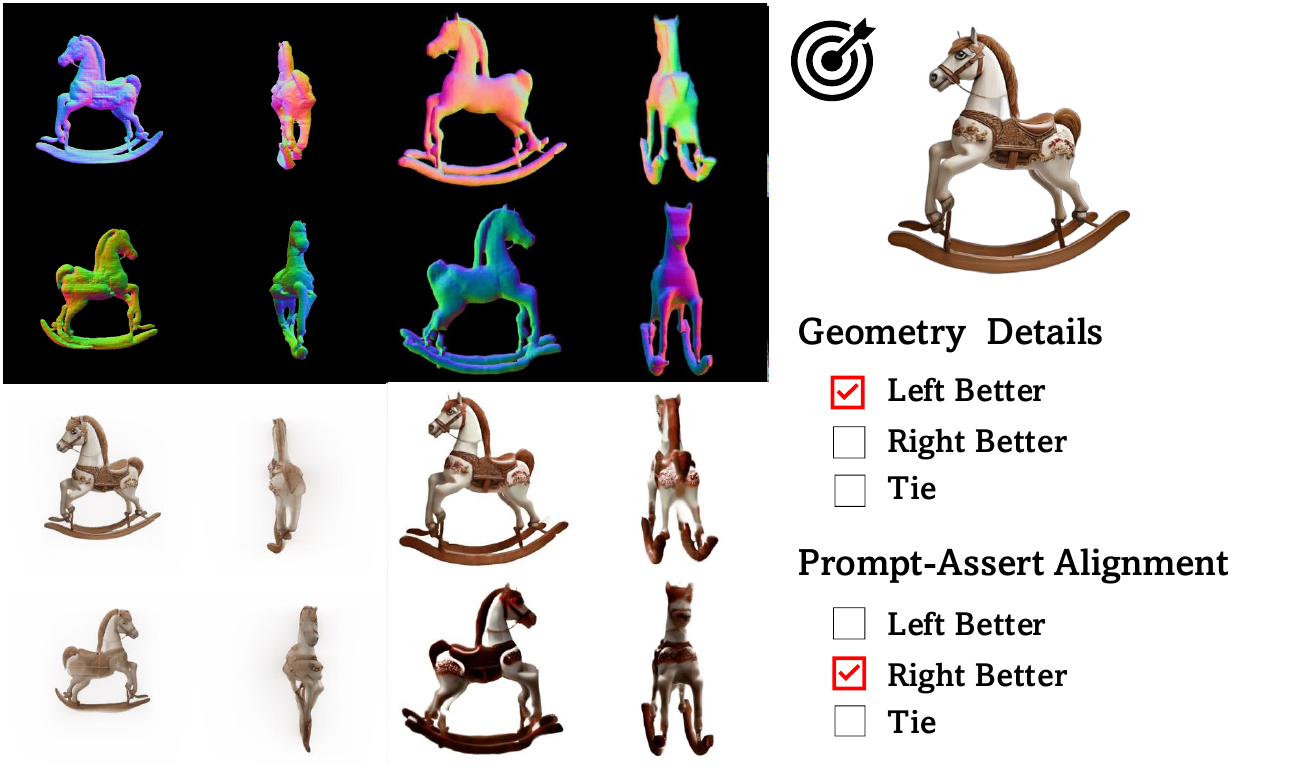}
    \caption{\textbf{Examples of evaluation criteria.} We try to give an intuitive explanation by means of pairwise comparison, where the top row for text-to-3D generation and the bottom row for image-to-3D generaion. Notice that the five dimensions are mutually independent, each with its own emphasis.
    }
    \label{fig:enter-label}
\end{figure*}

\subsection{Multi-criteria Human Voting Collection}
\label{sec3.2:multi-criteria_annotation}

\subsubsection{Evaluation Criteria}
\label{sec3.2.1:dataset_criteria}
When evaluating the quality of 3D model generation, we typically consider the following five criteria as outlined in GPTEval3D~\citep{Gpt4v_evaluation}, with some slight adjustments: 

\noindent{\textbf{1) Geometry Plausibility}} assesses whether the general shape resemble the real object logically and physically. A plausible 3D asset is expected to have recognizable structures and avoid improbable features, such as distorted faces (Multi-face Janus problem), floating fragments or noisy geometry. 

\noindent{\textbf{2) Geometry Details}} assesses the fineness and intricacy of shapes. High-detail models exhibit well-defined and intentional surface features, carefully distinguishing them from noise, whereas low-detail models may appear rough and simplistic.

\noindent{\textbf{3) Texture Quality}} underscores both aesthetic and consistency. The former includes aspects like realistic material representation and appropriate coloring, while the latter emphasizes uniformity and seamlessness across different viewing angles.

\noindent{\textbf{4) Geometry-Texture Coherence}} evaluates the alignment between geometric and texture characteristics. Textures should conform naturally to the model's contours without distortion and maintain a consistent appearance without compensating for or obscuring geometric details.

\noindent{\textbf{5) Prompt-Asset Alignment}} ensures that generated 3D assets accurately reflect their prompts in terms of category, quantity, attributes, and spatial relationships. For image-to-3D generations, maintaining identity consistency is also essential.

\noindent{}Figure~\ref{fig:enter-label} illustrates several examples of these criteria in a pairwise comparison style.

\subsubsection{3DGen-Arena}
\label{sec3.2.2:3dgen_arena}
Inspired by previous works~\citep{Chatbot-Arena, ImagenHub, tts-arena, WildWision-Arena}, we introduce \textit{3DGen-Arena}, the first arena-style benchmark platform in the 3D field. This platform employs an anonymous pairwise comparison manner, augmented with "named battles" and "direct chat" functionalities to encourage richer interactions. Specifically, visitors can sample prompts while the models in the battle remain anonymous until all five voting dimensions have been completed. For each dimension, visitors can choose from the options: \textit{`Model A is better', `Model B is better', `Tie', `Both bad'}.
Moreover, to facilitate user understanding, instead of simply displaying the generated 3D assets, \textit{3DGen-Arena} presents them through three distinct 360° panoramic videos: normal maps, textureless geometry, and fully textured renderings. The sreenshots of user interfaces can be found in Appendix~\ref{secA3.1:appendix_arena}. 
Additionally, we have received 8,045 anonymous votes from the public to date, with the corresponding leaderboard shown in Appendix~\ref{secA4:appendix_leaderboard}.

\subsection{Large-scale Human Annotation}
\label{sec3.3:human_annotation}
In addition to anonymous voting, we collect a substantial set of reliable labels from expert annotators. Specifically, we meticulously instruct 47 professional annotators with detailed guidelines and closely monitor their performance through regular feedback. Annotators are compensated at a rate of \$6 per hour, resulting in a total expenditure of approximately \$4,700. 
Additionally, during the annotation process, they are required to complete two types of annotations, as detailed below:
\textbf{1) Comparison votes} in a battle format, where annotators are asked to choose the better one given two anonymous models generated from the same prompt.
\textbf{2) Absolute scores} in numerical format, where annotators are asked to provide a reasonable value within a predefined range for each dimension given a single model.

\subsubsection{Comparison Votes}
\label{sec3.3.1:comparison_votes}

\noindent{\textbf{Rules.}}
We reuse the same annotation rules as 3DGen-Arena. Given a text/image prompt $p_i$, we sample two 3D models generated from different methods, ($S_0$, $S_1$), and showcase their 360° panoramic videos \{$(V_{i\_geo}$, $V_{i\_normal}$, $V_{i\_rgb}), i \in [0, 1]$\}. Annotators are expected to vote for the five dimensions outlined in Section~\ref{sec3.2.1:dataset_criteria} in a single route.
More details about the annotation platform can be found in Appendix~\ref{secA3.2:appendix_tag_platform}. 

\noindent{\textbf{Statistics.}
We randomly sampled 13,680 battle pairs and organized them into 456 packs, each containing 30 pairs. Finally, we collected 114,576 raw votes (the sum of 5 dimensions). Then, we employ three validation strategies simultaneously to clean raw votes: 1) We compute the strong-conflict ratio (when "left vote" meets "right vote") against valid pairs (287 pairs, 2\%) labeled by us; 2) We set up cross-annotation packs (240 packs, 52.6\%), which are labeled by two isolated annotators, and then compute the strong-conflict ratio between them; 3) We compute the "tie/both bad vote" ratio to identify potential lazy annotators. We revise our data with above three ratios, and ultimately obtain 68,400 unique votes.

\subsubsection{Absolute Scores}
\label{sec3.3.2:absolute_score}
\noindent{\textbf{Rules.}}
Given a text/image prompt $p_i$, we want to assign a score to the quality of a generated 3D model in absolute terms. Even with highly detailed instruction documentation, it remains challenging to ensure scoring consistency. To mitigate this, we display all 3D models (9 for text, 13 for image) generated from the same prompt simultaneously. Annotators are required to rank these models based on different dimensions before providing scores. We apply three score ranges based on the complexity of distinct dimensions: [0, 9], [0, 4], and [0, 1].
The screenshot of the annotation platform can be found in Appendix~\ref{secA3.2:appendix_tag_platform}. 

\noindent{\textbf{Statistics.}}
We generated 11,220 3D models and assigned two independent annotators to each model. Finally, we collected 124,695 raw scores (the sum of 5 dimensions). Similarly to preference data, we also employ three validation strategies: 1) We check the consistency between labeled ranking and score order; 2) We compute the error rate against valid samples (336 models, 3\%) labeled by us; 3) We compute the conflict ratio between two annotators. Noting that quality scoring is a very subjective task, we allow a certain degree of deviation when calculating the error rate and conflict ratio. In this paper, we set the absolute deviation threshold to 1. The final score for each dimension is weighted by two records, totaling 56,100 entries.

\begin{figure*}[t]
    \centering
    \includegraphics[width=\linewidth]{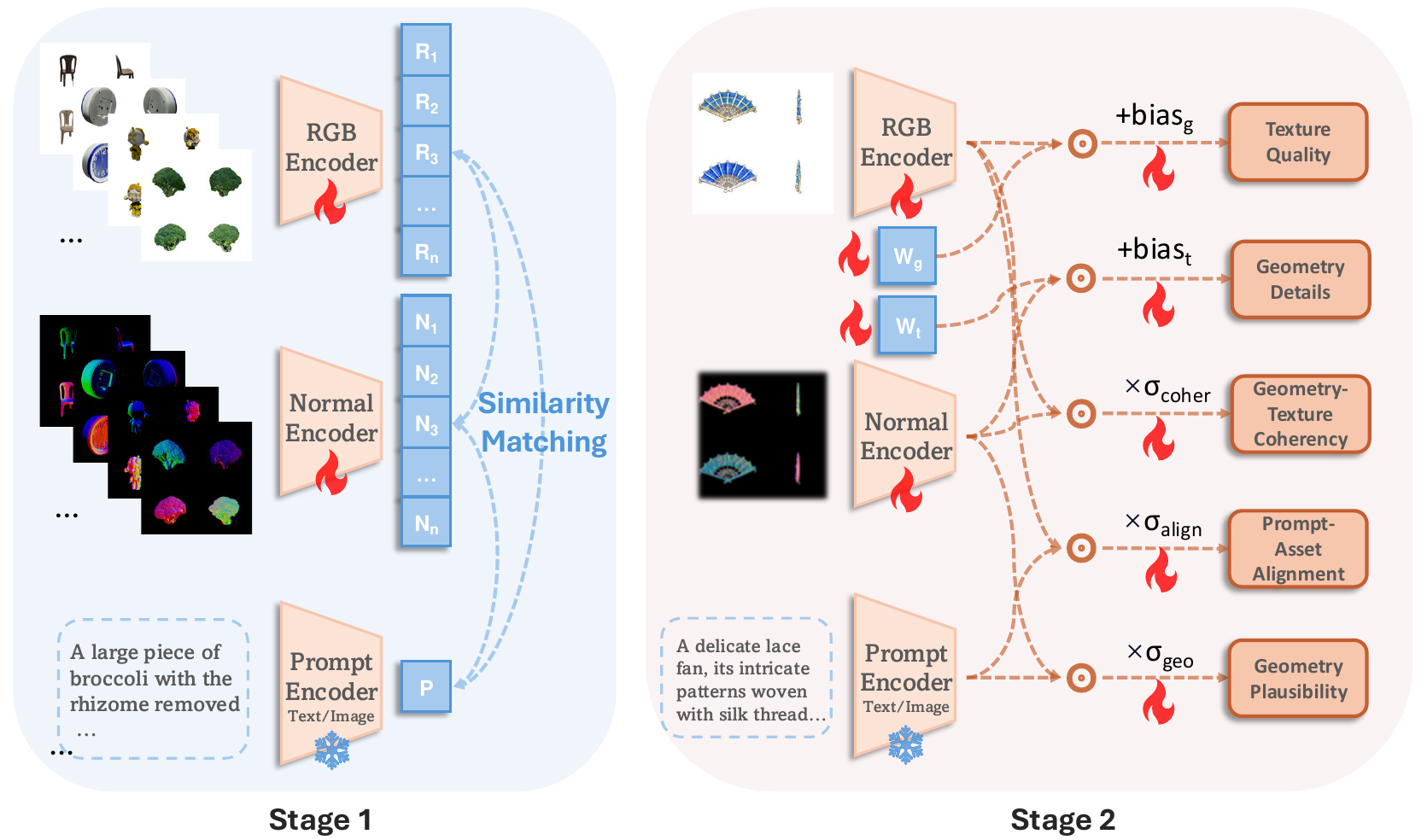}
    \vspace{-10px}
    \caption{\textbf{Overview of 3DGen-Score Model.} The model consists of three encoder modules, which can simultaneously process both multi-view normal and RGB images and support both text and image prompts. It works in a pairwise comparison style, and return a five dimension win-rate tuple for each input 3D model. And we apply a "two-stage" training stage. As shown in the left, we finetune visual encoders just like original CLIP to align embedding spaces in the first stage. Then in the second stage (right), we train the five predictors with the supervision of \textit{Comparison votes} data.}
    \label{fig:score_model}
\end{figure*}

\section{3DGen-Evaluator}
\label{sec4:ScoreModel}

An intuitive downstream application of our collected data is training automated evaluators. Given that 3D and video encoding have yet to reach the same level of maturity as image encoding, we propose utilizing multi-view image encoding as an alternative to direct 3D representation in this work.
Building on this idea, we propose the 3DGen-Score model, a CLIP-based evaluator, which inherits the advantages of CLIP-like architectures, including open-domain adaptability, ease of expansion, and computational efficiency. However, since it primarily returns numerical scores, its interpretability remains somewhat limited. 
To address this, we further introduce the 3DGen-Eval model, an MLLM-based evaluator, which leverages the strong visual-text reasoning capabilities of large models, offering enhanced open-world generalization and an interpretable reasoning process. 
However, MLLM-based models often suffer from inherent irreproducibility, where variations in prompts or even input timing can lead to different outputs. Therefore, these two models complement each other in practice, jointly forming a comprehensive evaluation system.

\subsection{CLIP-based 3DGen-Score Model}
\label{sec4.1:3degn_score_model}
Traditional CLIP-based metrics are designed for Image Generation tasks, primarily to evaluate the similarity between images and text. However, when applied to 3D generation, these metrics often overlook adverse aspects such as cross-view consistency and unrealistic repetition due to the lack of information exchange between views. To alleviate this, we modify the original CLIP framework to enable the simultaneous processing of multi-view images. Additionally, we incorporate normal maps, as they tend to capture geometric features more accurately than RGB images.

\noindent{\textbf{Model.}}
Our \textit{3DGen-Score} model $S$, as shown in Figure~\ref{fig:score_model}, is equipped with three encoder modules: a prompt encoder module $E_{p}$, a normal visual encoder $E_{n}$ and a RGB visual encoder $E_{r}$. 
Specifically, to support text and image prompts simultaneously, the prompt encoder module incorporates two distinct modal encoders, just like the original CLIP. The addition of two extra visual encoder modules allows for the separation of multi-view encoding from prompt encoding, simplifying the training process and facilitating easier expansion.

 \noindent{\textbf{Formula.}}
 To enhance the interpretability of our score model, we approximate each evaluation dimension in an abstract sense. For example, the \textit{Prompt-Asset Alignment} dimension is usually approximate as the similarity between RGB renderings and prompts. Similarly, we could calculate the similarity between normal renderings and prompts to estimate \textit{Geometry Plausibility} and the similarity between normal renderings and RGB renderings to estimate \textit{Geometry-Texture Coherence}. As for the \textit{Geometry Details} and \textit{Texture Quality}, illuminated by the LAION Aesthetic Predictor~\citep{Laion-5B}, we simplify them into two linear predictors. To be specific, given the prompt $p$ and the multi-view normal renderings $I_{n}$ and RGB renderings $I_{r}$ of a 3D model, the five dimensions can be formulated as follows:
\begin{align}
S_{geo}(p, I_{n}) &= \sigma_{geo} \cdot E_{p}(p) \cdot E_{n}(I_{n}), \\
S_{align}(p, I_{r}) &= \sigma_{align} \cdot E_{p}(p) \cdot E_{r}(I_{r}), \\
S_{coher}(I_{n}, I_{r}) &= \sigma_{coher} \cdot E_{n}(I_{n}) \cdot E_{r}(I_{r}), \\
S_{geo\_detail}(I_{n}) &= W_g \cdot E_{n}(I_{n}) + bias_g, \\
S_{texture}(I_{r}) &= W_t \cdot E_{r}(I_{r}) + bias_t,
 \end{align}
 where $\sigma_{geo}$, $\sigma_{align}$, $\sigma_{coher}$ are learnable scalar temperature parameters similar to CLIP, and $W_g$, $bias_g$, $W_t$, $bias_t$ are learnable parameters.

\begin{figure*}[t]
    \centering
    \includegraphics[width=\linewidth]{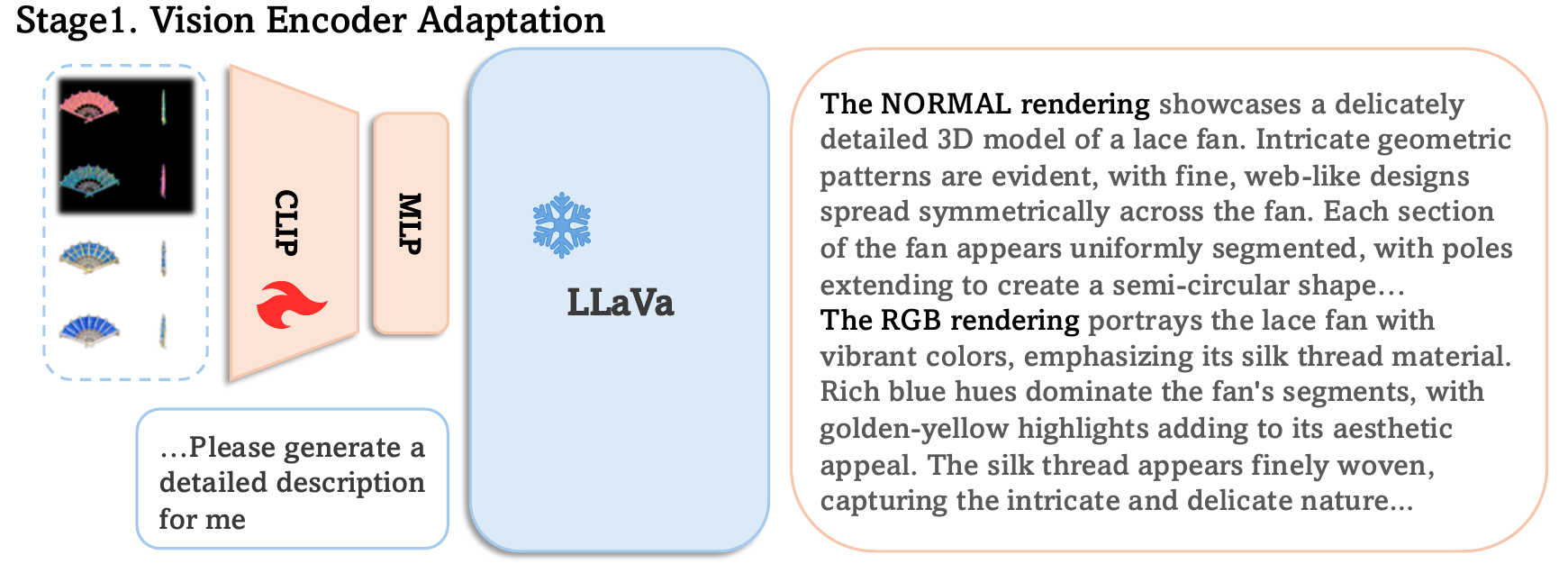}
    \vspace{-5px}
    \caption{\textbf{Vision Encoder Adaption}, the first training stage of \textbf{3DGen-Eval} model. In this stage, we serve the model as a captioner for multi-view images, supervised by captions generated by GPT-4V.}
    \label{fig:LlaVa_pretrain}
\end{figure*}

\noindent{\textbf{Training.}}
We initialize our model from \texttt{CLIP-ViT-H/14}~\citep{Clip, OpenCLIP}, and keep the prompt encoders frozen during training. Noting that CLIP has merely seen four-view normal and RGB images during the pretraining process, we propose a "two-stage" training strategy, shown in Figure~\ref{fig:score_model}. 
In the first stage, we fine-tune vision encoders only with contrastive loss to align the embedding spaces of concatenated normal and RGB renderings with the prompt embeddings. However, images rendered from generated models are often imperfectly aligned with their prompts or even severely collapsed. To address this, we additionally incorporate 3D assets from OmniObject3D~\citep{omniobject3d} and Cap3D~\citep{cap3d} datasets, which not only exhibit high-quality geometry and texture but are also well aligned with their annotations. Specifically, for each asset, we take its caption annotation as the text prompt and its front-view rendering image as the image prompt.
Then, in the second stage, we optimize the parameters of scoring functions supervised by human preference labels. The objective of the second training stage is to minimize the KL-divergence between the softmax-normalized scores and ground-truth labels, formulated as below. 
Notice that both training stages are trained on the \textit{Comparison votes} data. Meanwhile, rather than unfreezing all layers of the vision encoders, we selectively unfreeze specific layers during training, illustrated by~\citep{ShareGPT4v, bootstrap3d}. In this paper, we only unfreeze the last 4 layers.
We conducted experiments on two A100 GPUs for 4,000 steps, requiring 5 hours for a single stage with a batch size of 32 and a learning rate that warms up to 3e-6 over the first 500 steps.
\begin{small}
\begin{gather}
    min_\theta~KL(\hat{P}||P) = \sum_{i \in \{0,1\}} p_{i}\cdot log(p_{i}-\hat{p_{i}}), \\
    \hat{p_{i}} = \frac{exp(s_{i})} {exp(s_{0}) + exp(s_{1})}, i \in \{0,1\},
\end{gather}
\end{small}
\noindent{}where the ground truth $p_i$ $\in$ $\{0, 0.5, 1\}^5$, representing "win vote", "tie vote" and "fail vote", respectively.

\begin{figure*}[t]
    \centering
    \includegraphics[width=\linewidth]{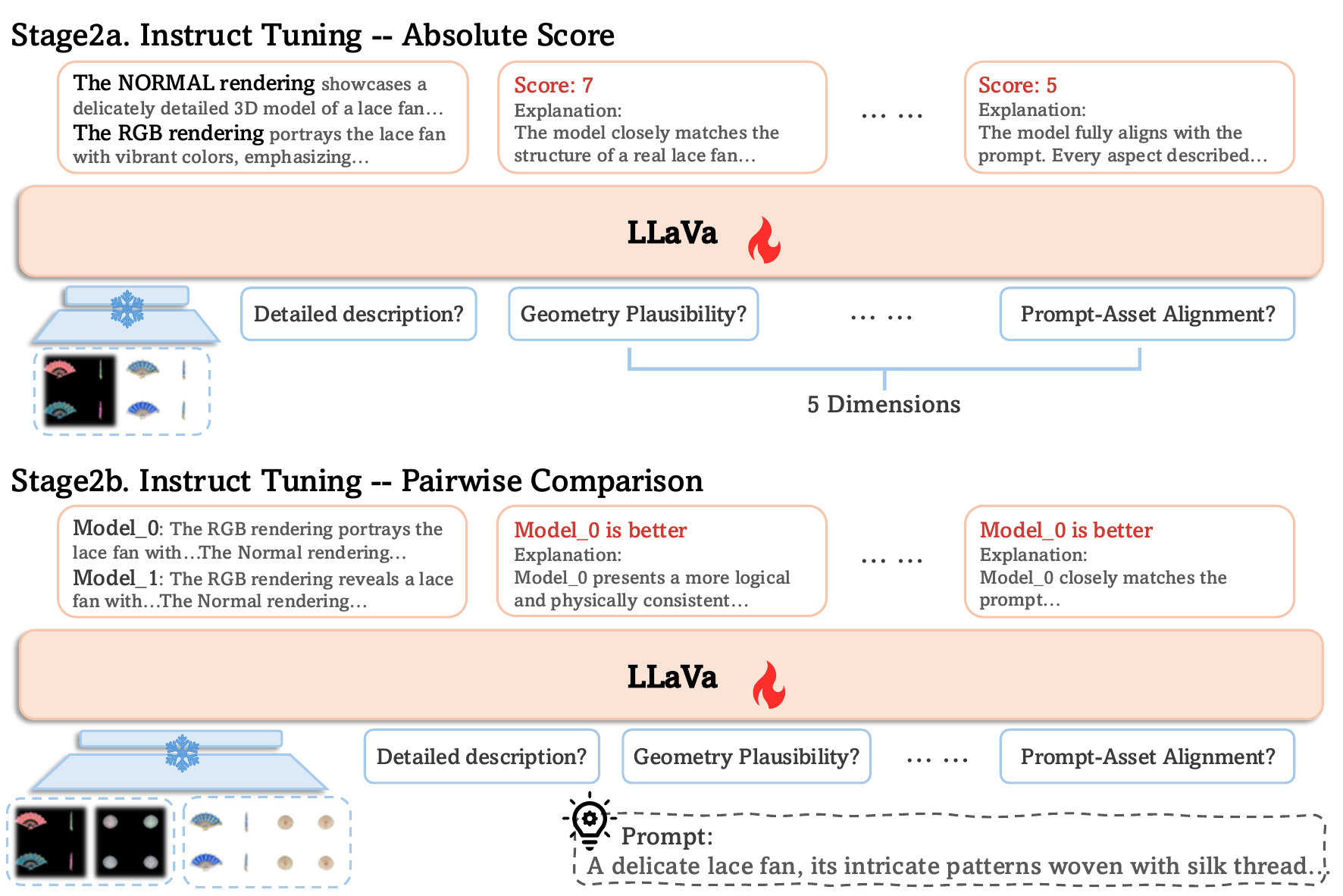}
    \vspace{-5px}
    \caption{\textbf{Instruct Tuning}, the second training stage of \textbf{3DGen-Eval} model, consisting of two sub-modules. The top sub-module is trained on \textit{Absolute Score} data, to help the model establish an overall understanding of each evaluation dimension. In the subsequent sub-module, the model learns to identify subtle differences between instances within each dimension, using \textit{Pairwise Comparison} data for supervison. Noticing that despite human annotations, we also encourage LLaVa to generate some explanations, guided by responses from GPT-4V, to enhance interpretability.}
    \label{fig:LlaVa_finetuen}
\end{figure*}

\subsection{MLLM-based 3DGen-Eval Model}
\label{sec4.2:3deval_model}
Building on the success of MLLMs, some explorations~\citep{Gpt4v_evaluation, bootstrap3d} have advocated the use of MLLMs to simulate human experts for evaluation. However, since GPTEval3D~\citep{gpt4point} calls the black-box GPT-4V directly, the inevitable and unpredictable bias often causes it to deviate from human judgments. Although MV-LLaVa~\citep{bootstrap3d} has been fine-tuned to preference data, potential biases remain, as pseudo-labels are generated by GPT-4V. To alleviate this, we propose fine-tuning MV-LLaVa with our human annotations. 

\noindent{\textbf{Preliminary.}}
MV-LLaVA~\citep{bootstrap3d} is a caption generation and overall quality scoring model for 3D assets, fine-tuned from instructive conversation pairs generated by GPT-4V. It takes four-view images as input and processes visual signals by feeding them into LLaVa~\citep{LLaVa} in the form of separate CLIP embeddings. Inspired by ShareGPT-4V~\citep{ShareGPT4v}, the model trains the vision encoder only in the first stage, using captions for supervision to enhance multi-view awareness and texture perception. Subsequently, the model evaluates quality based on the multi-view images and captions through a chain-of-thought reasoning process.
While MV-LLaVA relies solely on RGB renderings as input and is supervised by GPT-4V, it tends to provide only a coarse evaluation of 3D models and frequently deviates from humans.

\noindent{\textbf{Model.}}
We follow the same model structure as MV-LLaVa~\citep{bootstrap3d}, equipped with two primary modules: CLIP for visual embedding and LLaVa for evaluation in a style of Question\&Answer(QA). 
Instead of separate four-view RGB images, our \textit{3DGen-Eval} model takes concatenated images arranged in a 2*2 grid format as input and additionally incorporates normal renderings to enhance the evaluation of geometry-related dimensions.

\begin{figure*}[t]
  \centering
   \includegraphics[width=\linewidth]{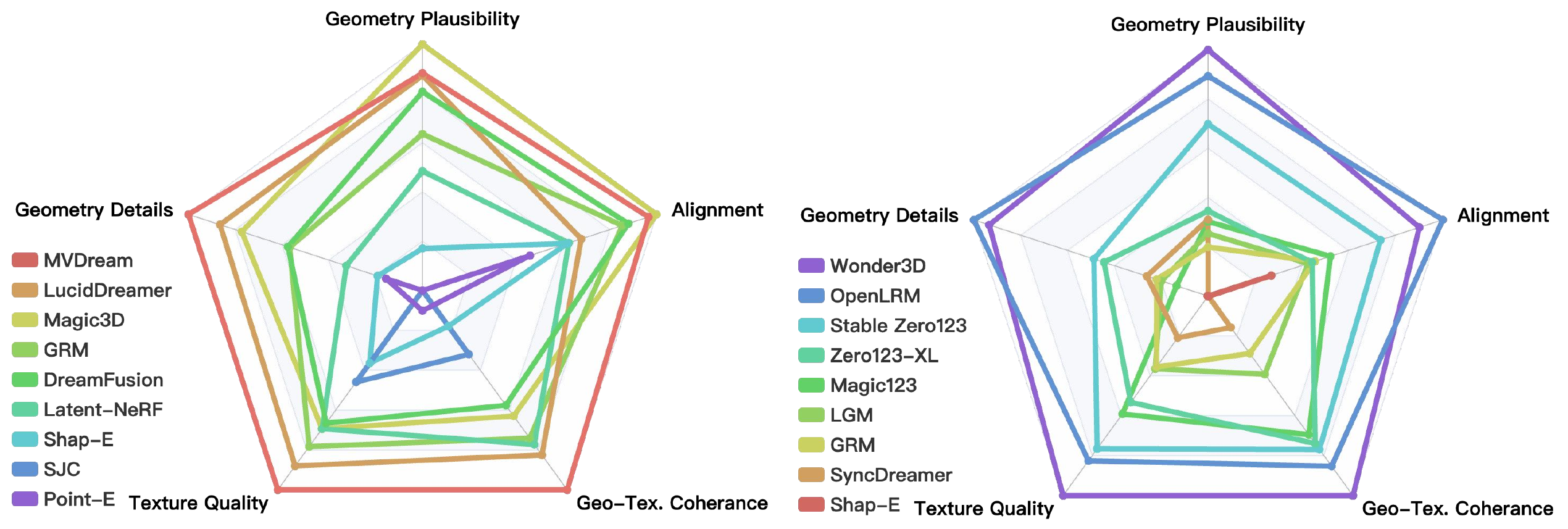}
   \vspace{-10px}
   \caption{\textbf{Radar visualization of Elo scores based on human annotations.} \textbf{Left:} text-to-3D generative models. \textbf{Right:} The top-9 image-to-3D generative models. The legend in the bottom-left corner of each image lists the model ranking based on the average Elo scores across five dimensions, from top to bottom. The detailed leaderboard can be checked in appendix~\ref{secA4:appendix_leaderboard}}
   \label{fig:radar}
\end{figure*}

\noindent{\textbf{Training.}}
We initialize our \textit{3DGen-Eval} model with MV-LLaVa and perform a "two-stage" training process. In the first stage, we finetune the visual encoder while keeping LLaVA frozen, allowing the model to learn to understand images in both the concatenated format and the normal format, as illustrated in Figure~\ref{fig:LlaVa_pretrain}. In the subsequent "Instruct Tuning" stage, we introduce two sub-modules designed to adapt to our data and distinct objectives, as shown in Figure~\ref{fig:LlaVa_finetuen}, which separately learn to assign an absolute score for a single 3D model and to compare between two models over five dimensions. Moreover, aimed at enhancing the confidence of the evaluator, except solely supervised by the human annotations, the model is also encouraged to generate some reasonable explanations, supervised by responses from GPT-4V. More details of the prompts used for pseudo-labels and templates used in training can be found in appendix~\ref{secA6.3:appendix_3dgen_eval_prompt}

\section{Evaluation}
\label{sec5:Evaluation}
In this section, we conduct a preliminary evaluation to assess the alignment of our metric with human judgments. To be specific, we introduce the baseline metrics in Section~\ref{sec5.1: base_metrics}, and present the human evaluation results based on the preference data in Section~\ref{sec5.2:human_evaluation}. Subsequently, we compare our models with baseline metrics in terms of the alignment of human preference in Section~\ref{sec5.3:human_correlation}. Furthermore, we conduct validation experiments for the generalization of our models in Section~\ref{sec5.4:generalization} and ablation experiments for the training strategy in Section~\ref{sec5.5:ablation_experiments}. Finally, we summarize the advantages of our two models and give some recommendations for automated evaluation in Section~\ref{sec5.6:discussion}

\begin{table*}[t]
    \caption{\textbf{Text-to-3D pairwise rating alignment with human judgment.} We assess the average probability of alignment with human judgments for each comparison. Our method demonstrates strong alignment across all criteria, with higher values indicating better performance. \textbf{Bold} and \underline{underlined} results indicate the best and second-best performers.}
    \label{pair-text}
    \centering
    \small
    \setlength{\tabcolsep}{8pt}
    \renewcommand{\arraystretch}{1.0}
    \begin{tabular}{lcccccc}
    \toprule
       Methods & Plausibility & Geo. Details & Tex. Quality & Geo-Tex. & Alignment & Average\\     
       \midrule
        CLIP & $0.666$ & $0.640$ & $0.684$ & $0.702$ & $0.612$ & $0.661$ \\
        Aesthetic score & $0.524$ & $0.536$ & $0.557$ & $0.549$ & $0.508$ & $0.535$ \\ 
        HPSv2.1 & $0.505$ & $0.505$ & $0.507$ & $0.507$ & $0.503$ & $0.505$ \\ 
        PickScore & $0.625$ & $0.646$ & $0.684$ & $0.705$ & $0.567$ & $0.645$ \\ 
        OpenShape & $0.502$ & $0.501$ & $0.500$ & $0.500$ & $0.502$ & $0.501$ \\ 
        MV-LLaVa & $0.493$ & $0.514$ & $0.495$ & $0.490$ & $0.511$ & $0.501$ \\
        GPTEval3D & $0.660$ & \underline{$0.667$} & \underline{$0.699$} & \textbf{$0.767$} & $0.592$ & \underline{$0.677$} \\ 
        \midrule
        3DGen-Eval(Ours) & \underline{$0.687$} & $0.665$ & $0.689$ & $0.696$ & \underline{$0.625$} & $0.672$ \\ 
        3DGen-Score(Ours) & \textbf{$0.729$} & \textbf{$0.707$} & \textbf{$0.760$} & \underline{$0.764$} & \textbf{$0.667$} & \textbf{$0.725$} \\
    \bottomrule
    \end{tabular}%
\end{table*}

\begin{table*}[t]
  \caption{\textbf{Image-to-3D pairwise rating alignment with human judgment.} We assess the average probability of alignment with human judgments for each comparison. Our method demonstrates strong alignment across all criteria, with higher values indicating better performance.}
  \label{pair-image}
  \centering
  \small
  \setlength{\tabcolsep}{8pt}
  \renewcommand{\arraystretch}{1.0}
  \begin{tabular}{lcccccc}
    \toprule
       Methods & Plausibility & Geo. Details & Tex. Quality & Geo-Tex. & Alignment & Average\\     \midrule
        CLIP similarity & 0.529 & 0.523 & 0.542 & 0.538 & 0.524 & 0.531 \\ 
        Aesthetic score & 0.560 & 0.560 & 0.567 & 0.566 & 0.545 & 0.560 \\  
        OpenShape & 0.500 & 0.501 & 0.499 & 0.499 & 0.500 & 0.500 \\ 
        MV-LLaVa & 0.463 & 0.447 & 0.441 & 0.45 & 0.485 & 0.457 \\
        GPTEval3D & 0.703 & 0.676 & 0.714 & 0.701 & \underline{0.654} & 0.670 \\ 
        \midrule
        3DGen-Eval(Ours) & \underline{0.744} & \underline{0.750} & \underline{0.754} & \underline{0.755} & 0.651 & \underline{0.731} \\
        3DGen-Score(Ours) & \textbf{0.777} & \textbf{0.768} & \textbf{0.778} & \textbf{0.798} & \textbf{0.712} & \textbf{0.767} \\
    \bottomrule
  \end{tabular}
\end{table*}

\subsection{Baseline Metrics}
\label{sec5.1: base_metrics}

We selected 7 evaluation metrics based on various considerations. 
\textbf{1) CLIP Score} \citep{CLIPScore} is widely used for \textit{Text–Asset Alignment}. The compatibility of normal-caption pairs helps measure \textit{Geometry Plausibility}. In this paper, we use \texttt{CLIP-ViT-H/14}.
\textbf{2) CLIP Similarity} measures the cosine distance between the CLIP features of the multi-view renderings and the image prompt, which is used to assess \textit{Image–Asset Alignment}.
\textbf{3) Aesthetic Score}~\citep{Laion-5B} is a CLIP-based linear estimator that predicts the aesthetic quality of images, which can be used to measure the \textit{Texture Quality}.
\textbf{4) HPSv2.1}~\citep{HPSv2} and \textbf{5) PickScore}~\citep{Pick-a-Pic} are human preference scoring models trained on user preference datasets. They are designed to predict preferences for generated images.
\textbf{6) OpenShape}~\citep{openshape} is a CLIP-based point cloud embedding. We calculate its similarity with the CLIP embedding of the input prompt.
\textbf{7) GPTEval3D}~\citep{Gpt4v_evaluation} is a human-aligned evaluator for text-to-3D generation utilizing GPT-4V. For image-to-3D generation, we use BLIP~\citep{BLIP} to obtain text captions.

\noindent{}All metrics are calculated on the annotated test dataset. For fairness, we average the four-view scores as the final value. Except for GPTEval3D~\citep{Gpt4v_evaluation}, limited by API, we uniformly sample 40 objects as test data. Additionally, due to its formatting restriction and modality limitation, we cannot achieve a fair comparison with \textbf{$T^3$Bench}~\citep{T3_Bench} in this section. However, we provide supplementary comparison experiments in Appendix~\ref{secA6.2:appendix_exp_t3bench}.

\begin{table*}[t]
  \caption{\textbf{Text-to-3D ranking alignment with human judgment.} The table illustrates Kendall’s tau ranking correlation~\citep{Kendall} between Text-to-3D rankings produced by various metrics and those determined by human experts. A higher correlation suggests a stronger alignment with human judgment. \textbf{Bold} and \underline{underlined} results indicate the best and second-best performers.}
  \label{rank-text}
  \centering
  \small
  \setlength{\tabcolsep}{8pt}
  \renewcommand{\arraystretch}{1.0}
  \begin{tabular}{lcccccc}
    \toprule
       Methods & Plausibility & Geo. Details & Tex. Quality & Geo-Tex. & Alignment & Average\\     \midrule
        CLIP-Score & \underline{0.667} & \underline{0.611} & \underline{0.722} & \underline{0.833} & 0.333 & 0.635 \\
        Aesthetic score & 0.333 & 0.389 & 0.556 & 0.667 & 0.167 & 0.422 \\ 
        HPSv2.1 & 0.389 & 0.444 & 0.611 & 0.722 & 0.222 & 0.478 \\ 
        PickScore & 0.333 & 0.389 & 0.556 & 0.667 & 0.167 & 0.422 \\ 
        OpenShape & 0.071 & 0.071 & 0.000 & -0.143 & 0.143 & 0.029 \\ 
        MV-LLaVa & -0.278 & 0.056 & -0.056 & -0.111 & 0.111 & -0.056 \\
        GPTEval3D & 0.500 & 0.500 & \textbf{0.889} & \underline{0.833} & \underline{0.389} & 0.622 \\ 
         \midrule
        3DGen-Eval(Ours) & \textbf{0.722} & \textbf{0.722} & \underline{0.722} & \underline{0.833} & 0.333 & \underline{0.667} \\
        3DGen-Score(Ours) & \underline{0.667} & 0.556 & \textbf{0.889} & \textbf{0.944} & \textbf{0.500} & \textbf{0.711} \\
    \bottomrule
  \end{tabular}
\end{table*}

\begin{table*}[t]
  \caption{\textbf{Image-to-3D ranking alignment with human judgment.} The table illustrates Kendall’s tau ranking correlation~\citep{Kendall} between Image-to-3D rankings from various metrics and human experts. A higher correlation suggests a stronger alignment with human judgment.}
  \label{rank-image}
  \centering
  \small
  \setlength{\tabcolsep}{9.5pt}
  \renewcommand{\arraystretch}{1.0}
  \begin{tabular}{lcccccc}
    \toprule
       Methods & Plausibility & Geo. Details & Tex. Quality & Geo-Tex. & Alignment & Average\\     \midrule
        CLIP similarity & 0.256 & 0.154 & 0.256 & 0.256 & 0.128 & 0.210 \\ 
        Aesthetic score & 0.487 & 0.436 & 0.590 & 0.538 & 0.513 & 0.513 \\ 
        OpenShape & 0.600 & 0.527 & 0.636 & 0.636 & 0.636 & 0.607 \\ 
        MV-LLaVa & -0.179 & -0.231 & -0.308 & -0.282 & -0.154 & -0.231 \\
        GPTEval3D & 0.564 & 0.590 & 0.667 & 0.667 & 0.513 & 0.600 \\ 
         \midrule
        3DGen-Eval(Ours) & \underline{0.744} & \underline{0.744} & \underline{0.744} & \underline{0.795} & \underline{0.692} & \underline{0.744} \\
        3DGen-Score(Ours) & \textbf{0.897} & \textbf{0.872} & \textbf{0.821} & \textbf{0.923} & \textbf{0.769} & \textbf{0.856} \\
    \bottomrule
  \end{tabular}
\end{table*}

\subsection{Human Preference Evaluations}
\label{sec5.2:human_evaluation}
 In this section, we conduct a comprehensive assessment of each generative model based on \emph{human expert annotations}. Specifically, we calculate the average Elo scores for each model across the five criteria and present radar charts in Figure~\ref{fig:radar}, which depict the top 9 models for Text-to-3D and Image-to-3D tracks. 

\noindent{\textbf{Text-to-3D.}}
According to our criteria, the top-3 methods in text-to-3D track are MVDream~\citep{Mvdream}, LucidDreamer~\citep{Luciddreamer}, and Magic3D~\citep{Magic3D}, achieving average Elo scores of 1177.66, 1112.21, and 1088.93, respectively.
These results shed light on the distinct strengths and weaknesses of each method, offering insights that can inform future improvements in 3D generation techniques.
The success of MVDream~\citep{Mvdream} demonstrates the advancement of multi-view diffusion model, which not only achieves the effective combination of 2D prior and 3D prior but also facilitates the information exchange between different views. Building on the 3DGS representation, LucidDreamer~\citep{Luciddreamer} performs well on geometry and texture quality. Magic3D~\citep{Magic3D} benefits from a "two-stage" optimization strategy, significantly enhancing the geometric plausibility of generated 3D models. However, its texture quality remains somewhat unsatisfactory, primarily due to the inherent limitations of SDS loss.

\noindent{\textbf{Image-to-3D.}}
On the image-to-3D track, Wonder3D~\citep{Wonder3D}, OpenLRM~\citep{OpenLRM}, and Stable Zero123~\citep{Zero-1-to-3} emerge as the frontrunners, achieving average Elo scores of 1304.05, 1279.67, and 1200.69, respectively. 
Wonder3D~\citep{Wonder3D} employs a multi-view cross-domain attention mechanism, which again verifies the significance of 2D\&3D domain blending and multi-view consistency. Benefiting from Objarverse~\citep{Objaverse, Objaverse-XL} datasets, OpenLRM reinvigorates the transformer-based encoder-decoder architecture, enhancing the quality and diversity of generated 3D objects. Meanwhile, Stable Zero123~\citep{Zero-1-to-3} achieves a comprehensive improvement over Zero123-XL~\citep{Zero-1-to-3} in multiple dimensions, but still leaves room for improvement in geometric details. 

\begin{table*}[ht]
    \caption{\textbf{Pairwise rating alignment with additional Instant-Mesh.} We assess the average probability of alignment with human judgments for each comparison between Instant-Mesh and other 13 image-to-3D models.}
    \label{tab:pair-instantmesh}
    \centering
    \small
    \setlength{\tabcolsep}{8pt}
    \renewcommand{\arraystretch}{1.0}
    \begin{tabular}{lcccccc}
    \toprule
       Methods & Plausibility & Geo. Details & Tex. Quality & Geo-Tex. & Alignment & Average\\     \midrule
        CLIP similarity & 0.667 & 0.654 & 0.703 & 0.714 & 0.751 & 0.698 \\ 
        Aesthetic score & 0.529 & 0.544 & 0.558 & 0.547 & 0.559 & 0.547 \\  
        OpenShape & 0.544 & 0.580 & 0.603 & 0.587 & 0.553 & 0.573 \\ 
        MV-LLaVa & 0.553 & 0.550 & 0.558 & 0.565 & 0.535 & 0.552 \\
        GPTEval3D & \underline{0.711} & \textbf{0.782} & \underline{0.789} & \textbf{0.776} & \underline{0.757} & \underline{0.763} \\ 
        \midrule
        3DGen-Eval(Ours) & 0.700 & 0.712 & 0.727 & 0.708 & 0.692 & 0.708 \\
        3DGen-Score(Ours) & \textbf{0.754} & \underline{0.741} & \textbf{0.826} & \underline{0.765} & \textbf{0.815} & \textbf{0.780} \\
    \bottomrule
    \end{tabular}
\end{table*}

\begin{table*}[ht]
    \caption{\textbf{Pairwise rating alignment compared on 3DRewardDB.} We assess the average alignment across considered dimensions for each method.}
    \label{tab:pair-3DRewardDB}
    \centering
    \small
    \setlength{\tabcolsep}{7.5pt}
    \renewcommand{\arraystretch}{1.0}
    \begin{tabular}{lcccccc}
    \toprule 
        Methods & 3DGen-Score(Ours) & 3DGen-Eval(Ours) & GPTEval3D & PickScore & HPSv2.1 & CLIP\\   
       \midrule
        Alignment & \textbf{0.716} & 0.576 & 0.676 & 0.505 & 0.675 & 0.671 \\ 
    \bottomrule
    \end{tabular}%
\end{table*}

\subsection{Comparisons on Human Preference Correlation}
\label{sec5.3:human_correlation}
In this section, we assess the alignment between our proposed metric and human preferences. To accomplish this, we employ the Softmax function over basic metrics to assign a win probability for each evaluation match. Experimental results demonstrate a strong correlation between our models and human judgment.

\noindent{\textbf{Pairwise Alignment.}}
Let $p_i$ represent the predicted win rating of the left model player in the $i$-th battle and $q_i$ denote the human judgment of making the same choice. Thus, the pairwise rating alignment is given by $\mathbb{E}_i[p_iq_i +(1-p_i)(1-q_i)]$. A higher value signifies a stronger alignment with human experts.
 As demonstrated in Table~\ref{pair-text} and Table~\ref{pair-image}, our \textit{3DGen-Score} model consistently outperforms other metrics, highlighting its versatility across various evaluation criteria. Moreover, our \textit{3DGen-Eval} model achieves significant improvements in all dimensions compared to MV-LLaVa~\citep{bootstrap3d}. In some dimensions, it even outperforms GPT-4V, demonstrating the effectiveness of the fine-tuning process. 

\noindent{\textbf{Ranking Alignment.}}
Table~\ref{rank-text} and Table~\ref{rank-image} display the ranking correlations between various evaluation metrics and the reference Elo scores computed from the expert annotations. All metric scores, excluding GPTEval3D~\citep{gpt4point}, are based on the average scores of prompts in the test set, while GPTEval3D and our models calculate scores using the Elo rating system, starting with an initial score of 1,000 and a K factor of 32.
Table~\ref{rank-image} shows that our methods achievesthe highest ranking across all dimensions on the image-to-3D track, securing the 1st and 2nd positions.
As for the text-to-3D ranking, our \textit{3DGen-Eval} model demonstrates impressive performance in the geometric-related dimension, while our \textit{3DGen-Score} model achieves the highest overall ranking on average, with a notable improvement in "Prompt-Asset Alignment" dimension.

\noindent{\textbf{Analysis}}
Experimental results demonstrate that our models outperform all existing metrics across comprehensive evaluation dimensions, regardless of whether the assessment is conducted on pairwise alignment or ranking correlation and in both text-to-3D and image-to-3D tracks. Notably, among our models, the CLIP-based 3DGen-Score outperforms the MLLM-based 3DGen-Eval model. We attribute this to CLIP’s simpler architecture and fewer parameters, which enable more efficient training and contribute to stronger performance. 
Additionally, model performance is generally stronger on the image-to-3D track compared to the text-to-3D track. We hypothesize that this discrepancy arises from two key factors: first, the image-to-3D track benefits from a larger volume of training data; second, the process of aligning text and images within encoders may introduce biases. Furthermore, the performance of 3DGen-Score on the text-to-3D track in the ranking setting highlights the necessity of training 3DGen-Eval to improve generalization to normal images.

\begin{table*}[t]
    \caption{\textbf{Human alignment results of ablation experiments}. The first 3 columns demonstrate training strategiesm, where \textbf{C\_loss} means "contrastive loss" applied in stage-1, and \textbf{V\_enc} means "finetune vision encoders" applid in stage-2. The rest columns demonstrate the \textbf{pairwise alignment} on test data, and the 'Average' column is weighted by the count of involved models. The higher values indicate better alignment with human judgments.}
    \label{tab: ablation_training_alignment}
    \centering
    \small
    \setlength{\tabcolsep}{8pt}
    \renewcommand{\arraystretch}{1.0}

    \begin{tabular}{cc|ccccc|c}
    \toprule
    \multicolumn{2}{c|}{\centering Strategy} & \multicolumn{5}{c|}{\centering Dimensions} & \multirow{2}{*}{Avgerage} \\   
    \cmidrule(lr){1-2} \cmidrule(lr){3-7}
    C\_loss & V\_enc. & Plausibility & Geo. Details & Tex. Quality & Geo-Tex. & Alignment & \\
    \midrule
    \checkmark & & \textbf{0.759} & 0.714 & 0.735 & 0.759 & 0.684 & 0.730 \\
     & \checkmark  & 0.746 & \underline{0.742} & \underline{0.771} & \underline{0.776} & \underline{0.686} & \underline{0.744} \\
    \checkmark & \checkmark &  \underline{0.757}& \textbf{0.747} & \textbf{0.771} & \textbf{0.784} & \textbf{0.694} & \textbf{0.750} \\
    \bottomrule
    \end{tabular}
\end{table*}

\begin{table*}[t]
    \caption{\textbf{Model robustness results of ablation experiments}. We calculate the \textbf{pairwise alignment} on comparisons between Instant-Mesh and other 13 image-to-3D models. The higher values indicate better model robustness.}
    \label{tab: ablation_training_robustness}
    \centering
    \small
    \setlength{\tabcolsep}{8pt}
    \renewcommand{\arraystretch}{1.0}
    \begin{tabular}{cc|ccccc|c}
    \toprule
    \multicolumn{2}{c|}{\centering Strategy} & \multicolumn{5}{c|}{\centering Dimensions} & \multirow{2}{*}{Avgerage} \\   
    \cmidrule(lr){1-2} \cmidrule(lr){3-7}
    C\_loss & V\_enc. & Plausibility & Geo. Details & Tex. Quality & Geo-Tex. & Alignment & \\
    \midrule
    \checkmark & & \underline{0.715} & 0.599 & \underline{0.781} & 0.665 & 0.738 & 0.700 \\
     & \checkmark & 0.677 & \underline{0.757} & 0.774 & \underline{0.781} & \underline{0.800} & \underline{0.758} \\
    \checkmark & \checkmark &  \textbf{0.754}& \textbf{0.741} & \textbf{0.826} & \textbf{0.765} & \textbf{0.815} & \textbf{0.780} \\
    \bottomrule
    \end{tabular}
\end{table*}

\subsection{Generalization Performance}
\label{sec5.4:generalization}
In Section~\ref{sec5.3:human_correlation}, we conduct evaluation experiments only on our proprietary dataset, which may introduce specific biases and limit our ability to assess the model's performance and robustness.
To alleviate it, we expand our testing to include external data, such as new 3d models and external dataset.

\noindent{\textbf{Evaluation on New Model}}
To verify the robustness of our model on unseen data, we utilize Instant-Mesh~\citep{InstantMesh}, an advanced image-to-3D model, to generate 510 3D assets using our prompt set. We then randomly pair these assets with those from 13 other image-to-3D models, resulting in 130 comparison pairs. Subsequently, we calculate pairwise rating alignment between our model and annotations from experts, shown in Table~\ref{tab:pair-instantmesh}. 
The results indicate that our \textit{3DGen-Score} model exhibits strong generalization capabilities, allowing it to assess previously unseen models while surpassing existing methods across multiple dimensions. This underscores its effectiveness as a robust and reliable evaluation metric for 3D assessment tasks.

\noindent{\textbf{Evaluation on External Dataset}}
 We conduct experiments using 3DRewardDB~\citep{DreamReward}, a coarse preference dataset of generated text-to-3D models and, to our knowledge, the only available relative dataset. However, 3DRewardDB provides four-view perspectives only for each model, lacking normal maps or raw 3D assets. To address this limitation, we apply Metric3D~\citep{metric3d}, the state-of-the-art method for surface normal estimation, to predict normal maps. We then sample 120 comparison pairs as the test set. 
Notably, 3DRewardDB derives final scores based on alignment, consistency, and overall quality while omitting criteria such as Geometry Details and Geometry-Texture Alignment during its labeling process. To align its criteria with ours, we exclude those less-considered dimensions and aggregate scores for "3D Plausibility", "Texture Quality", and "Asset-Prompt Alignment" to compute a final score. The same approach is applied to GPTEval3D~\citep{Gpt4v_evaluation}.
The experimental results are presented in Table~\ref{tab:pair-3DRewardDB}. Since the normal maps used here are estimated using Metric3D rather than ground truth, this may introduce visual bias, potentially affecting the performance of our scoring models in geometry-related dimensions. Despite this limitation, our scoring models still achieve outstanding performance on external datasets, demonstrating not only strong generalization capabilities but also remarkable robustness in handling diverse data sources and inherent estimation noise.

\subsection{Ablation Experiments}
\label{sec5.5:ablation_experiments}

We introduce a "two-stage" training strategy in Section~\ref{sec4:ScoreModel}. To validate its effectiveness and necessity, we carefully design and conduct ablation experiments on the \textit{3DGen-Score} model. Furthermore, considering the requirements of the evaluation task, we thoroughly assess two aspects: 
\textbf{1) alignment with human judgments} and \textbf{2) robustness to new models}. 
Specifically, we compute pairwise alignment on test pairs to measure the models' alignment with human judgments, and use Instant-Mesh pairs to assess their generalization performance. The experimental results, presented in Table~\ref{tab: ablation_training_alignment} and Table~\ref{tab: ablation_training_robustness}, demonstrate that the two-stage training strategy generally outperforms single-stage training. In particular, the second-stage training enhances the model's alignment with human judgments, while incorporating the first stage further strengthens its robustness to new models. 



\subsection{Discussion}
\label{sec5.6:discussion}
In summary, our two models each have distinct advantages. The 3DGen-Score model aligns well with human preferences while remaining computationally efficient and highly scalable. In contrast, the 3DGen-Eval model provides superior interpretability, matching or even surpassing GPT-4V in human preference alignment without incurring additional costs. Based on these strengths, we recommend the 3DGen-Eval model for detailed quality analysis of individual samples. For method-level comparisons, the 3DGen-Score model proves to be a suitable choice, particularly for the image-to-3D track. However, for those geometry-related dimensions of text-to-3D models, we recommend referring to the 3DGen-Eval model's results.
Additionally, we also provide a CLIP-based absolute scoring predictor in Appendix~\ref{secA5:appendix_3dgen_score_predictor}, which is more computationally efficient, allowing model ranking without the need for comparisons. 

\begin{figure*}[t]
    \centering
    \includegraphics[width=0.98\textwidth]{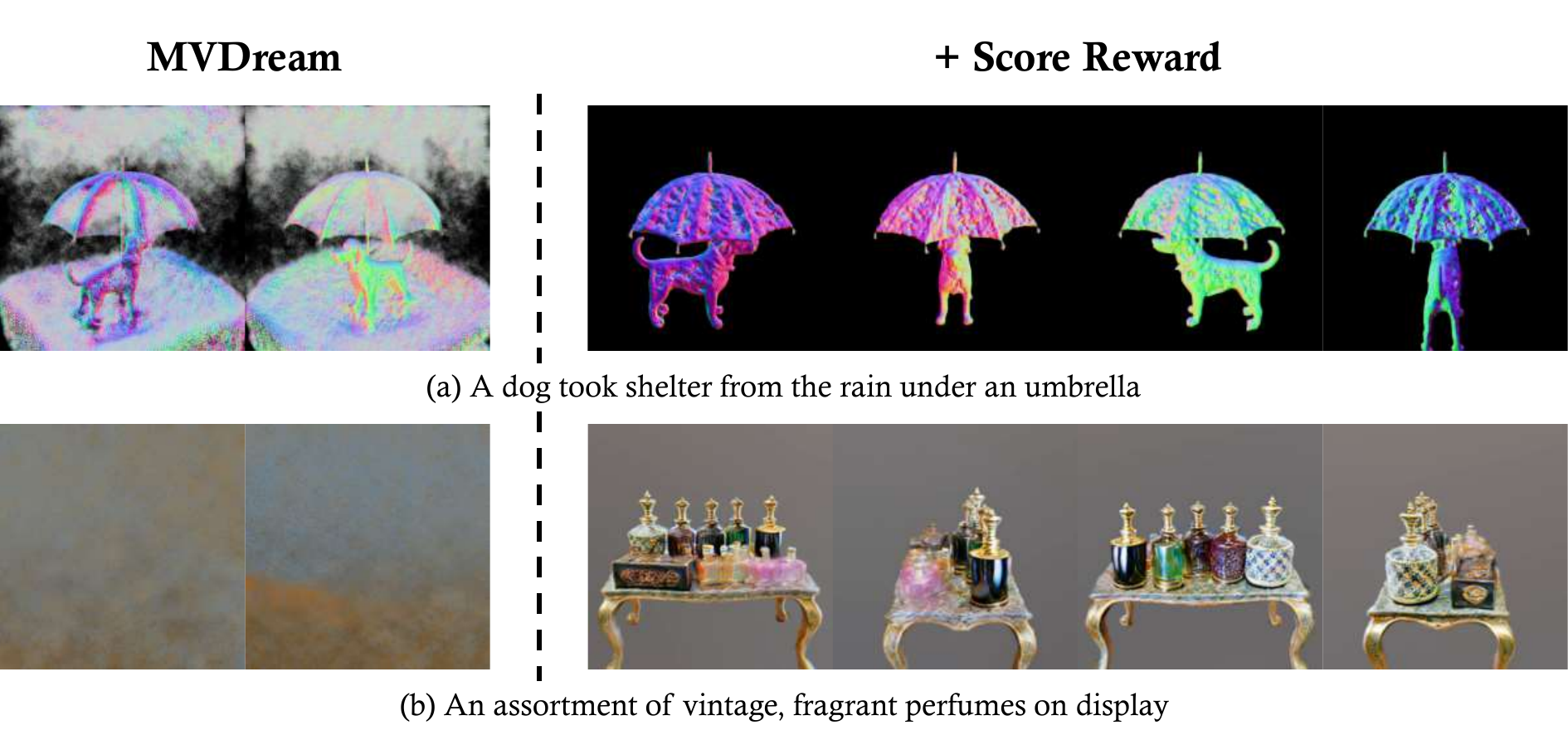}
    \includegraphics[width=0.98\textwidth]{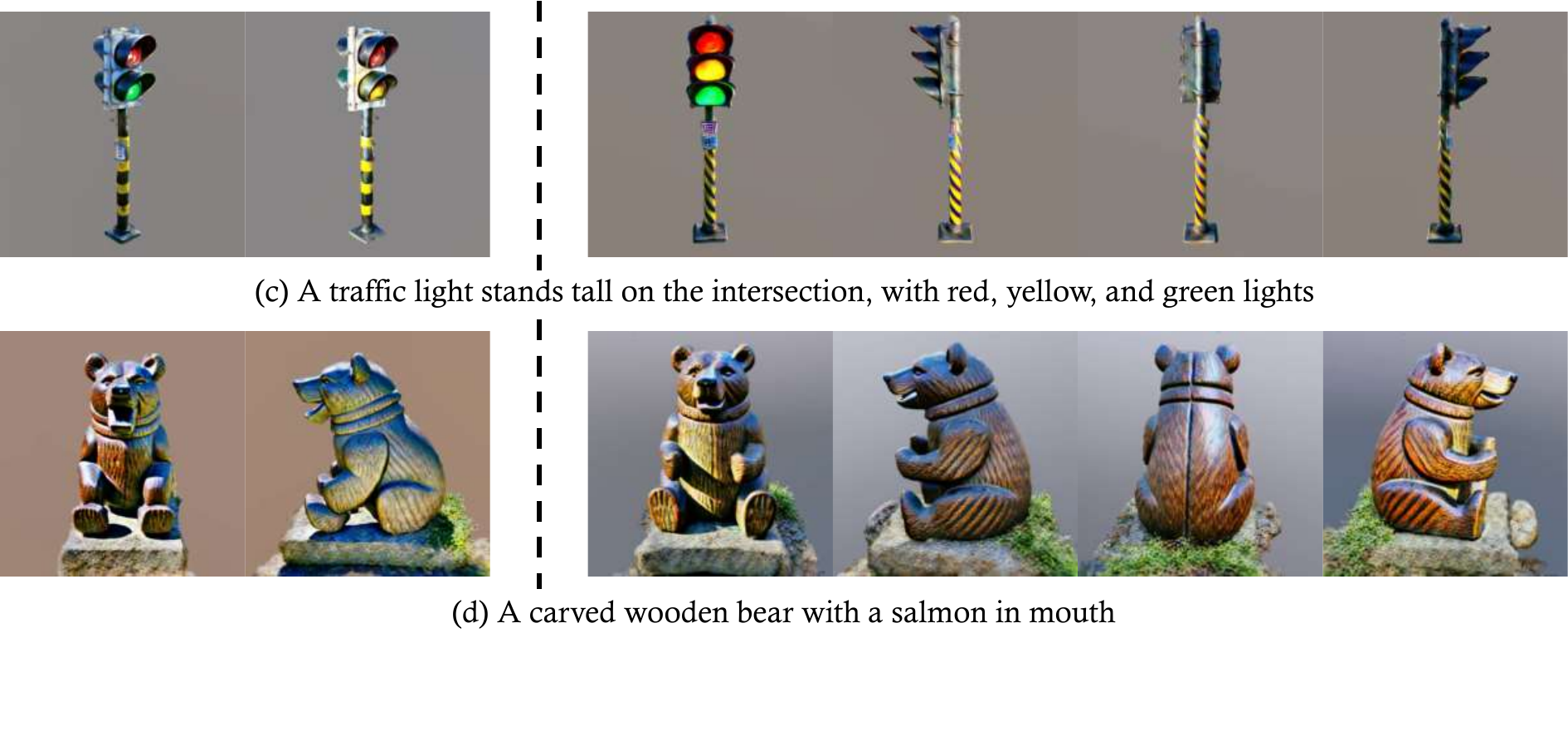}
    \vspace{-28pt}
    \caption{\textbf{Visual results of Reward Generation} We operate experiments over MVDream~\citep{Mvdream}, with original results on \textbf{Left} and RM-optimized on \textbf{Right}.}
    \label{fig:reward_model}
\end{figure*}

\section{Application}
\label{sec6:applications}

Using human preference data to establish leaderboards or train automatic scoring models is a straightforward application, but its potential extends beyond these use cases. One of the most critical applications is leveraging it as a reward signal to optimize generative models. In practice, Reinforcement Learning from Human Feedback (RLHF) has gained significant traction in the field of language and image generation~\citep{InstructGPT, HH-RLHF, Openai_RLHF, Webgpt, ImageReward}. Inspired by recent works such as~\citep{ImageReward, DreamReward}, we similarly employ \textit{3DGen-Score} as a Reward Model (RM) to simulate human feedback, allowing us to refine the behavior of generative models. To achieve this, we simplify the final objective function into the following formula:
\begin{align}
    \mathcal{L}_{Reward}(\theta) \approx \mathcal{L}_{SDS} - \lambda_{r}r(p, I_{normal}, I_{rgb}),
\end{align}
where $r$ represents the 3DGen-Score model, $\lambda_r$ is a hyperparameter, $p$ denotes the prompt, and $I_{normal}$ and $I_{rgb}$ are rendering images arranged in four views. For a detailed derivation, please refer to DreamReward~\citep{DreamReward}.

\begin{figure*}[ht]
    \centering
    \includegraphics[width=0.9\textwidth]{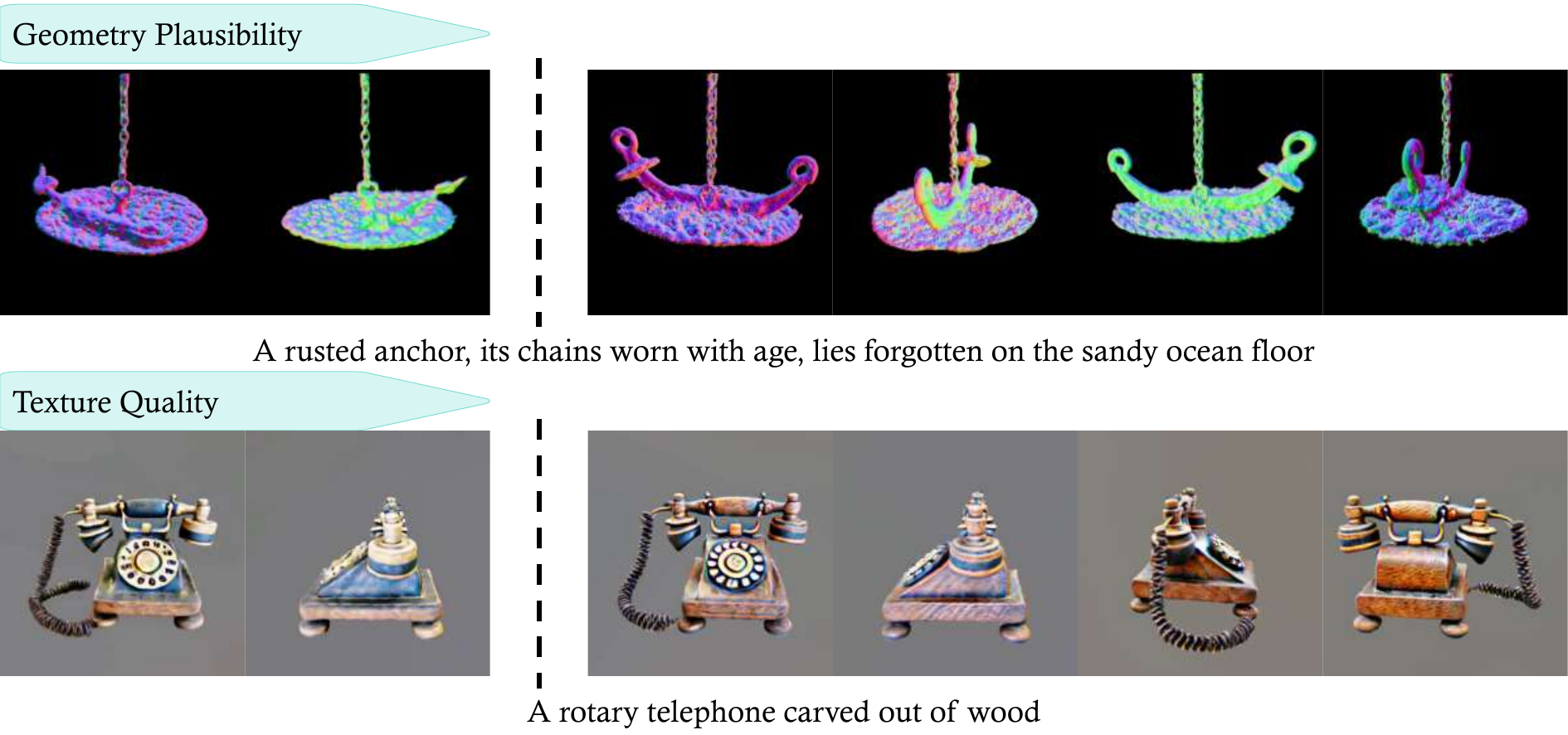}
    \includegraphics[width=0.9\textwidth]{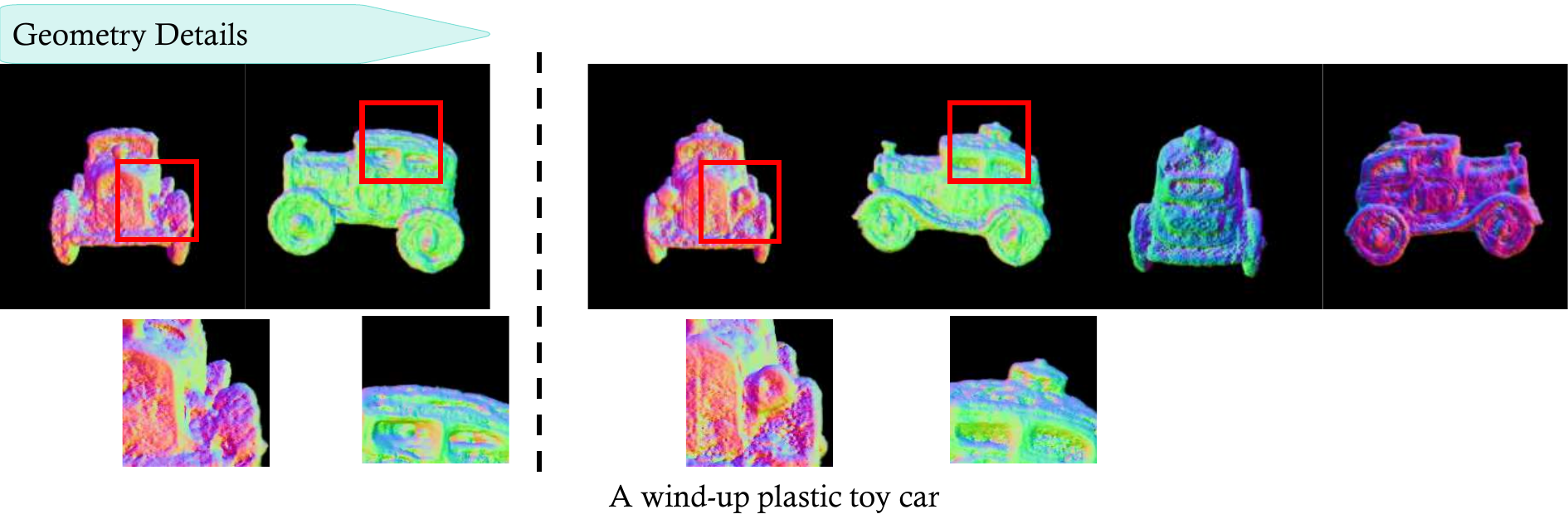}
    \includegraphics[width=0.9\textwidth]{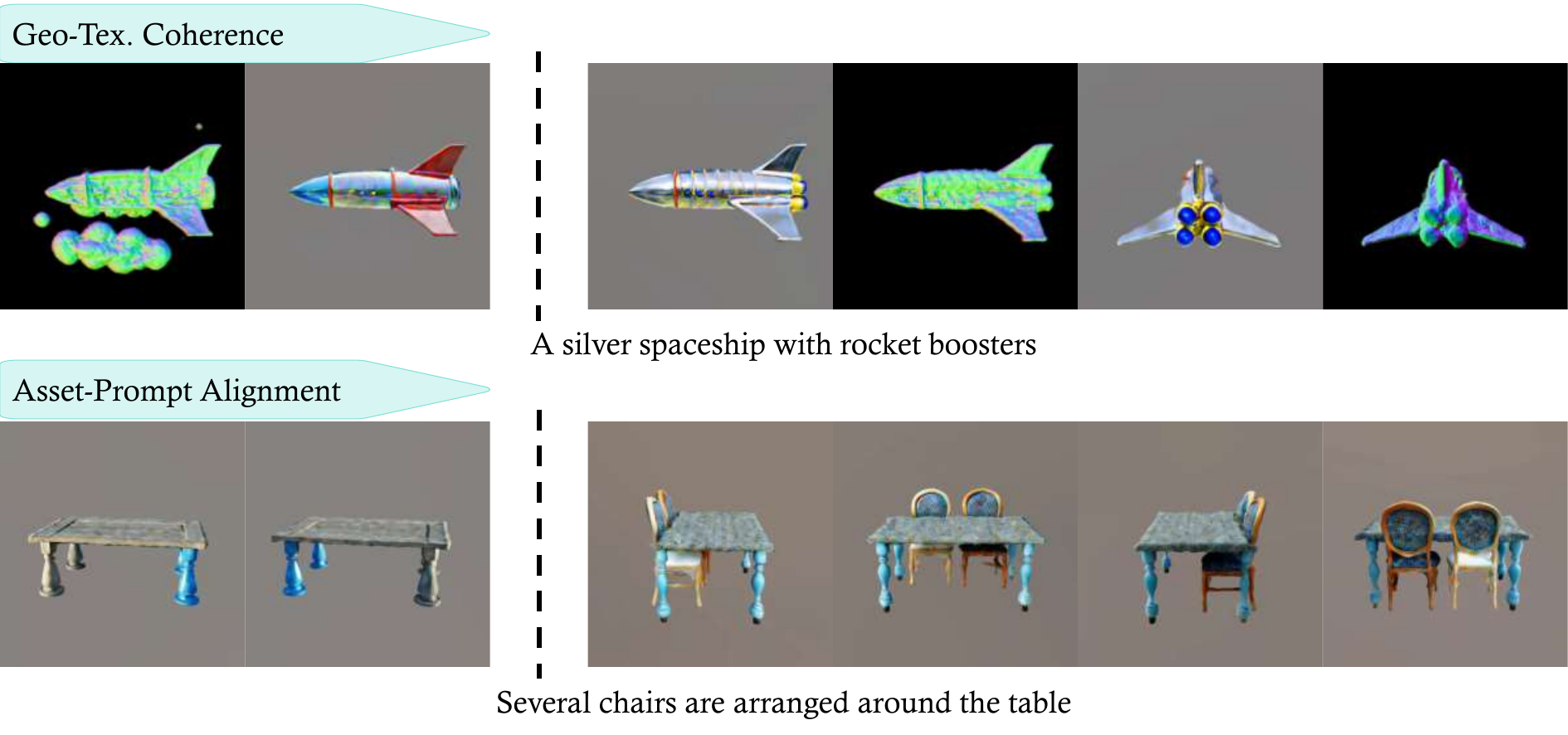}    
    \vspace{-5pt}
    \caption{\textbf{Visual results of Reward Generation for each dim}}
    \label{fig:reward_dim}
\end{figure*}

\begin{figure*}[t]
    \centering
    \includegraphics[width=\linewidth]{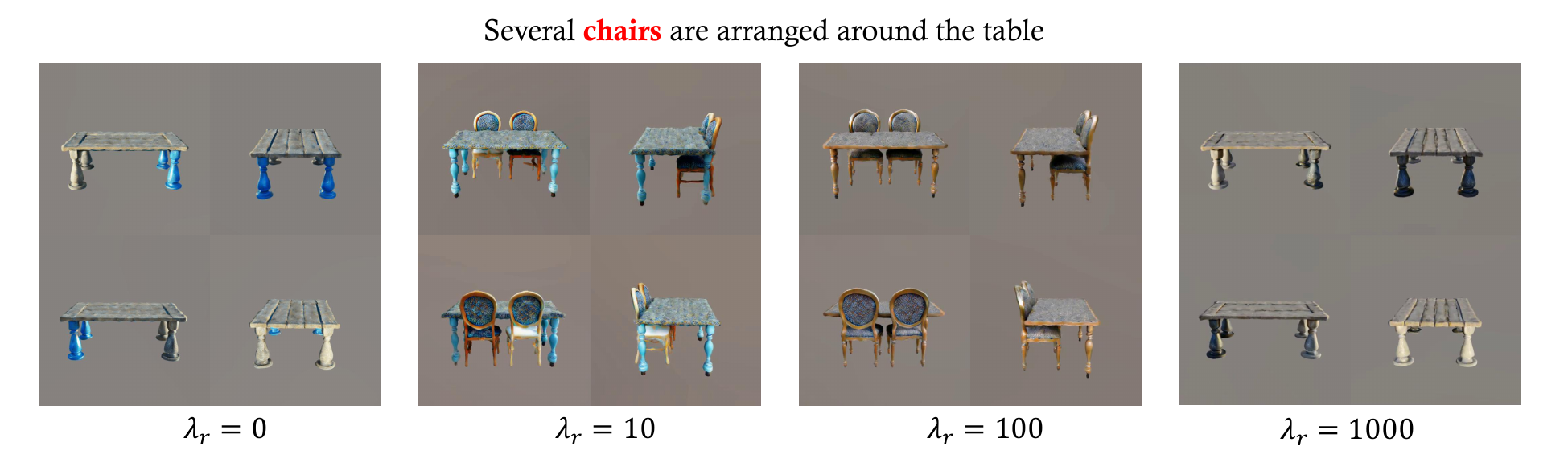}
    \vspace{-20px}
    \caption{\textbf{Visual Comparison on various $\lambda_r$}}
    \label{fig:reward_lambda}
\end{figure*}

\noindent{\textbf{Experiments.}}
In this paper, we conduct experiments on MVDream\citep{Mvdream}, which ranks as the top text-to-3D generative model according to our criteria. Moreover, MVDream generates multi-view images in parallel during optimization, which is particularly compatible with our objective function. 
In practice, we derive the final reward for each asset by averaging the scores across five evaluation dimensions.
As shown in Figure~\ref{fig:reward_model}, our approach effectively enhances the quality and coherence of 3D generative outputs across critical dimensions. For instance, examples (a) and (b) illustrate improvements in noise reduction, while examples (c) and (d) exhibit the enhanced capability of simulating 3D priors and refining textures. These results indicate that our \textit{3DGen-Score} model provides valuable insights for advancing generative performance.

\noindent{\textbf{Dimension Study.}}
We explored the impacts on each evaluation dimension through taking single dimension scores as the final reward. Visual results in Figure~\ref{fig:reward_dim} highlight the potential for targeted improvements in specific dimensions. Interestingly, we find that the effect of the reward score is not entirely isolated to each dimension. Instead, there is notable interaction across different dimensions, as optimizing for one dimension can also enhance or impact performance in others.

\noindent{\textbf{Ablation Study.}}
We investigate the impact of varying the value of $\lambda_r$. As shown in Figure~\ref{fig:reward_lambda}, the effects of $\lambda_r$ are not purely linear. Positive influence is observed only within an optimal range. When $\lambda_r$ is set too low, the improving potential remains underutilized. Conversely, if $\lambda_r$ is too large, the benefits diminish, and the generation process even begins to show signs of collapse.

\section{Conclusions}
\label{sec7:conclusions}
In this work, we present \textit{3DGen-Bench}, the first large-scale human preference dataset for 3D models. To construct this dataset, we developed \textit{3DGen-Arena}, a data platform that integrates 11,200 3D models generated by 19 different 3D generative models. By collecting votes from both public users and expert annotators, \textit{3DGen-Bench} eventually compiles over 68,000 preference votes and more than 56,000 score labels. Leveraging this data, we conduct a comprehensive evaluation and analysis of 3D generative models across five criteria, providing valuable insights for the advancement of 3D generation research. Additionally, we train two scoring models, \textit{3DGen-Score} and \textit{3DGen-Eval}, which serve as automated 3D evaluators, demonstrating superior alignment with human judgment and greater robustness to new models compared to existing metrics.

\noindent{\textbf{Limitations.}}
Our collection now includes 19 open-source generative models, but many works in this area still remain closed-source. Our goal is to involve state-of-the-art models and consistently update our leaderboard with the latest advancements. Additionally, due to the lack of a robust 3D embedding model, we opt for 2D CLIP embedding as an alternative. Developing more advanced 3D embedding techniques remains a priority to fully leverage the naive 3D data for evaluation.

\noindent{\textbf{Future Works.}}
Our 3DGen-Bench dataset innovatively realizes the subdivision of the annotation dimension, rescuing it from a holistic and rough score. While our annotations currently operate at the instance level, evolving evaluation standards may encourage finer-grained labeling, such as region-level annotations. Additionally, there remains a need for a more advanced 3DGen-Evaluator. Future iterations should explore more sophisticated model architectures or improved embedding techniques to enhance the evaluator's effectiveness and alignment.


\backmatter

\bmhead{Data Availability}
The dataset used in this study is publicly available in the 3DGen-Bench repository:~\url{https://huggingface.co/datasets/3DGen/3DGen-Bench}. Noting that the 3D assets adhere to the licensing terms of the respective methods from which they are derived, any use of the 3D data must comply with the original licensing terms and copyright policies.
\section*{Appendix}
\addcontentsline{toc}{section}{Appendix}

\appendix
\setcounter{table}{0}
\setcounter{figure}{0}
\renewcommand{\thetable}{R\arabic{table}}
\renewcommand\thefigure{S\arabic{figure}}

In Appendix, we provide the basic information and licensing terms of our dataset in Section~\ref{secA1:dataset_info}, along with additional details regarding its construction in Section~\ref{secA2:appendix_dataset}. 
We then present further instructions on the annotation platform and human leaderboard in Section~\ref{secA3:appendix_platform} and Section~\ref{secA4:appendix_leaderboard}, respectively.
Finally, we introduce a new scoring model in absolute form in Section~\ref{secA5:appendix_3dgen_score_predictor} and provide more experimental details in Section~\ref{secA6:appendix_3dgen-eval}.

\section{Dataset Information}
\label{secA1:dataset_info}
\subsection{Dataset Link and Documentation}
The dataset we established in this study, together with its metadata and license, is now publicly available in the 3DGen-Bench
repository:~\url{https://huggingface.co/datasets/3DGen/3DGen-Bench}. 
Specifically, we release the curated prompts, generated 3D models along with their renderings, and annotated human preferences.
Detailed documentation of its structure and usage can be found in the dataset card, and the Croissant metadata record can be viewed and downloaded by reviewers at~\url{https://huggingface.co/api/datasets/3DGen/3DGen-Bench/croissant}. 
Additionally, we host the 3DGen-Arena app on Huggingface Space~\url{https://huggingface.co/spaces/ZhangYuhan/3DGen-Arena}, enabling online voting among our diverse 3D assets.

\subsection{Licensing Terms}
Depending on the data source, the prompts and annotations are released under the ~\href{https://opensource.org/license/mit}{MIT License}, while the 3D assets follow the original licenses of the respective methods. Any use of the data must comply with the corresponding licensing terms and copyright policies.

\section{Additional Details on Dataset Construction}
\label{secA2:appendix_dataset}

\subsection{Prompt Generation}
\label{secA2.1:appendix_prompt}

We provide 510 text prompts in TXT format and 510 image prompts in RGBA format, generated respectively from ChatGPT and Stable Diffision. 
Specifically, for text prompts, we first identified 6 basic category domain: "\textbf{Vehicle}", "\textbf{Plant}", "\textbf{Animal}", "\textbf{Food}", "\textbf{Indoor Object}" and "\textbf{Outdoor Object}". Then, aiming to obtain as many specific categories as possible, we prompted ChatGPT with "\textit{We want to generate a prompt suite for text-to-3D generation evaluation. We hope the suite is comprehensive and challenging. The first step is to define category domain. We totally spit into 6 prime domain: {vehicle, animal, plant, food, indoor, outdoor}. Now we need you to list specific categories for each domain as many as possible}". 
Subsequently, equipped with diverse categories, we started to generate descriptive sentences. In order to generate prompts at different difficulty levels, we employed the following templates in turn.
\begin{itemize}
\item \textbf{Noun phrase}: \\
\{a/an\}\{attribution\}\{category\}\{with ...\}
\item \textbf{Simple sentence}: \\
\{count\}\{attribution\}\{category\}\{adv./verb-phrase /with ...\}
\item \textbf{Composite sentence}: \\
\{count\}\{attribution\}\{category\}\{verb-phrase\}\{count\}\{attribution\}\{category\}
\item \textbf{Other templates}: \\
allow ChatGPT to generate freely
\end{itemize} 
 where \{attribution\} and \{adv./verb-phrase/with ...\} are optional, \{count\} should smaller than six if defined explicitly, and prompts are encouraged to contain descriptive part to describe either geometric features or appearance features concretely.

\noindent{}Through heuristic generation in different domains, we obtained 600 raw prompts. We then do manual screening and targeted regeneration to ensure the balance of categories and difficulties. Eventually, we obtained 510 text prompts.

\begin{table*}[t]
    \centering
    \caption{\textbf{Leaderboard of Text-to-3D generative models.} Ranked by the average Elo score computed from comparison annotations.}
    \small
    \setlength{\tabcolsep}{9.3pt}
    \renewcommand{\arraystretch}{1.0}
    \begin{tabular}{c|cccccc}
        \toprule
         Method & Plausibility & Geo. Details & Tex. Quality & Geo-Tex. & Alignment & Average \\
         \midrule
         MvDream & 1107.90 & 1182.60 & 1229.61 & 1270.78 & 1095.39 & 1177.66 \\
         LucidDreamer & 1105.05 & 1133.85 & 1158.18 & 1163.19 & 1000.79 & 1122.21\\
         Magic3D & 1143.94 & 1100.30 & 1046.93 & 1042.77 & 1106.71 & 1088.93 \\
         GRM & 1032.35 & 1026.28 &  1100.23 & 1111.33 & 1058.49 & 1065.74 \\
         Dreamfusion & 1084.69 & 1029.05 &  1030.02 & 1009.73 & 1066.99 & 1044.10 \\
         Latent-NeRF & 986.09 & 938.78 &  1046.76 & 1130.76 & 983.47 & 1017.97 \\
         Shap-E & 889.87 & 890.49 &   849.57 & 764.86 & 981.75 & 875.71 \\
         SJC & 837.64 & 821.15 & 906.59 & 852.02 & 777.70 & 839.71 \\
         Point-E & 812.47 & 877.50 & 632.11 & 654.56 & 928.72 & 781.71 \\
         \bottomrule
    \end{tabular}
    \label{tab:leaderboard_t2s}
\end{table*}

\begin{table*}[t]
    \centering
    \caption{\textbf{Leaderboard of Image-to-3D generative models.} Ranked by the average Elo score computed from comparison annotations.}
    \small
    \setlength{\tabcolsep}{8.5pt}
    \renewcommand{\arraystretch}{1.0}
    \begin{tabular}{c|cccccc}
        \toprule
         Method & Plausibility & Geo. Details & Tex. Quality & Geo-Tex. & Alignment & Average \\
         \midrule
         Wonder3D & 1333.95 & 1308.03 & 1347.60 & 1321.71 & 1210.94 & 1304.05 \\
         OpenLRM & 1294.69 & 1330.10 & 1270.31 & 1260.36 & 1244.88 & 1279.67 \\
         Stable Zero123 & 1222.58 & 1157.20 & 1243.73 & 1225.31 & 1154.64 & 1200.69 \\
         Zero123-XL & 1092.45 & 1142.73 & 1140.74 & 1213.14 & 1055.74 & 1128.96 \\
         Magic123 & 1076.61 & 1038.36 & 1166.43 & 1194.64 & 1082.10 & 1111.23 \\
         LGM & 1058.78 & 1060.23 & 1066.24 & 1068.82 & 1049.68 & 1060.35 \\
         GRM & 1038.48 & 1067.10 & 1062.29 & 1025.53 & 1059.53 & 1042.99 \\
         SyncDreamer & 1079.61 & 1081.16 & 997.50 & 971.29 & 905.29 & 1006.97 \\
         Shap-E & 965.11 & 993.74 & 905.02 & 906.95 & 996.59 & 953.48 \\
         TriplaneGaussian & 840.28 & 819.21 &   880.53 & 893.25 & 836.79 & 854.01 \\
         Point-E & 740.57 & 769.02 & 707.68 & 704.43 & 916.69 & 767.68 \\
         EscherNet & 721.99 & 690.57 & 689.32 & 728.21 & 843.21 & 734.66 \\
         Free3D & 534.90 & 542.55 & 522.61 & 486.36 & 643.91 & 546.47 \\
         \bottomrule
    \end{tabular}
    \label{tab:leaderboard_i2s}
\end{table*}

\subsection{3D Generative Models}
\label{secA2.2:appendix_3dmodel}

In our benchmark, we include 9 text-to-3D generative models: Mvdream~\citep{Mvdream}, Lucid-dreamer~\citep{Luciddreamer}, Magic3D~\citep{Magic3D}, GRM~\citep{GRM}, Dreamfusion~\citep{Dreamfusion}, Latent-NeRF~\citep{Latent-NeRF}, Shap-E~\citep{Shap•E}, SJC~\citep{Score_Jacobian_Chaining}, and Point-E~\citep{Ponit-e}, and 13 image-to-3D generative models: Wonder3D~\citep{Wonder3D}, OpenLRM~\citep{OpenLRM}, Stable Zero123~\citep{Zero-1-to-3}, Zero-1-to-3 XL~\citep{Zero-1-to-3}, Magic123~\citep{Magic123}, LGM~\citep{LGM}, GRM~\citep{GRM}, SyncDreamer~\citep{SyncDreamer}, Shap-E~\citep{Shap•E}, Triplane-Gaussian~\citep{Triplane}, Point-E~\citep{Ponit-e}, Escher-Net~\citep{EscherNet}, and Free3D~\citep{Free3D}.

\subsection{Annotation Cost}
\label{secA2.3:appendix_cost}
Throughout the annotation process, we employed 47 expert annotators at a rate of \$6 per hour, spending approximately \$4,700 in total.
The cleaned results are published on Huggingface~\footnote{\label{fn:dataset}\url{https://huggingface.co/datasets/3DGen/3DGen-Bench}}.

\section{Additional Details about Annotation Platform}
\label{secA3:appendix_platform}

\subsection{3DGen-Arena}
\label{secA3.1:appendix_arena}
We run the 3DGen-Arena app on Huggingface Space~\footnote{\label{fn:arena}\url{https://huggingface.co/spaces/ZhangYuhan/3DGen-Arena}}. It supports online voting across our manifold 3D assets, offering three interaction modes: '\textbf{Anonymous battle models}', '\textbf{Named battle models}' and '\textbf{Single model chat}'. The first two modes enable model comparisons, differing in whether the voter is aware of the participant models' identities, while the third mode allows for individual model interactions.
The corresponding screenshots are listed in Figure~\ref{fig:arena_1},~\ref{fig:arena_2},~\ref{fig:arena_3}.

\begin{figure*}[t]
  \centering
   \includegraphics[width=\linewidth]{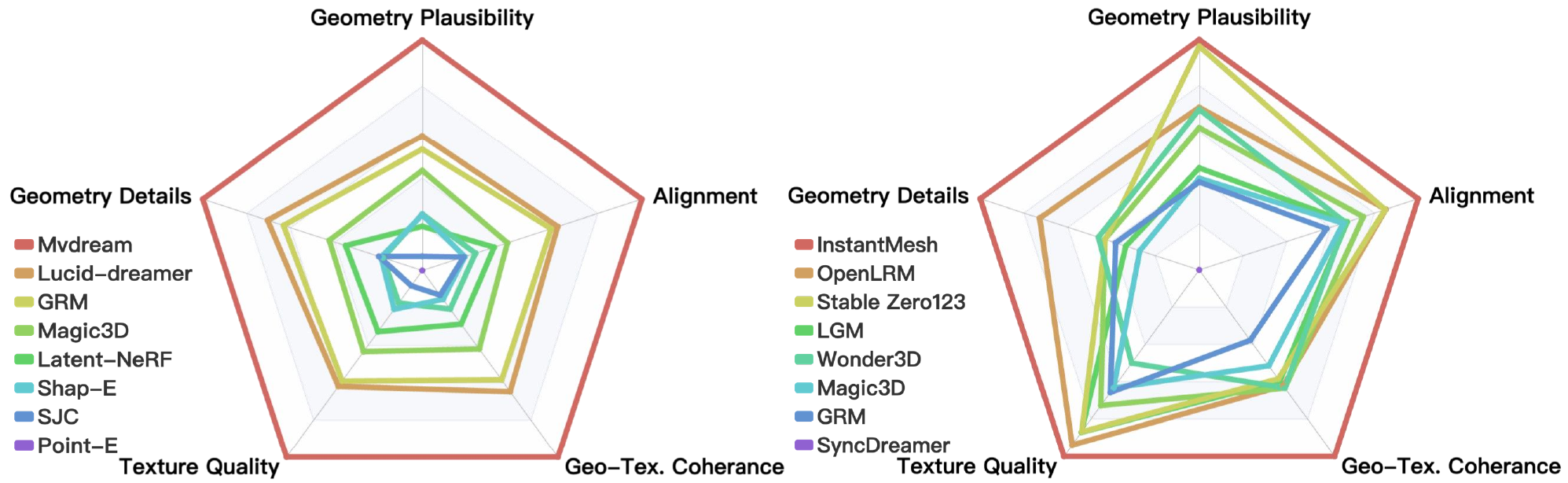}
   \vspace{-5px}
   \caption{\textbf{Elo scores of Generative models calculated by anonymous votes gathered from 3DGen-Arena.} \textbf{Left:} Text-to-3D generative models. \textbf{Right:} The top-9 Image-to-3D generative models. The legend in the down left corner of each image lists the ranking of each method based on the average across five dimensions of Elo scores, from top to bottom. The accurate scores can be checked in the leaderboard page running on huggingface~\textsuperscript{\ref{fn:arena}}.}
   \vspace{-10pt}
   \label{fig:arena_leaderboard}
\end{figure*}

\begin{figure*}[t]
    \centering
    \includegraphics[width=\linewidth]{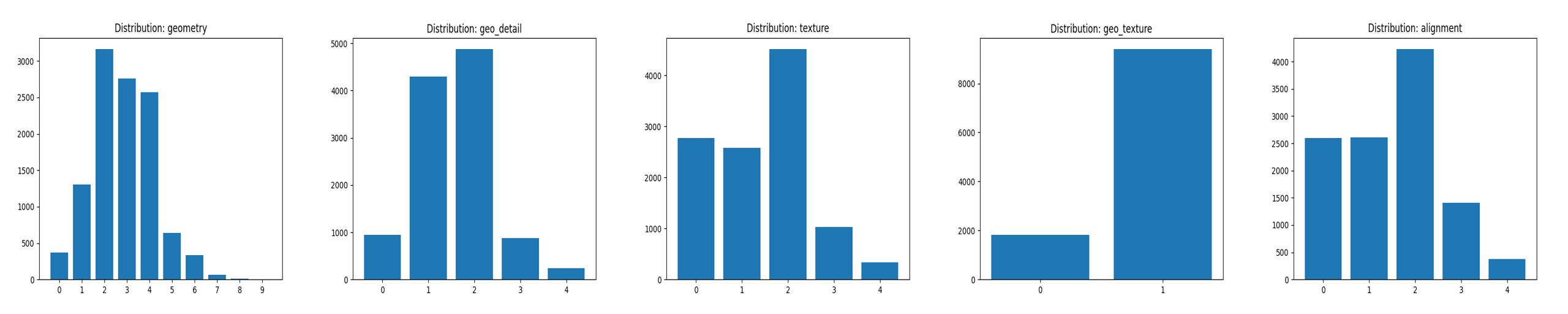}
    \vspace{-10px}
    \caption{\textbf{Distribution of \textit{Absolute Score} data}. It shows that high-quality 3D models are few in the collected dataset, annotated only over generated models.}
    \label{fig:score_anno_distribution}
\end{figure*}

\subsection{More Platforms}
\label{secA3.2:appendix_tag_platform}
\noindent{\textbf{Comparison Votes.}}
This process follows the same rules as the anonymous comparison in 3DGen-Arena, but given fixed battle pairs sampled by us to maintain balanced data distribution. In addition, to maintain the integrity and authenticity of the voting process, participants are required to vote using their real names, adding a layer of accountability to the evaluation.
The screenshot of voting platform is in Figure~\ref{fig:vote_platform}. 

\noindent{\textbf{Absolute Scores.}}
To enhance the credibility of the scoring results, we command annotators to rank all 3D models generated from the same prompt firstly, ensuring a consistent and fair assessment across similar generations. Subsequently, annotators proceed to label each model systematically, maintaining uniform criteria throughout the evaluation process.  This dual-step approach helps standardize scoring, reduce bias, and improve the reliability of our results.
The scoring platform screenshot is in Figure~\ref{fig:tag_platform}.

\section{Human Leaderboard}
\label{secA4:appendix_leaderboard}

\noindent{\textbf{Expert Annotations}}
We employ the Elo rating algorithm~\citep{elo} to compute the human leaderboard on our collected 13.8k comparison annotations. This approach enables a dynamic and adaptive ranking system, where models are evaluated based on their relative performance against each other. The accurate elo scores are listed in Table~\ref{tab:leaderboard_t2s} and Table~\ref{tab:leaderboard_i2s}.

\noindent{\textbf{Anonymous Votes}}
After hosting the 3DGen-Arena app for several months, we have collected 8,045 votes from anonymous visitors until now, with 6,351 votes for text-to-3D track and 1,694 for image-to-3D track. Based on these anonymous votes, we establish the leaderboard similarly by calculating Elo rankings, as shown in Figure~\ref{fig:arena_leaderboard}. 
Additionally, we also collected some votes from 'Named battle models' mode. And the detailed leaderboard that aggregates these two types of data is presented in huggingface~\textsuperscript{\ref{fn:arena}}.

\begin{table*}[ht]
    \centering
    \caption{\textbf{Ranking alignment with human judgment.} we compute the Kendall’s tau ranking correlation~\citep{Kendall}, where a higher value suggests a stronger alignment with human judgment.}
    \small
    \setlength{\tabcolsep}{4.5pt}
    \renewcommand{\arraystretch}{1.0}
    \begin{tabular}{c|ccccccc}
        \toprule
         Task & Method & Plausibility & Geo. Details & Tex. Quality & Geo-Tex. & Alignment & Average\\
         \midrule
         \multirow{2}{*}{Text-to-3D} & CLIP & 0.833 & 0.722 & 0.833 & 0.2778 & 0.889 & 0.711\\
         & 3DGen-SP(ours) & \textbf{0.833} & \textbf{0.833} & \textbf{0.889} & \textbf{0.389} & \textbf{0.889} & \textbf{0.767} \\
         \midrule
         \multirow{2}{*}{Image-to-3D} & CLIP & 0.154 & 0.00 & 0.256 & 0.462 & 0.180 & 0.210 \\
         & 3DGen-SP(ours) & \textbf{0.897} & \textbf{0.821} & \textbf{0.821} & \textbf{0.692} & \textbf{0.780} & \textbf{0.802} \\
         \bottomrule
    \end{tabular}
    \label{tab:rank_align_3dgen_sp}
\end{table*}

\begin{table*}[ht]
    \centering
    \caption{\textbf{Ranking alignment with additional Instant-Mesh.} we compute the Kendall’s tau ranking correlation~\citep{Kendall}, where a higher value suggests a stronger alignment with human judgment.}
    \small
    \setlength{\tabcolsep}{8.5pt}
    \renewcommand{\arraystretch}{1.0}
    \begin{tabular}{c|cccccc}
        \toprule
         Method & Plausibility & Geo. Details & Tex. Quality & Geo-Tex. & Alignment & Average\\
         \midrule
         CLIP & 0.209 & 0.077 & 0.297 & 0.473 & 0231 & 0.257 \\
         3DGen-SP(Ours) & \textbf{0.802} & \textbf{0.736} & \textbf{0.846} & \textbf{0.692} & \textbf{0.780} & \textbf{0.771} \\
         \bottomrule
    \end{tabular}
    \label{tab:rank_robust_3dgen_sp}
\end{table*}

\section{Absolute Scoring Predictor}
\label{secA5:appendix_3dgen_score_predictor}

\subsection{3DGen-Score Predictor}
\label{secA5.1:appendix_3dgen_sp_model}
We proposed the \textit{3DGen-Score} model in Section~\ref{sec4.1:3degn_score_model} to predict the win rate between two candidate models, learned from \textit{Comparison Votes} data,  while \textit{Absolute Scores} data remain under exploration. To fully leverage the potential of these data, we introduce \textit{3DGen-SP}, short for 3DGen-Score Predictor. Inspired by LAION Aesthetic Predictor V1~\citep{Laion-5B}, we extend vision encoders by incorporating five additional predictor heads to predict absolute scores for each dimension of the criterion.

\noindent{\textbf{Data Preparation}} 
As shown in Figure~\ref{fig:score_anno_distribution}, the distribution of absolute score annotations for the 11.2k generated 3D models is imbalanced, with high-quality samples being underrepresented in some dimensions.
To address this, we expand our score dataset by incorporating models from OmniObject3D~\citep{omniobject3d} and Cap3D~\citep{cap3d}. Specifically, we assign full scores for 6k scanned models from OmniObject3D and filter 3.4k high-quality models rated as "good" or higher by GPT-4 from Cap3D~\citep{bootstrap3d}. For each model, we treat its caption as the text prompt and its rendering image as the image prompt. As a result, we obtain approximately 19k high-quality models and 11.2k human-annotated models.

\noindent{\textbf{Training Strategy}}
We follow a similar model structure to \textit{3DGen-Score} model and employ a "two-stage" training strategy. In the first stage, called the Encoder-Adaptation stage, we minimize contrastive loss using only the 19k high-quality samples. In the subsequent stage, we minimize MSE loss with supervision from a total of 30k training samples. For implementation, we initialize the encoders from CLIP-ViT-H/14~\citep{Clip} and initialize the learnable predictor heads using Xavier Uniform. And we only unfreeze selected final layers of vision encoders during training, specifically the last 4 layers in this paper. The model is trained with a batch size of 32 and a learning rate of 3e-6, using the Adam optimizer to optimize the training process.

\subsection{Evaluation}
\label{secA5.2:appendix_3dgen_sp_eval}

\noindent{\textbf{Alignment with Human}} 
As an evaluation task, we prioritize consistency across distributions rather than precise score values of individual models. Therefore, we assess the performance of the scoring model by computing ranking consistency with human annotations. Specifically, we rank models by calculating the average score over test dataset predicted by CLIP and \textit{3DGen-SP}, respectively, and calculate the Kendall’s tau ranking correlation~\citep{Kendall} to assess alignment with human annotations. As shown in Table~\ref{tab:rank_align_3dgen_sp},
our model consistently outperforms CLIP in all tested scenarios.

\noindent{\textbf{Robustness to New Model}}
In a similar setting, we further conduct experiments on Instant-Mesh pairs to evaluate the generalization capability of \textit{3DGen-SP} model. As shown in Table~\ref{tab:rank_robust_3dgen_sp},  our \textit{3DGen-SP} model demonstrates strong robustness against novel models, consistently outperforming CLIP across all dimensions. These results further validate the effectiveness of our training strategy.


\section{Additional Details about 3DGen-Evaluator}
\label{secA6:appendix_3dgen-eval}

\begin{figure*}[t]
    \centering
    \includegraphics[width=\textwidth]{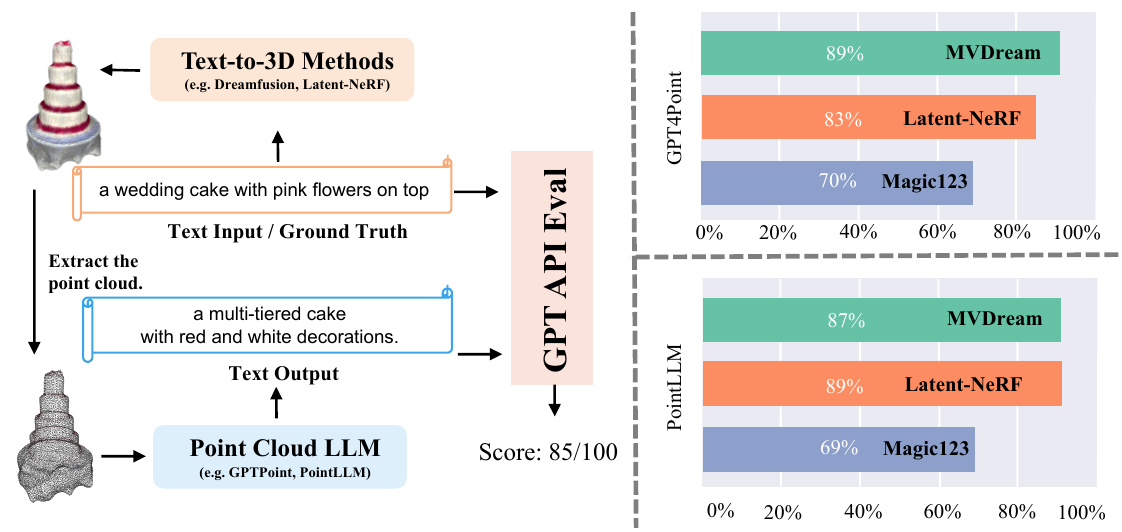}
    \vspace{-10pt}
    \caption{\textbf{Evaluation with point-based MLLMs} \textbf{Left}: An overview of the evaluation pipeline. \textbf{Right}: Model rankings on the test data.}
    \label{fig:gpt4point}
\end{figure*}

\begin{table*}[t]
    \caption{\textbf{Pairwise rating alignment compared with $T^3$Bench.} We assess the average probability that our model's decisions align with human judgments for each comparison.}
    \label{pair-t3bench}
    \centering
     \small
    \setlength{\tabcolsep}{10pt}
    \renewcommand{\arraystretch}{1.0}
    \scalebox{0.95}{
    \begin{tabular}{lcccccc}
    \toprule
       Methods & Plausibility & Geo. Details & Tex. Quality & Geo-Tex. & Alignment & Average\\     
       \midrule
        $T^3$Bench & 0.686 & 0.675 & 0.693 & 0.665 & 0.608 & 0.665 \\ 
        3DGen-Score(Ours) & \textbf{0.789} & \textbf{0.714} & \textbf{0.742} & \textbf{0.716} & \textbf{0.747} & \textbf{0.742} \\ 
    \bottomrule
    \end{tabular}%
    }
\end{table*}

\subsection{3D Prior Exploration}
\label{secA6.1:appendix_exp_gpt4point}
We employ 3D point-based MLLMs~\citep{gpt4point, pointllm} to explore the feasibility of assessing quality directly using 3D representations. The primary process involves evaluating the alignment between ground-truth prompts and generated captions, as shown in Figure~\ref{fig:gpt4point}. First, we extract a colored point cloud from the generated 3D models, typically represented in formats such as Mesh, NeRF, or Gaussian, and organize them into the format (N\_points, 6), consisting of xyz coordinates and RGB colors. Next, we prompt the 3D MLLMs with "\textit{What is the object?}" to generate corresponding captions. These text pairs are then submitted to ChatGPT for human-like assessment. Finally, we sample 100 battle pairs from the test dataset, and the ranking results are presented in Figure~\ref{fig:gpt4point}.

\subsection{Comparison with $T^3$Bench}
\label{secA6.2:appendix_exp_t3bench}
Since $T^3$Bench~\citep{T3_Bench} has stricter requirements on the input format, converting the format for some models in the 3DGS representation proved challenging. So we only conduct experiments on five models that are supported: Dreamfusion~\citep{Dreamfusion}, Latent-NeRF~\citep{Latent-NeRF}, Mvdream~\citep{Mvdream}, SJC~\citep{Score_Jacobian_Chaining} and Magic3D~\citep{Magic3D}. We compare pairwise alignment with human on our test dataset, as $T^3$Bench has not provided or released validation data. The results, shown in Table~\ref{pair-t3bench}, reveal that our scoring model consistently outperforms T3Bench across all dimensions, demonstrating its impressive ability to simulate human preferences.

\subsection{Prompt Design for 3DGen-Eval}
\label{secA6.3:appendix_3dgen_eval_prompt}

\noindent{\textbf{Prompts for GPT-4V}}
To generate pseudo labels for training the \textit{3DGen-Eval} model, we prompt GPT-4V to produce detailed descriptions for each 3D model and well-reasoned explanations for each dimension to justify the assigned scores. The template used for \textit{Absolute Scores} and \textit{Comparison Votes} are shown in Figure~\ref{fig:prompt_gpt4v_score} and Figure~\ref{fig:prompt_gpt4v_compair}, respectively.

\noindent{\textbf{Templates for Instruct Tuning}}
We then structure the responses generated from GPT-4V into templates compatible with LLaVA~\citep{LLaVa} for better understanding and processing. Example diagrams for \textit{Abosulte Score} and \textit{Comparison Votes} are shown in Figure~\ref{fig:prompt_instruct_score} and Figure~\ref{fig:prompt_instruct_compair}, respectively.

\clearpage

\begin{figure}[]
    \vspace{-10px}
    \centering
    \includegraphics[width=\linewidth]{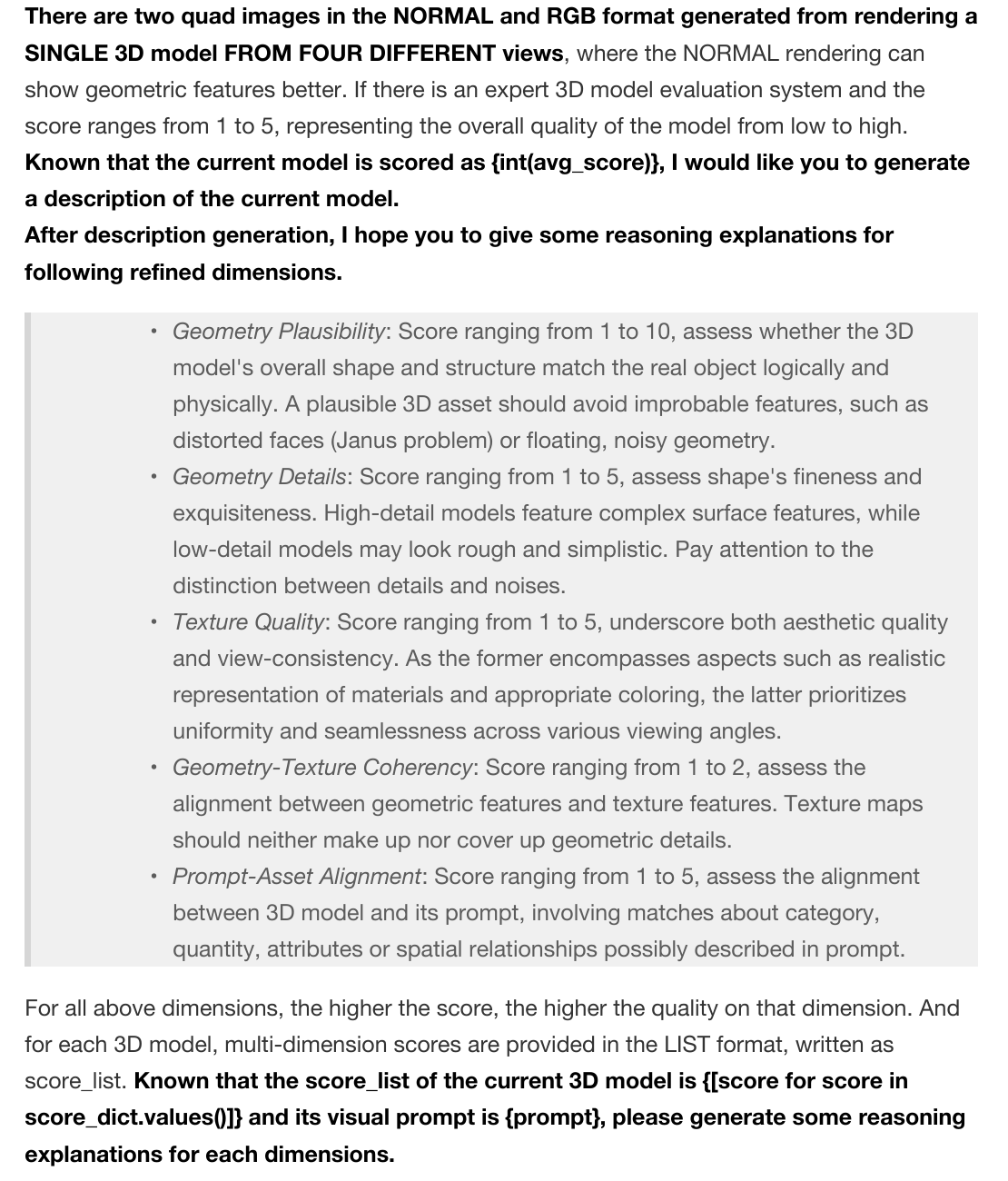}
    \vspace{-10px}
    \caption{\textbf{Prompts for GPT-4V to generate descriptions and explanations of \textit{Absolute score} for single 3D model}, where the \{score\_dict\} is score annotations of 5 dimensions, and the \{avg\_score\} is averaged from \{score\_dict\}.}
    \label{fig:prompt_gpt4v_score}
\end{figure}

\begin{figure}[]
    \centering
    \includegraphics[width=\linewidth]{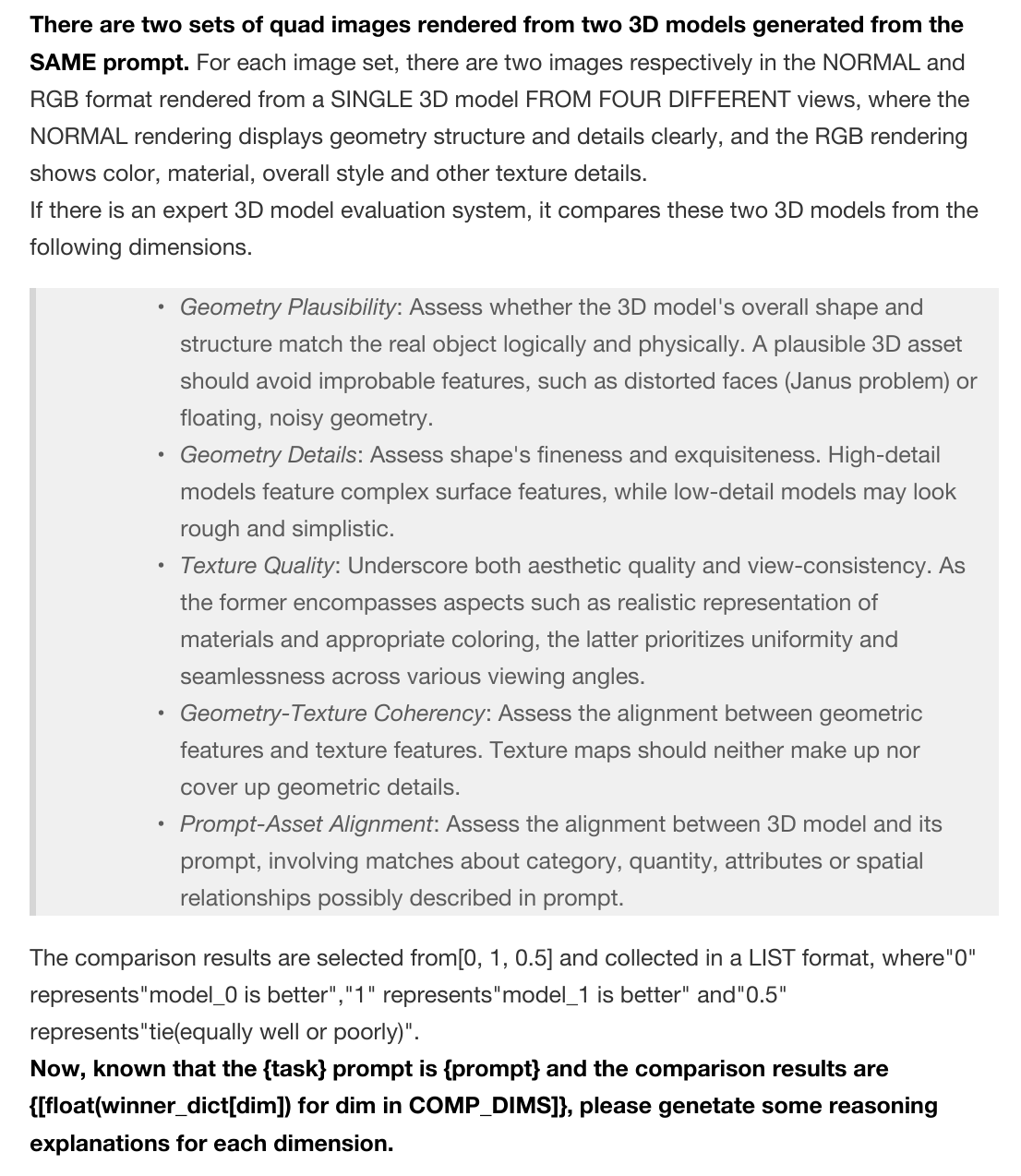}
    \vspace{-10px}
    \caption{\textbf{Prompts for GPT-4V to generate explanations of \textit{Pairwise comparison} for two 3D models generated from the same prompt}, where the \{prompt\} is the corresponding prompt input and the \{winner\_dict\} represents comparison annotations of five dimensions.}
    \label{fig:prompt_gpt4v_compair}
\end{figure}

\begin{figure}[h]
    \vspace{-10px}
    \centering
    \includegraphics[width=0.95\linewidth]{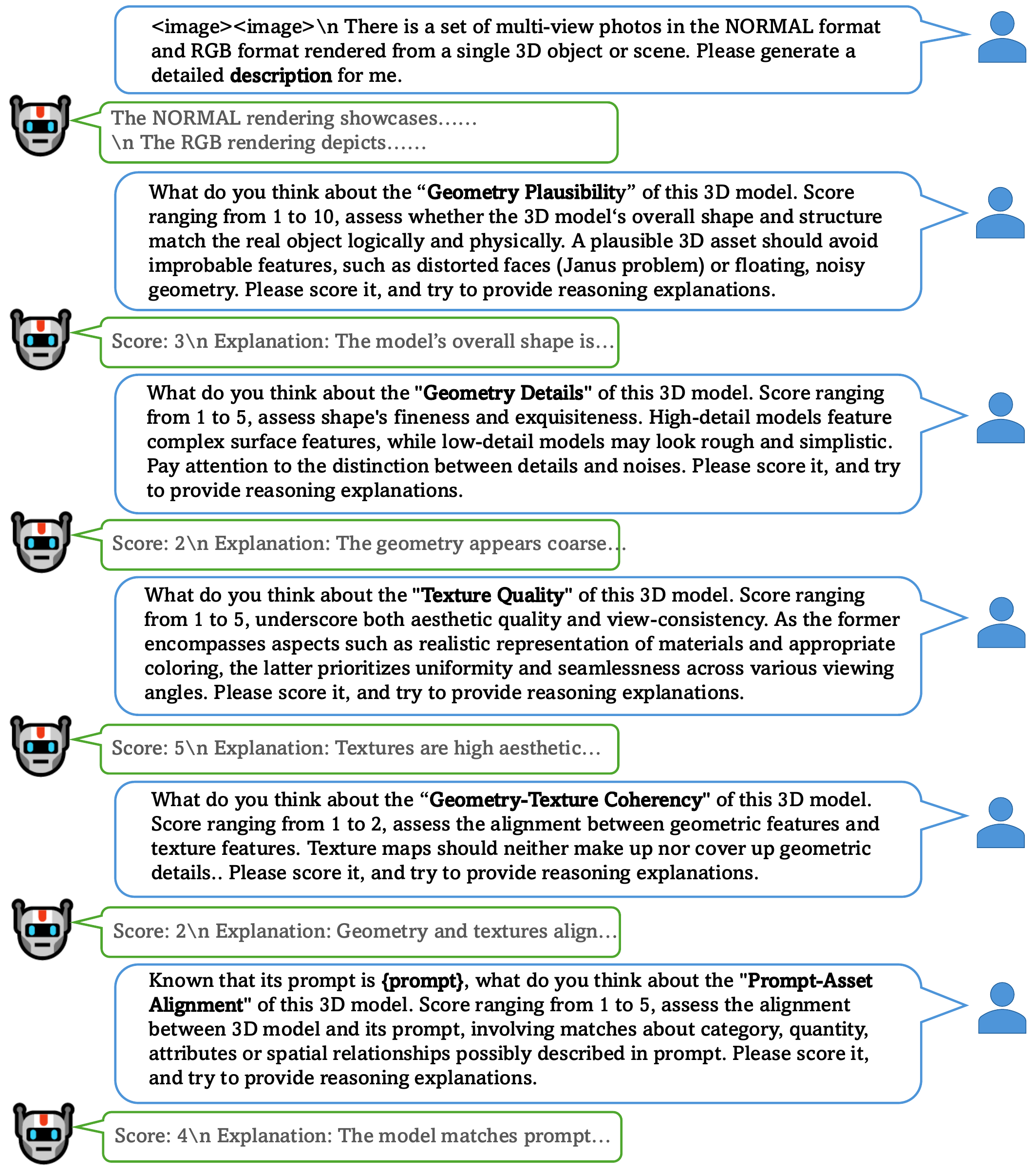}
    \caption{\textbf{Prompts for \textit{3DGen-Eval} model's Instruct-Tuning stage(a) trained with \textit{Absolute scores}}, where the \{score\} is supervised by the human annotations, and the \{explanation\} is supervised by GPT-4V's responses.}
    \vspace{-10px}
    \label{fig:prompt_instruct_score}
\end{figure}

\begin{figure}[b]
    \centering
    \includegraphics[width=0.95\linewidth]{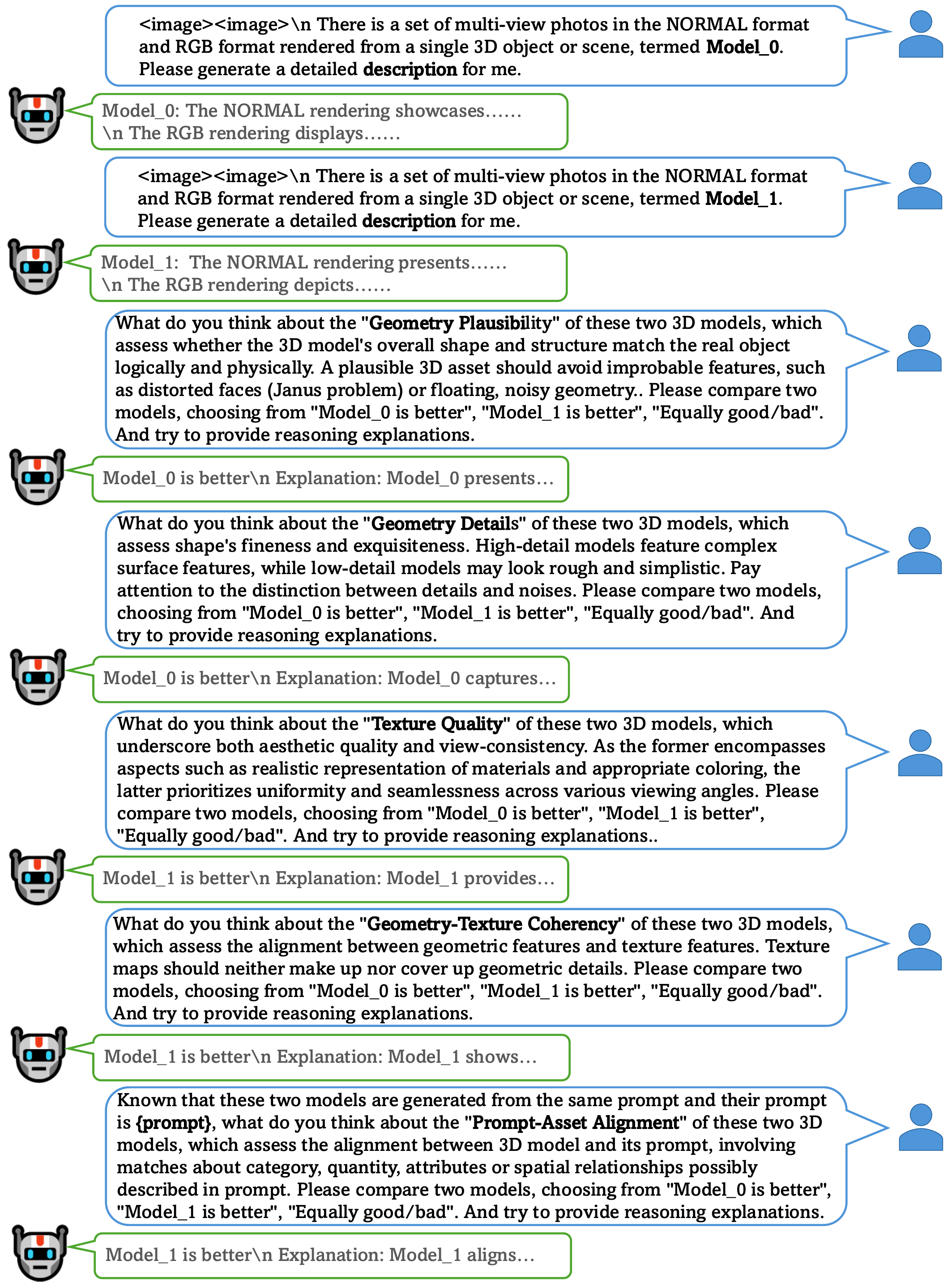}
    \caption{\textbf{Prompts for \textit{3DGen-Eval} model's Instruct-Tuning stage(b) trained with \textit{Comparison votes}}.}
    \label{fig:prompt_instruct_compair}
\end{figure}

\begin{figure*}[t]
    \centering
    \includegraphics[width=0.85\textwidth]{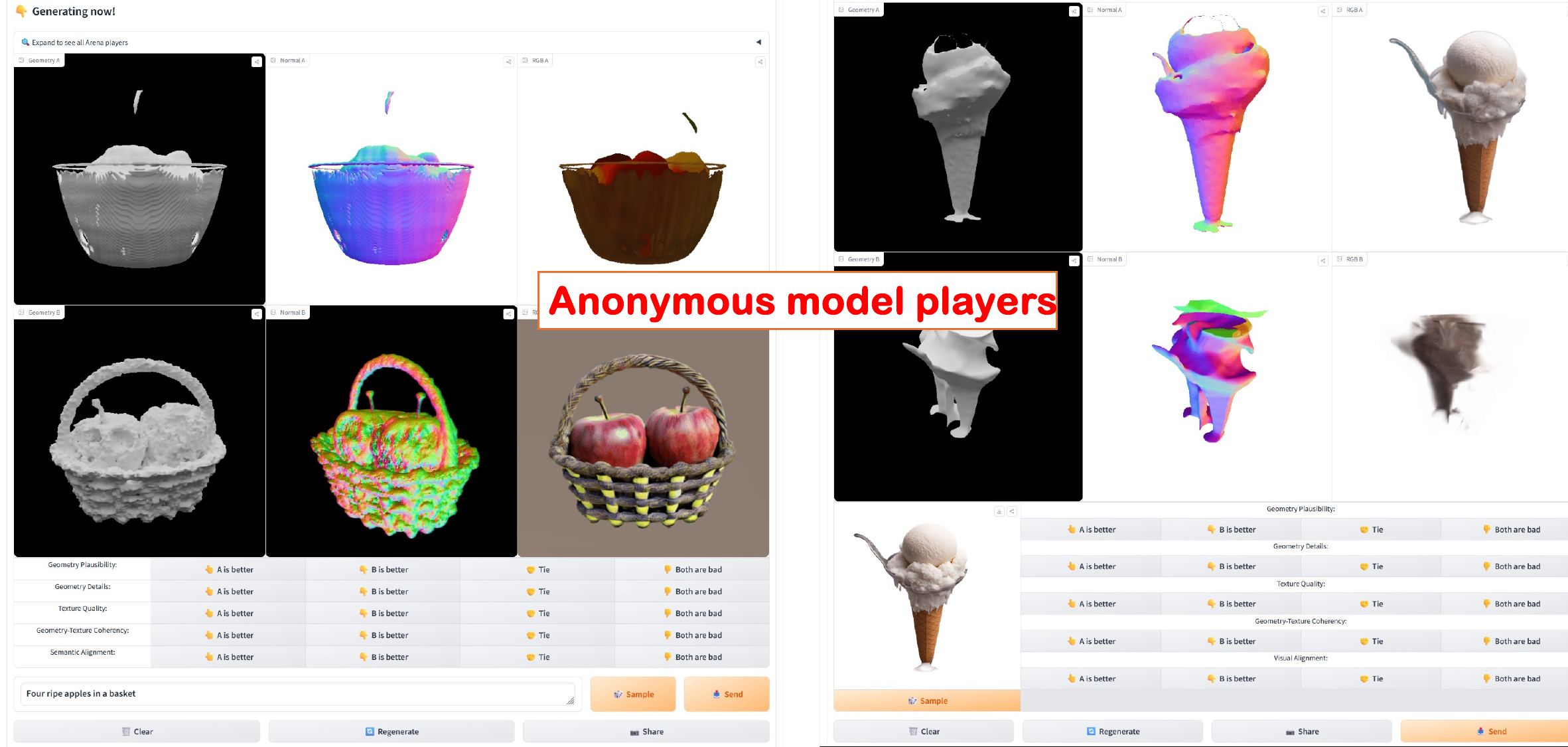}
    \caption{3DGen-Arena Page1: Anonymous battle models}
    \label{fig:arena_1}
    
    \includegraphics[width=0.85\textwidth]{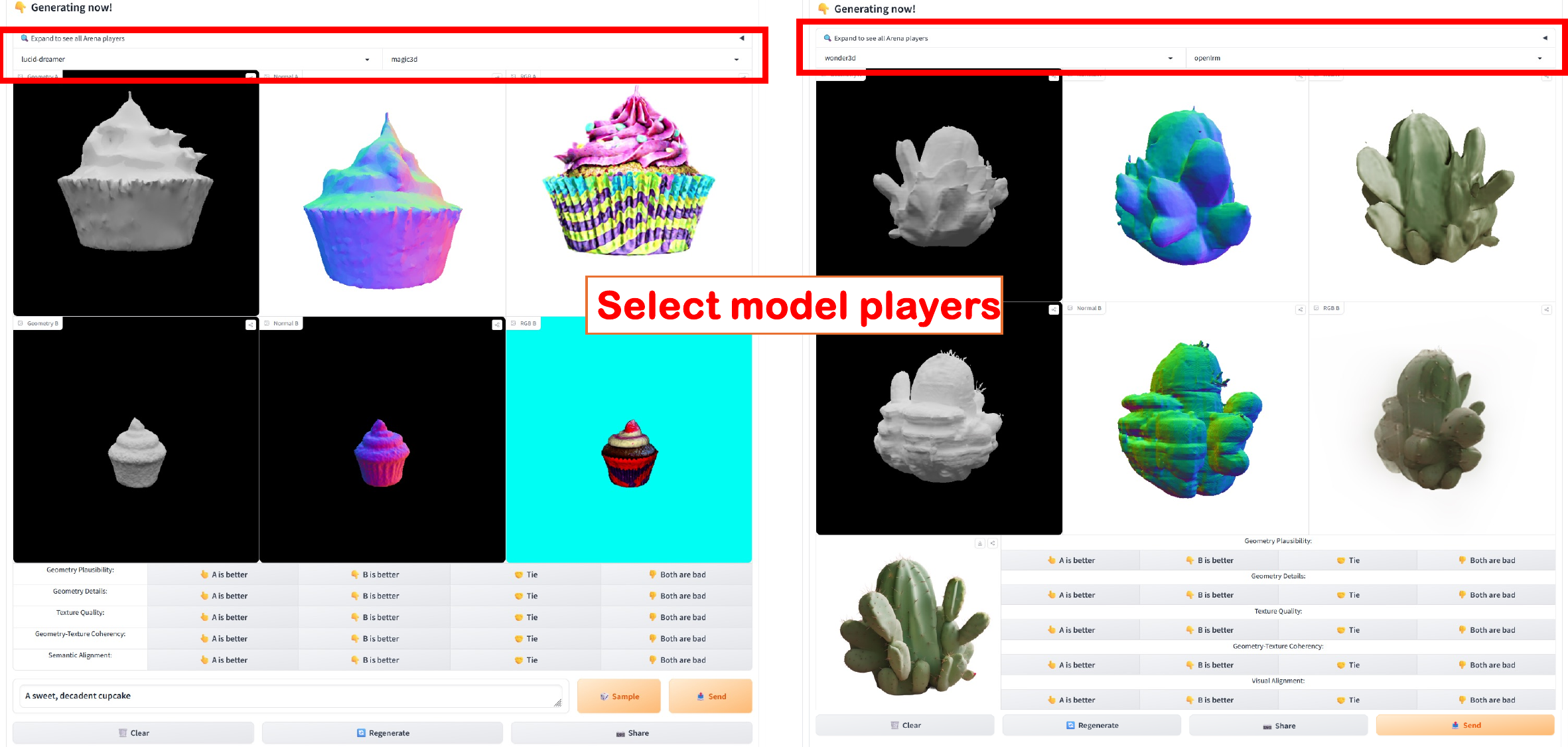}
    \caption{3DGen-Arena Page2: Named battle models}
    \label{fig:arena_2}
    
    \includegraphics[width=0.85\textwidth]{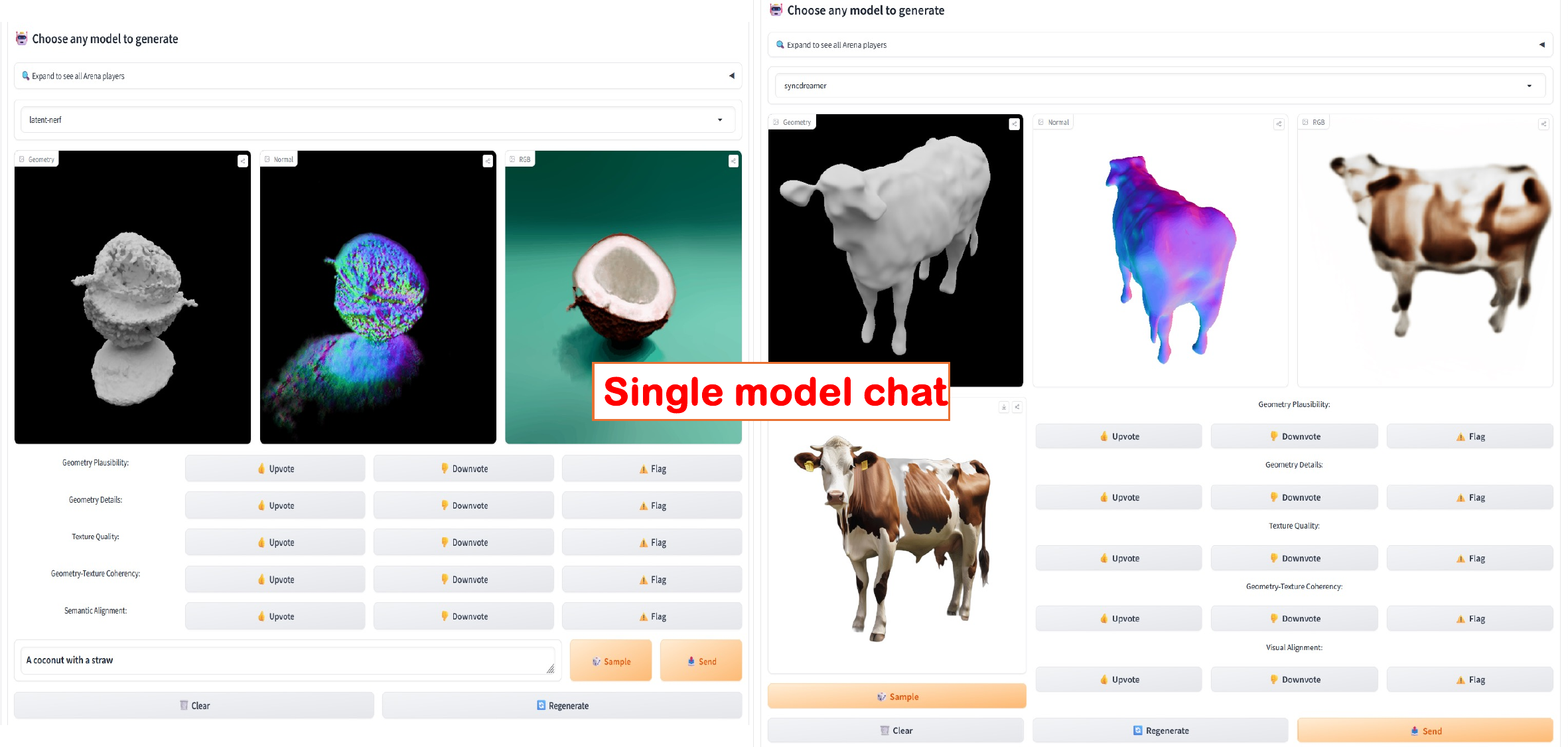}
    \caption{3DGen-Arena Page3: Single model chat}
    \label{fig:arena_3}
\end{figure*}

\begin{figure*}[]
    \centering
    \includegraphics[width=0.85\textwidth]{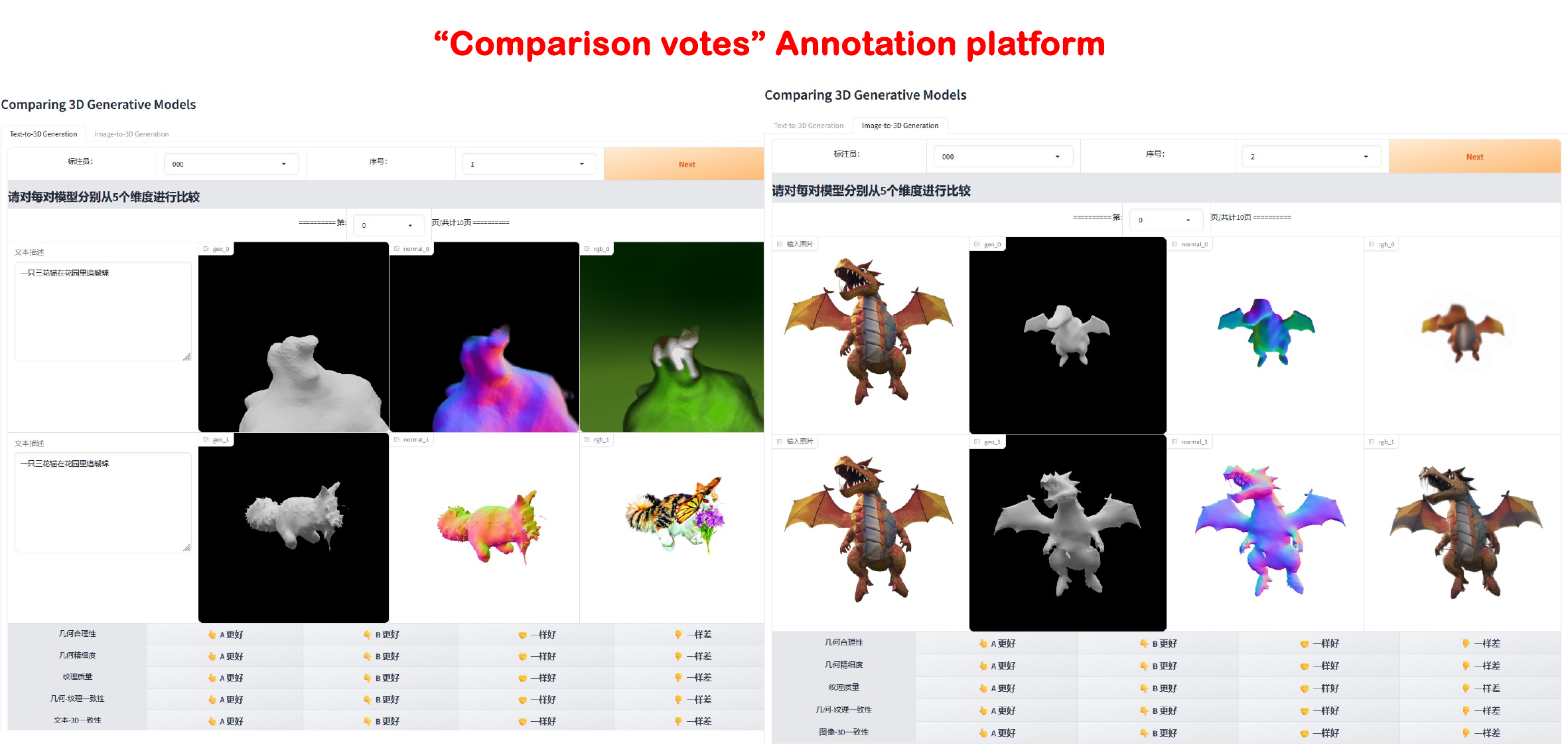}
    \caption{Screenshots of annotation platform for \textbf{Comparison Votes}}
    \label{fig:vote_platform}
    
    \includegraphics[width=0.855\textwidth]{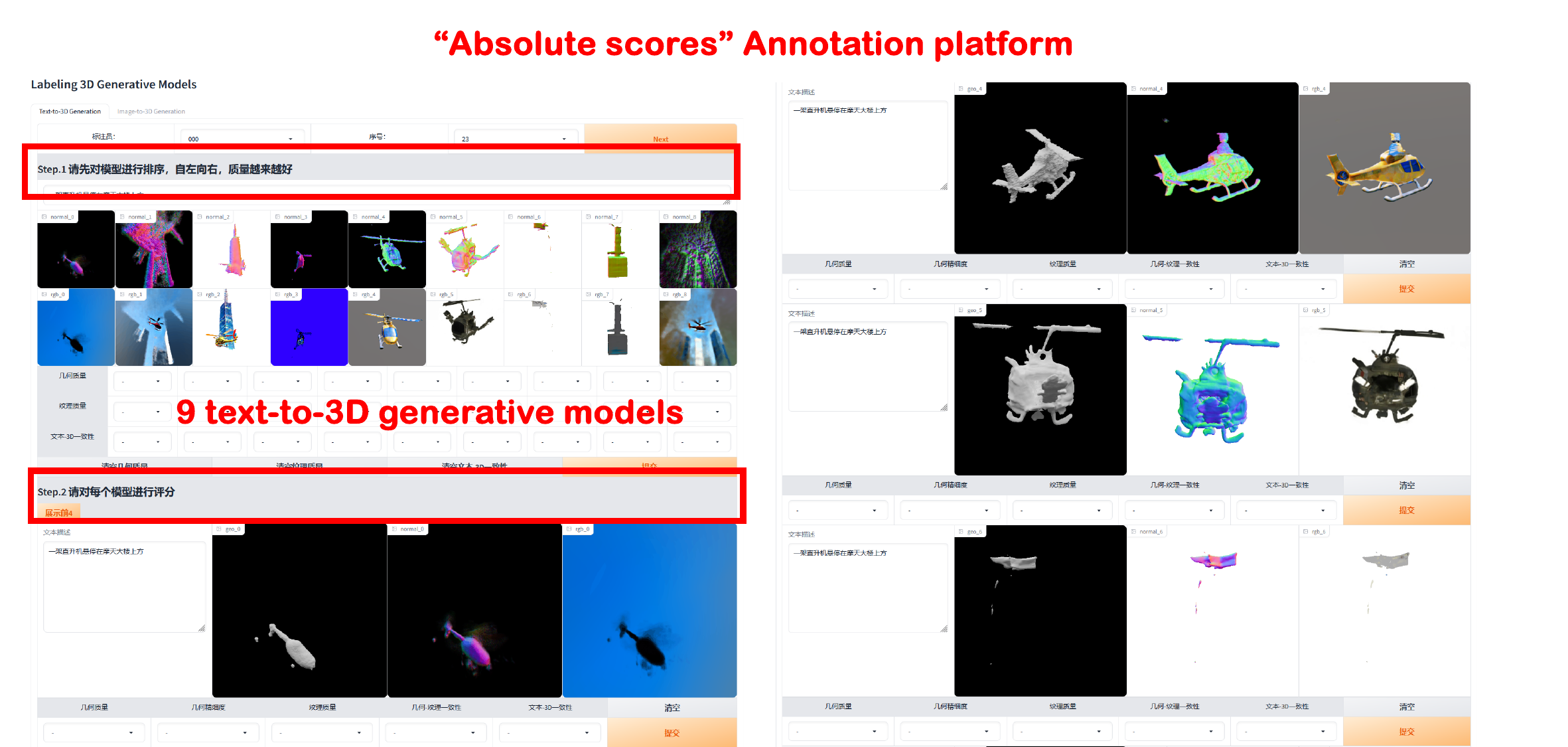}
    \includegraphics[width=0.85\textwidth]{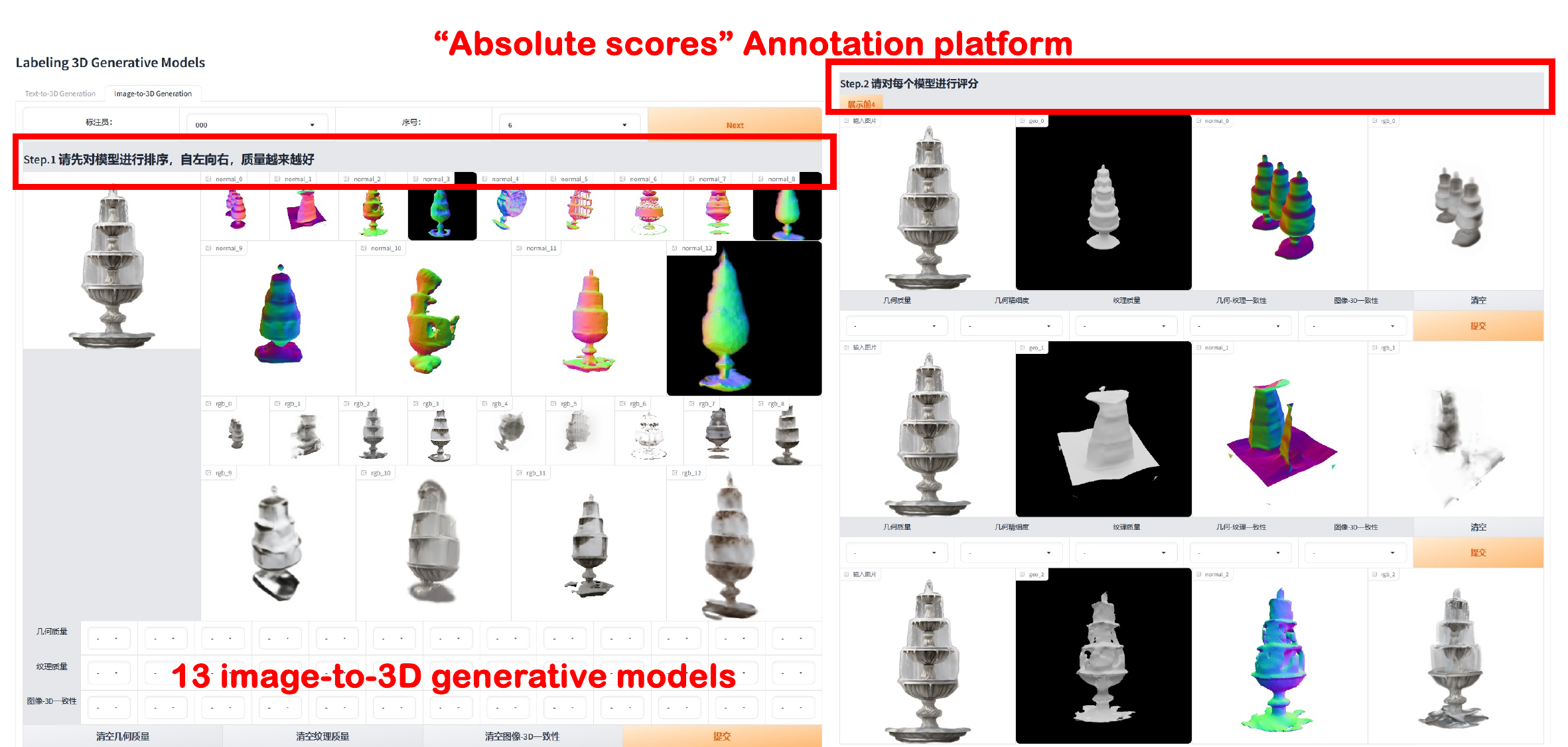}
    \caption{Screenshots of annotation platform for \textbf{Absolute Scores}}
    \label{fig:tag_platform}
\end{figure*}





\clearpage

\bibliography{sn-bibliography}

\begin{thebibliography}{}
\renewcommand{\doi}[1]{\url{https://doi.org/#1}}
\bibcommenthead

\bibitem [\protect \citeauthoryear {%
Bai%
\ \protect \BOthers {.}}{%
Bai%
\ \protect \BOthers {.}}{%
{\protect \APACyear {2022}}%
}]{%
HH-RLHF}
\APACinsertmetastar {%
HH-RLHF}%
\begin{APACrefauthors}%
Bai, Y.%
, Jones, A.%
, Ndousse, K.%
, Askell, A.%
, Chen, A.%
, DasSarma, N.%
\BDBL {}others%
\end{APACrefauthors}%
\unskip\
\newblock
\APACrefYearMonthDay{2022}{}{}.
\newblock
{\BBOQ}\APACrefatitle {Training a helpful and harmless assistant with reinforcement learning from human feedback} {Training a helpful and harmless assistant with reinforcement learning from human feedback}.{\BBCQ}
\newblock
\APACjournalVolNumPages{arXiv preprint arXiv:2204.05862}{}{}{,}
\newblock

\newblock

\PrintBackRefs{\CurrentBib}

\bibitem [\protect \citeauthoryear {%
L.~Chen%
\ \protect \BOthers {.}}{%
L.~Chen%
\ \protect \BOthers {.}}{%
{\protect \APACyear {2023}}%
}]{%
ShareGPT4v}
\APACinsertmetastar {%
ShareGPT4v}%
\begin{APACrefauthors}%
Chen, L.%
, Li, J.%
, Dong, X.%
, Zhang, P.%
, He, C.%
, Wang, J.%
\BDBL {}Lin, D.%
\end{APACrefauthors}%
\unskip\
\newblock
\APACrefYearMonthDay{2023}{}{}.
\newblock
{\BBOQ}\APACrefatitle {Sharegpt4v: Improving large multi-modal models with better captions} {Sharegpt4v: Improving large multi-modal models with better captions}.{\BBCQ}
\newblock
\APACjournalVolNumPages{arXiv preprint arXiv:2311.12793}{}{}{,}
\newblock

\newblock

\PrintBackRefs{\CurrentBib}

\bibitem [\protect \citeauthoryear {%
R.~Chen%
, Chen%
, Jiao%
\BCBL {}\ \BBA {} Jia%
}{%
R.~Chen%
\ \protect \BOthers {.}}{%
{\protect \APACyear {2023}}%
}]{%
Fantasia3D}
\APACinsertmetastar {%
Fantasia3D}%
\begin{APACrefauthors}%
Chen, R.%
, Chen, Y.%
, Jiao, N.%
\BCBL {} Jia, K.%
\end{APACrefauthors}%
\unskip\
\newblock
\APACrefYearMonthDay{2023}{}{}.
\newblock
{\BBOQ}\APACrefatitle {Fantasia3d: Disentangling geometry and appearance for high-quality text-to-3d content creation} {Fantasia3d: Disentangling geometry and appearance for high-quality text-to-3d content creation}.{\BBCQ}
\newblock
 \APACrefbtitle {Proceedings of the IEEE/CVF international conference on computer vision} {Proceedings of the ieee/cvf international conference on computer vision}\ (\BPGS\ 22246--22256).
\PrintBackRefs{\CurrentBib}

\bibitem [\protect \citeauthoryear {%
Y.~Chen%
\ \protect \BOthers {.}}{%
Y.~Chen%
\ \protect \BOthers {.}}{%
{\protect \APACyear {2024}}%
}]{%
chen2024comboverse}
\APACinsertmetastar {%
chen2024comboverse}%
\begin{APACrefauthors}%
Chen, Y.%
, Wang, T.%
, Wu, T.%
, Pan, X.%
, Jia, K.%
\BCBL {} Liu, Z.%
\end{APACrefauthors}%
\unskip\
\newblock
\APACrefYearMonthDay{2024}{}{}.
\newblock
\APACrefbtitle {ComboVerse: Compositional 3D Assets Creation Using Spatially-Aware Diffusion Guidance.} {Comboverse: Compositional 3d assets creation using spatially-aware diffusion guidance.}
\PrintBackRefs{\CurrentBib}

\bibitem [\protect \citeauthoryear {%
Z.~Chen%
, Wang%
\BCBL {}\ \BBA {} Liu%
}{%
Z.~Chen%
\ \protect \BOthers {.}}{%
{\protect \APACyear {2023}}%
}]{%
GSGEN}
\APACinsertmetastar {%
GSGEN}%
\begin{APACrefauthors}%
Chen, Z.%
, Wang, F.%
\BCBL {} Liu, H.%
\end{APACrefauthors}%
\unskip\
\newblock
\APACrefYearMonthDay{2023}{}{}.
\newblock
{\BBOQ}\APACrefatitle {Text-to-3D using Gaussian Splatting} {Text-to-3d using gaussian splatting}.{\BBCQ}
\newblock
\APACjournalVolNumPages{CoRR}{abs/2309.16585}{}{,}
\newblock

\newblock

\PrintBackRefs{\CurrentBib}

\bibitem [\protect \citeauthoryear {%
Cheng%
, Lee%
, Tuyakov%
, Schwing%
\BCBL {}\ \BBA {} Gui%
}{%
Cheng%
\ \protect \BOthers {.}}{%
{\protect \APACyear {2023}}%
}]{%
cheng2023sdfusion}
\APACinsertmetastar {%
cheng2023sdfusion}%
\begin{APACrefauthors}%
Cheng, Y\BHBI C.%
, Lee, H\BHBI Y.%
, Tuyakov, S.%
, Schwing, A.%
\BCBL {} Gui, L.%
\end{APACrefauthors}%
\unskip\
\newblock
\APACrefYearMonthDay{2023}{}{}.
\newblock
{\BBOQ}\APACrefatitle {{SDFusion}: Multimodal 3D Shape Completion, Reconstruction, and Generation} {{SDFusion}: Multimodal 3d shape completion, reconstruction, and generation}.{\BBCQ}
\newblock
 \APACrefbtitle {CVPR.} {Cvpr.}
\PrintBackRefs{\CurrentBib}

\bibitem [\protect \citeauthoryear {%
Deitke%
\ \protect \BOthers {.}}{%
Deitke%
\ \protect \BOthers {.}}{%
{\protect \APACyear {2024}}%
}]{%
Objaverse-XL}
\APACinsertmetastar {%
Objaverse-XL}%
\begin{APACrefauthors}%
Deitke, M.%
, Liu, R.%
, Wallingford, M.%
, Ngo, H.%
, Michel, O.%
, Kusupati, A.%
\BDBL {}others%
\end{APACrefauthors}%
\unskip\
\newblock
\APACrefYearMonthDay{2024}{}{}.
\newblock
{\BBOQ}\APACrefatitle {Objaverse-xl: A universe of 10m+ 3d objects} {Objaverse-xl: A universe of 10m+ 3d objects}.{\BBCQ}
\newblock
\APACjournalVolNumPages{Advances in Neural Information Processing Systems}{36}{}{,}
\newblock

\newblock

\PrintBackRefs{\CurrentBib}

\bibitem [\protect \citeauthoryear {%
Deitke%
\ \protect \BOthers {.}}{%
Deitke%
\ \protect \BOthers {.}}{%
{\protect \APACyear {2023}}%
}]{%
Objaverse}
\APACinsertmetastar {%
Objaverse}%
\begin{APACrefauthors}%
Deitke, M.%
, Schwenk, D.%
, Salvador, J.%
, Weihs, L.%
, Michel, O.%
, VanderBilt, E.%
\BDBL {}Farhadi, A.%
\end{APACrefauthors}%
\unskip\
\newblock
\APACrefYearMonthDay{2023}{}{}.
\newblock
{\BBOQ}\APACrefatitle {Objaverse: A universe of annotated 3d objects} {Objaverse: A universe of annotated 3d objects}.{\BBCQ}
\newblock
 \APACrefbtitle {Proceedings of the IEEE/CVF Conference on Computer Vision and Pattern Recognition} {Proceedings of the ieee/cvf conference on computer vision and pattern recognition}\ (\BPGS\ 13142--13153).
\PrintBackRefs{\CurrentBib}

\bibitem [\protect \citeauthoryear {%
Di%
\ \protect \BOthers {.}}{%
Di%
\ \protect \BOthers {.}}{%
{\protect \APACyear {2024}}%
}]{%
Hyper-3DG}
\APACinsertmetastar {%
Hyper-3DG}%
\begin{APACrefauthors}%
Di, D.%
, Yang, J.%
, Luo, C.%
, Xue, Z.%
, Chen, W.%
, Yang, X.%
\BCBL {} Gao, Y.%
\end{APACrefauthors}%
\unskip\
\newblock
\APACrefYearMonthDay{2024}{}{}.
\newblock
{\BBOQ}\APACrefatitle {Hyper-3DG: Text-to-3D Gaussian Generation via Hypergraph} {Hyper-3dg: Text-to-3d gaussian generation via hypergraph}.{\BBCQ}
\newblock
\APACjournalVolNumPages{CoRR}{abs/2403.09236}{}{,}
\newblock

\newblock

\PrintBackRefs{\CurrentBib}

\bibitem [\protect \citeauthoryear {%
Downs%
\ \protect \BOthers {.}}{%
Downs%
\ \protect \BOthers {.}}{%
{\protect \APACyear {2022}}%
}]{%
GSO}
\APACinsertmetastar {%
GSO}%
\begin{APACrefauthors}%
Downs, L.%
, Francis, A.%
, Koenig, N.%
, Kinman, B.%
, Hickman, R.%
, Reymann, K.%
\BDBL {}Vanhoucke, V.%
\end{APACrefauthors}%
\unskip\
\newblock
\APACrefYearMonthDay{2022}{}{}.
\newblock
\APACrefbtitle {Google Scanned Objects: A High-Quality Dataset of 3D Scanned Household Items.} {Google scanned objects: A high-quality dataset of 3d scanned household items.}
\PrintBackRefs{\CurrentBib}

\bibitem [\protect \citeauthoryear {%
Elo%
}{%
Elo%
}{%
{\protect \APACyear {1967}}%
}]{%
elo}
\APACinsertmetastar {%
elo}%
\begin{APACrefauthors}%
Elo, A.E.%
\end{APACrefauthors}%
\unskip\
\newblock
\APACrefYearMonthDay{1967}{}{}.
\newblock
{\BBOQ}\APACrefatitle {The proposed uscf rating system, its development, theory, and applications} {The proposed uscf rating system, its development, theory, and applications}.{\BBCQ}
\newblock
\APACjournalVolNumPages{Chess life}{22}{8}{242--247,}
\newblock

\newblock

\PrintBackRefs{\CurrentBib}

\bibitem [\protect \citeauthoryear {%
Ethayarajh%
, Choi%
\BCBL {}\ \BBA {} Swayamdipta%
}{%
Ethayarajh%
\ \protect \BOthers {.}}{%
{\protect \APACyear {2022}}%
}]{%
SHP}
\APACinsertmetastar {%
SHP}%
\begin{APACrefauthors}%
Ethayarajh, K.%
, Choi, Y.%
\BCBL {} Swayamdipta, S.%
\end{APACrefauthors}%
\unskip\
\newblock
\APACrefYearMonthDay{2022}{}{}.
\newblock
{\BBOQ}\APACrefatitle {Understanding Dataset Difficulty with $\mathcal{V}$-Usable Information} {Understanding dataset difficulty with $\mathcal{V}$-usable information}.{\BBCQ}
\newblock
 \APACrefbtitle {International Conference on Machine Learning} {International conference on machine learning}\ (\BPGS\ 5988--6008).
\PrintBackRefs{\CurrentBib}

\bibitem [\protect \citeauthoryear {%
Gao*%
\ \protect \BOthers {.}}{%
Gao*%
\ \protect \BOthers {.}}{%
{\protect \APACyear {2024}}%
}]{%
gao2024cat3d}
\APACinsertmetastar {%
gao2024cat3d}%
\begin{APACrefauthors}%
Gao*, R.%
, Holynski*, A.%
, Henzler, P.%
, Brussee, A.%
, Martin-Brualla, R.%
, Srinivasan, P.P.%
\BDBL {}Poole*, B.%
\end{APACrefauthors}%
\unskip\
\newblock
\APACrefYearMonthDay{2024}{}{}.
\newblock
{\BBOQ}\APACrefatitle {CAT3D: Create Anything in 3D with Multi-View Diffusion Models} {Cat3d: Create anything in 3d with multi-view diffusion models}.{\BBCQ}
\newblock
\APACjournalVolNumPages{arXiv}{}{}{,}
\newblock

\newblock

\PrintBackRefs{\CurrentBib}

\bibitem [\protect \citeauthoryear {%
Guo%
\ \protect \BOthers {.}}{%
Guo%
\ \protect \BOthers {.}}{%
{\protect \APACyear {2023}}%
}]{%
threestudio2023}
\APACinsertmetastar {%
threestudio2023}%
\begin{APACrefauthors}%
Guo, Y\BHBI C.%
, Liu, Y\BHBI T.%
, Shao, R.%
, Laforte, C.%
, Voleti, V.%
, Luo, G.%
\BDBL {}Zhang, S\BHBI H.%
\end{APACrefauthors}%
\unskip\
\newblock
\APACrefYearMonthDay{2023}{}{}.
\newblock
\APACrefbtitle {threestudio: A unified framework for 3D content generation.} {threestudio: A unified framework for 3d content generation.}
\newblock
\APAChowpublished {\url{https://github.com/threestudio-project/threestudio}}.
\PrintBackRefs{\CurrentBib}

\bibitem [\protect \citeauthoryear {%
Gupta%
, Xiong%
, Nie%
, Jones%
\BCBL {}\ \BBA {} O{\u{g}}uz%
}{%
Gupta%
\ \protect \BOthers {.}}{%
{\protect \APACyear {2023}}%
}]{%
gupta20233dgen}
\APACinsertmetastar {%
gupta20233dgen}%
\begin{APACrefauthors}%
Gupta, A.%
, Xiong, W.%
, Nie, Y.%
, Jones, I.%
\BCBL {} O{\u{g}}uz, B.%
\end{APACrefauthors}%
\unskip\
\newblock
\APACrefYearMonthDay{2023}{}{}.
\newblock
{\BBOQ}\APACrefatitle {3dgen: Triplane latent diffusion for textured mesh generation} {3dgen: Triplane latent diffusion for textured mesh generation}.{\BBCQ}
\newblock
\APACjournalVolNumPages{arXiv preprint arXiv:2303.05371}{}{}{,}
\newblock

\newblock

\PrintBackRefs{\CurrentBib}

\bibitem [\protect \citeauthoryear {%
Y.~He%
\ \protect \BOthers {.}}{%
Y.~He%
\ \protect \BOthers {.}}{%
{\protect \APACyear {2023}}%
}]{%
T3_Bench}
\APACinsertmetastar {%
T3_Bench}%
\begin{APACrefauthors}%
He, Y.%
, Bai, Y.%
, Lin, M.%
, Zhao, W.%
, Hu, Y.%
, Sheng, J.%
\BDBL {}Liu, Y\BHBI J.%
\end{APACrefauthors}%
\unskip\
\newblock
\APACrefYearMonthDay{2023}{}{}.
\newblock
{\BBOQ}\APACrefatitle {T$^{3}$ Bench: Benchmarking Current Progress in Text-to-3D Generation} {T$^{3}$ bench: Benchmarking current progress in text-to-3d generation}.{\BBCQ}
\newblock
\APACjournalVolNumPages{arXiv preprint arXiv:2310.02977}{}{}{,}
\newblock

\newblock

\PrintBackRefs{\CurrentBib}

\bibitem [\protect \citeauthoryear {%
Z.~He%
\ \BBA {} Wang%
}{%
Z.~He%
\ \BBA {} Wang%
}{%
{\protect \APACyear {2023}}%
}]{%
OpenLRM}
\APACinsertmetastar {%
OpenLRM}%
\begin{APACrefauthors}%
He, Z.%
\BCBT {}\ \BBA {} Wang, T.%
\end{APACrefauthors}%
\unskip\
\newblock
\APACrefYearMonthDay{2023}{}{}.
\newblock
\APACrefbtitle {OpenLRM: Open-Source Large Reconstruction Models.} {Openlrm: Open-source large reconstruction models.}
\newblock
\APAChowpublished {\url{https://github.com/3DTopia/OpenLRM}}.
\PrintBackRefs{\CurrentBib}

\bibitem [\protect \citeauthoryear {%
Hessel%
, Holtzman%
, Forbes%
, Bras%
\BCBL {}\ \BBA {} Choi%
}{%
Hessel%
\ \protect \BOthers {.}}{%
{\protect \APACyear {2022}}%
}]{%
CLIPScore}
\APACinsertmetastar {%
CLIPScore}%
\begin{APACrefauthors}%
Hessel, J.%
, Holtzman, A.%
, Forbes, M.%
, Bras, R.L.%
\BCBL {} Choi, Y.%
\end{APACrefauthors}%
\unskip\
\newblock
\APACrefYearMonthDay{2022}{}{}.
\newblock
\APACrefbtitle {CLIPScore: A Reference-free Evaluation Metric for Image Captioning.} {Clipscore: A reference-free evaluation metric for image captioning.}
\PrintBackRefs{\CurrentBib}

\bibitem [\protect \citeauthoryear {%
F.~Hong%
\ \protect \BOthers {.}}{%
F.~Hong%
\ \protect \BOthers {.}}{%
{\protect \APACyear {2024}}%
}]{%
hong20243dtopia}
\APACinsertmetastar {%
hong20243dtopia}%
\begin{APACrefauthors}%
Hong, F.%
, Tang, J.%
, Cao, Z.%
, Shi, M.%
, Wu, T.%
, Chen, Z.%
\BDBL {}Liu, Z.%
\end{APACrefauthors}%
\unskip\
\newblock
\APACrefYearMonthDay{2024}{}{}.
\newblock
{\BBOQ}\APACrefatitle {3dtopia: Large text-to-3d generation model with hybrid diffusion priors} {3dtopia: Large text-to-3d generation model with hybrid diffusion priors}.{\BBCQ}
\newblock
\APACjournalVolNumPages{arXiv preprint arXiv:2403.02234}{}{}{,}
\newblock

\newblock

\PrintBackRefs{\CurrentBib}

\bibitem [\protect \citeauthoryear {%
Y.~Hong%
\ \protect \BOthers {.}}{%
Y.~Hong%
\ \protect \BOthers {.}}{%
{\protect \APACyear {2023}}%
}]{%
Lrm}
\APACinsertmetastar {%
Lrm}%
\begin{APACrefauthors}%
Hong, Y.%
, Zhang, K.%
, Gu, J.%
, Bi, S.%
, Zhou, Y.%
, Liu, D.%
\BDBL {}Tan, H.%
\end{APACrefauthors}%
\unskip\
\newblock
\APACrefYearMonthDay{2023}{}{}.
\newblock
{\BBOQ}\APACrefatitle {Lrm: Large reconstruction model for single image to 3d} {Lrm: Large reconstruction model for single image to 3d}.{\BBCQ}
\newblock
\APACjournalVolNumPages{arXiv preprint arXiv:2311.04400}{}{}{,}
\newblock

\newblock

\PrintBackRefs{\CurrentBib}

\bibitem [\protect \citeauthoryear {%
Ilharco%
\ \protect \BOthers {.}}{%
Ilharco%
\ \protect \BOthers {.}}{%
{\protect \APACyear {2021}}%
}]{%
OpenCLIP}
\APACinsertmetastar {%
OpenCLIP}%
\begin{APACrefauthors}%
Ilharco, G.%
, Wortsman, M.%
, Wightman, R.%
, Gordon, C.%
, Carlini, N.%
, Taori, R.%
\BDBL {}Schmidt, L.%
\end{APACrefauthors}%
\unskip\
\newblock
\APACrefYearMonthDay{2021}{{\APACmonth{07}}}{}.
\newblock
\APACrefbtitle {OpenCLIP.} {Openclip.}
\newblock
\APACaddressPublisher{}{Zenodo}.
\newblock
\begin{APACrefURL} {https://doi.org/10.5281/zenodo.5143773} \end{APACrefURL}
\newblock
\APACrefnote{If you use this software, please cite it as below.}
\PrintBackRefs{\CurrentBib}

\bibitem [\protect \citeauthoryear {%
Jun%
\ \BBA {} Nichol%
}{%
Jun%
\ \BBA {} Nichol%
}{%
{\protect \APACyear {2023}}%
}]{%
Shap•E}
\APACinsertmetastar {%
Shap•E}%
\begin{APACrefauthors}%
Jun, H.%
\BCBT {}\ \BBA {} Nichol, A.%
\end{APACrefauthors}%
\unskip\
\newblock
\APACrefYearMonthDay{2023}{}{}.
\newblock
{\BBOQ}\APACrefatitle {Shap-e: Generating conditional 3d implicit functions} {Shap-e: Generating conditional 3d implicit functions}.{\BBCQ}
\newblock
\APACjournalVolNumPages{arXiv preprint arXiv:2305.02463}{}{}{,}
\newblock

\newblock

\PrintBackRefs{\CurrentBib}

\bibitem [\protect \citeauthoryear {%
Kendall%
}{%
Kendall%
}{%
{\protect \APACyear {1938}}%
}]{%
Kendall}
\APACinsertmetastar {%
Kendall}%
\begin{APACrefauthors}%
Kendall, M.G.%
\end{APACrefauthors}%
\unskip\
\newblock
\APACrefYearMonthDay{1938}{}{}.
\newblock
{\BBOQ}\APACrefatitle {A new measure of rank correlation} {A new measure of rank correlation}.{\BBCQ}
\newblock
\APACjournalVolNumPages{Biometrika}{30}{1/2}{81--93,}
\newblock

\newblock

\PrintBackRefs{\CurrentBib}

\bibitem [\protect \citeauthoryear {%
Kerbl%
, Kopanas%
, Leimk{\"u}hler%
\BCBL {}\ \BBA {} Drettakis%
}{%
Kerbl%
\ \protect \BOthers {.}}{%
{\protect \APACyear {2023}}%
}]{%
kerbl20233d}
\APACinsertmetastar {%
kerbl20233d}%
\begin{APACrefauthors}%
Kerbl, B.%
, Kopanas, G.%
, Leimk{\"u}hler, T.%
\BCBL {} Drettakis, G.%
\end{APACrefauthors}%
\unskip\
\newblock
\APACrefYearMonthDay{2023}{}{}.
\newblock
{\BBOQ}\APACrefatitle {3d gaussian splatting for real-time radiance field rendering} {3d gaussian splatting for real-time radiance field rendering}.{\BBCQ}
\newblock
\APACjournalVolNumPages{ACM Transactions on Graphics}{42}{4}{1--14,}
\newblock

\newblock

\PrintBackRefs{\CurrentBib}

\bibitem [\protect \citeauthoryear {%
Kirstain%
\ \protect \BOthers {.}}{%
Kirstain%
\ \protect \BOthers {.}}{%
{\protect \APACyear {2023}}%
}]{%
Pick-a-Pic}
\APACinsertmetastar {%
Pick-a-Pic}%
\begin{APACrefauthors}%
Kirstain, Y.%
, Polyak, A.%
, Singer, U.%
, Matiana, S.%
, Penna, J.%
\BCBL {} Levy, O.%
\end{APACrefauthors}%
\unskip\
\newblock
\APACrefYearMonthDay{2023}{}{}.
\newblock
{\BBOQ}\APACrefatitle {Pick-a-pic: An open dataset of user preferences for text-to-image generation} {Pick-a-pic: An open dataset of user preferences for text-to-image generation}.{\BBCQ}
\newblock
\APACjournalVolNumPages{Advances in Neural Information Processing Systems}{36}{}{36652--36663,}
\newblock

\newblock

\PrintBackRefs{\CurrentBib}

\bibitem [\protect \citeauthoryear {%
Kong%
\ \protect \BOthers {.}}{%
Kong%
\ \protect \BOthers {.}}{%
{\protect \APACyear {2024}}%
}]{%
EscherNet}
\APACinsertmetastar {%
EscherNet}%
\begin{APACrefauthors}%
Kong, X.%
, Liu, S.%
, Lyu, X.%
, Taher, M.%
, Qi, X.%
\BCBL {} Davison, A.J.%
\end{APACrefauthors}%
\unskip\
\newblock
\APACrefYearMonthDay{2024}{}{}.
\newblock
{\BBOQ}\APACrefatitle {EscherNet: A Generative Model for Scalable View Synthesis} {Eschernet: A generative model for scalable view synthesis}.{\BBCQ}
\newblock
\APACjournalVolNumPages{arXiv preprint arXiv:2402.03908}{}{}{,}
\newblock

\newblock

\PrintBackRefs{\CurrentBib}

\bibitem [\protect \citeauthoryear {%
K{\"o}pf%
\ \protect \BOthers {.}}{%
K{\"o}pf%
\ \protect \BOthers {.}}{%
{\protect \APACyear {2024}}%
}]{%
OASST}
\APACinsertmetastar {%
OASST}%
\begin{APACrefauthors}%
K{\"o}pf, A.%
, Kilcher, Y.%
, von R{\"u}tte, D.%
, Anagnostidis, S.%
, Tam, Z.R.%
, Stevens, K.%
\BDBL {}others%
\end{APACrefauthors}%
\unskip\
\newblock
\APACrefYearMonthDay{2024}{}{}.
\newblock
{\BBOQ}\APACrefatitle {Openassistant conversations-democratizing large language model alignment} {Openassistant conversations-democratizing large language model alignment}.{\BBCQ}
\newblock
\APACjournalVolNumPages{Advances in Neural Information Processing Systems}{36}{}{,}
\newblock

\newblock

\PrintBackRefs{\CurrentBib}

\bibitem [\protect \citeauthoryear {%
Ku%
\ \protect \BOthers {.}}{%
Ku%
\ \protect \BOthers {.}}{%
{\protect \APACyear {2023}}%
}]{%
ImagenHub}
\APACinsertmetastar {%
ImagenHub}%
\begin{APACrefauthors}%
Ku, M.%
, Li, T.%
, Zhang, K.%
, Lu, Y.%
, Fu, X.%
, Zhuang, W.%
\BCBL {} Chen, W.%
\end{APACrefauthors}%
\unskip\
\newblock
\APACrefYearMonthDay{2023}{}{}.
\newblock
{\BBOQ}\APACrefatitle {Imagenhub: Standardizing the evaluation of conditional image generation models} {Imagenhub: Standardizing the evaluation of conditional image generation models}.{\BBCQ}
\newblock
\APACjournalVolNumPages{arXiv preprint arXiv:2310.01596}{}{}{,}
\newblock

\newblock

\PrintBackRefs{\CurrentBib}

\bibitem [\protect \citeauthoryear {%
J.~Li%
, Li%
, Xiong%
\BCBL {}\ \BBA {} Hoi%
}{%
J.~Li%
\ \protect \BOthers {.}}{%
{\protect \APACyear {2022}}%
}]{%
BLIP}
\APACinsertmetastar {%
BLIP}%
\begin{APACrefauthors}%
Li, J.%
, Li, D.%
, Xiong, C.%
\BCBL {} Hoi, S.%
\end{APACrefauthors}%
\unskip\
\newblock
\APACrefYearMonthDay{2022}{}{}.
\newblock
{\BBOQ}\APACrefatitle {BLIP: Bootstrapping Language-Image Pre-training for Unified Vision-Language Understanding and Generation} {Blip: Bootstrapping language-image pre-training for unified vision-language understanding and generation}.{\BBCQ}
\newblock
 \APACrefbtitle {ICML.} {Icml.}
\PrintBackRefs{\CurrentBib}

\bibitem [\protect \citeauthoryear {%
J.~Li%
\ \protect \BOthers {.}}{%
J.~Li%
\ \protect \BOthers {.}}{%
{\protect \APACyear {2023}}%
}]{%
Instant3d}
\APACinsertmetastar {%
Instant3d}%
\begin{APACrefauthors}%
Li, J.%
, Tan, H.%
, Zhang, K.%
, Xu, Z.%
, Luan, F.%
, Xu, Y.%
\BDBL {}Bi, S.%
\end{APACrefauthors}%
\unskip\
\newblock
\APACrefYearMonthDay{2023}{}{}.
\newblock
{\BBOQ}\APACrefatitle {Instant3d: Fast text-to-3d with sparse-view generation and large reconstruction model} {Instant3d: Fast text-to-3d with sparse-view generation and large reconstruction model}.{\BBCQ}
\newblock
\APACjournalVolNumPages{arXiv preprint arXiv:2311.06214}{}{}{,}
\newblock

\newblock

\PrintBackRefs{\CurrentBib}

\bibitem [\protect \citeauthoryear {%
W.~Li%
, Chen%
, Chen%
\BCBL {}\ \BBA {} Tan%
}{%
W.~Li%
\ \protect \BOthers {.}}{%
{\protect \APACyear {2023}}%
}]{%
SweetDreamer}
\APACinsertmetastar {%
SweetDreamer}%
\begin{APACrefauthors}%
Li, W.%
, Chen, R.%
, Chen, X.%
\BCBL {} Tan, P.%
\end{APACrefauthors}%
\unskip\
\newblock
\APACrefYearMonthDay{2023}{}{}.
\newblock
{\BBOQ}\APACrefatitle {SweetDreamer: Aligning Geometric Priors in 2D Diffusion for Consistent Text-to-3D} {Sweetdreamer: Aligning geometric priors in 2d diffusion for consistent text-to-3d}.{\BBCQ}
\newblock
\APACjournalVolNumPages{arxiv:2310.02596}{}{}{,}
\newblock

\newblock

\PrintBackRefs{\CurrentBib}

\bibitem [\protect \citeauthoryear {%
Liang%
\ \protect \BOthers {.}}{%
Liang%
\ \protect \BOthers {.}}{%
{\protect \APACyear {2023}}%
}]{%
Luciddreamer}
\APACinsertmetastar {%
Luciddreamer}%
\begin{APACrefauthors}%
Liang, Y.%
, Yang, X.%
, Lin, J.%
, Li, H.%
, Xu, X.%
\BCBL {} Chen, Y.%
\end{APACrefauthors}%
\unskip\
\newblock
\APACrefYearMonthDay{2023}{}{}.
\newblock
{\BBOQ}\APACrefatitle {Luciddreamer: Towards high-fidelity text-to-3d generation via interval score matching} {Luciddreamer: Towards high-fidelity text-to-3d generation via interval score matching}.{\BBCQ}
\newblock
\APACjournalVolNumPages{arXiv preprint arXiv:2311.11284}{}{}{,}
\newblock

\newblock

\PrintBackRefs{\CurrentBib}

\bibitem [\protect \citeauthoryear {%
Lin%
\ \protect \BOthers {.}}{%
Lin%
\ \protect \BOthers {.}}{%
{\protect \APACyear {2023}}%
}]{%
Magic3D}
\APACinsertmetastar {%
Magic3D}%
\begin{APACrefauthors}%
Lin, C\BHBI H.%
, Gao, J.%
, Tang, L.%
, Takikawa, T.%
, Zeng, X.%
, Huang, X.%
\BDBL {}Lin, T\BHBI Y.%
\end{APACrefauthors}%
\unskip\
\newblock
\APACrefYearMonthDay{2023}{}{}.
\newblock
{\BBOQ}\APACrefatitle {Magic3d: High-resolution text-to-3d content creation} {Magic3d: High-resolution text-to-3d content creation}.{\BBCQ}
\newblock
 \APACrefbtitle {Proceedings of the IEEE/CVF Conference on Computer Vision and Pattern Recognition} {Proceedings of the ieee/cvf conference on computer vision and pattern recognition}\ (\BPGS\ 300--309).
\PrintBackRefs{\CurrentBib}

\bibitem [\protect \citeauthoryear {%
F.~Liu%
, Wu%
, Wei%
, Rao%
\BCBL {}\ \BBA {} Duan%
}{%
F.~Liu%
\ \protect \BOthers {.}}{%
{\protect \APACyear {2023}}%
}]{%
Sherpa3D}
\APACinsertmetastar {%
Sherpa3D}%
\begin{APACrefauthors}%
Liu, F.%
, Wu, D.%
, Wei, Y.%
, Rao, Y.%
\BCBL {} Duan, Y.%
\end{APACrefauthors}%
\unskip\
\newblock
\APACrefYearMonthDay{2023}{}{}.
\newblock
\APACrefbtitle {Sherpa3D: Boosting High-Fidelity Text-to-3D Generation via Coarse 3D Prior.} {Sherpa3d: Boosting high-fidelity text-to-3d generation via coarse 3d prior.}
\PrintBackRefs{\CurrentBib}

\bibitem [\protect \citeauthoryear {%
H.~Liu%
, Li%
, Wu%
\BCBL {}\ \BBA {} Lee%
}{%
H.~Liu%
\ \protect \BOthers {.}}{%
{\protect \APACyear {2024}}%
}]{%
LLaVa}
\APACinsertmetastar {%
LLaVa}%
\begin{APACrefauthors}%
Liu, H.%
, Li, C.%
, Wu, Q.%
\BCBL {} Lee, Y.J.%
\end{APACrefauthors}%
\unskip\
\newblock
\APACrefYearMonthDay{2024}{}{}.
\newblock
{\BBOQ}\APACrefatitle {Visual instruction tuning} {Visual instruction tuning}.{\BBCQ}
\newblock
\APACjournalVolNumPages{Advances in neural information processing systems}{36}{}{,}
\newblock

\newblock

\PrintBackRefs{\CurrentBib}

\bibitem [\protect \citeauthoryear {%
M.~Liu%
\ \protect \BOthers {.}}{%
M.~Liu%
\ \protect \BOthers {.}}{%
{\protect \APACyear {2023}}%
}]{%
One-2-3-45++}
\APACinsertmetastar {%
One-2-3-45++}%
\begin{APACrefauthors}%
Liu, M.%
, Shi, R.%
, Chen, L.%
, Zhang, Z.%
, Xu, C.%
, Wei, X.%
\BDBL {}Su, H.%
\end{APACrefauthors}%
\unskip\
\newblock
\APACrefYearMonthDay{2023}{}{}.
\newblock
{\BBOQ}\APACrefatitle {One-2-3-45++: Fast Single Image to 3D Objects with Consistent Multi-View Generation and 3D Diffusion} {One-2-3-45++: Fast single image to 3d objects with consistent multi-view generation and 3d diffusion}.{\BBCQ}
\newblock
\APACjournalVolNumPages{arXiv preprint arXiv:2311.07885}{}{}{,}
\newblock

\newblock

\PrintBackRefs{\CurrentBib}

\bibitem [\protect \citeauthoryear {%
M.~Liu%
, Shi%
\BCBL {}\ \protect \BOthers {.}}{%
M.~Liu%
, Shi%
\BCBL {}\ \protect \BOthers {.}}{%
{\protect \APACyear {2024}}%
}]{%
openshape}
\APACinsertmetastar {%
openshape}%
\begin{APACrefauthors}%
Liu, M.%
, Shi, R.%
, Kuang, K.%
, Zhu, Y.%
, Li, X.%
, Han, S.%
\BDBL {}Su, H.%
\end{APACrefauthors}%
\unskip\
\newblock
\APACrefYearMonthDay{2024}{}{}.
\newblock
{\BBOQ}\APACrefatitle {Openshape: Scaling up 3d shape representation towards open-world understanding} {Openshape: Scaling up 3d shape representation towards open-world understanding}.{\BBCQ}
\newblock
\APACjournalVolNumPages{Advances in neural information processing systems}{36}{}{,}
\newblock

\newblock

\PrintBackRefs{\CurrentBib}

\bibitem [\protect \citeauthoryear {%
M.~Liu%
, Xu%
\BCBL {}\ \protect \BOthers {.}}{%
M.~Liu%
, Xu%
\BCBL {}\ \protect \BOthers {.}}{%
{\protect \APACyear {2024}}%
}]{%
One-2-3-45}
\APACinsertmetastar {%
One-2-3-45}%
\begin{APACrefauthors}%
Liu, M.%
, Xu, C.%
, Jin, H.%
, Chen, L.%
, Varma~T, M.%
, Xu, Z.%
\BCBL {} Su, H.%
\end{APACrefauthors}%
\unskip\
\newblock
\APACrefYearMonthDay{2024}{}{}.
\newblock
{\BBOQ}\APACrefatitle {One-2-3-45: Any single image to 3d mesh in 45 seconds without per-shape optimization} {One-2-3-45: Any single image to 3d mesh in 45 seconds without per-shape optimization}.{\BBCQ}
\newblock
\APACjournalVolNumPages{Advances in Neural Information Processing Systems}{36}{}{,}
\newblock

\newblock

\PrintBackRefs{\CurrentBib}

\bibitem [\protect \citeauthoryear {%
R.~Liu%
\ \protect \BOthers {.}}{%
R.~Liu%
\ \protect \BOthers {.}}{%
{\protect \APACyear {2023}}%
}]{%
Zero-1-to-3}
\APACinsertmetastar {%
Zero-1-to-3}%
\begin{APACrefauthors}%
Liu, R.%
, Wu, R.%
, Hoorick, B.V.%
, Tokmakov, P.%
, Zakharov, S.%
\BCBL {} Vondrick, C.%
\end{APACrefauthors}%
\unskip\
\newblock
\APACrefYearMonthDay{2023}{}{}.
\newblock
\APACrefbtitle {Zero-1-to-3: Zero-shot One Image to 3D Object.} {Zero-1-to-3: Zero-shot one image to 3d object.}
\PrintBackRefs{\CurrentBib}

\bibitem [\protect \citeauthoryear {%
Y.~Liu%
\ \protect \BOthers {.}}{%
Y.~Liu%
\ \protect \BOthers {.}}{%
{\protect \APACyear {2023}}%
}]{%
SyncDreamer}
\APACinsertmetastar {%
SyncDreamer}%
\begin{APACrefauthors}%
Liu, Y.%
, Lin, C.%
, Zeng, Z.%
, Long, X.%
, Liu, L.%
, Komura, T.%
\BCBL {} Wang, W.%
\end{APACrefauthors}%
\unskip\
\newblock
\APACrefYearMonthDay{2023}{}{}.
\newblock
{\BBOQ}\APACrefatitle {SyncDreamer: Generating Multiview-consistent Images from a Single-view Image} {Syncdreamer: Generating multiview-consistent images from a single-view image}.{\BBCQ}
\newblock
\APACjournalVolNumPages{arXiv preprint arXiv:2309.03453}{}{}{,}
\newblock

\newblock

\PrintBackRefs{\CurrentBib}

\bibitem [\protect \citeauthoryear {%
Long%
\ \protect \BOthers {.}}{%
Long%
\ \protect \BOthers {.}}{%
{\protect \APACyear {2023}}%
}]{%
Wonder3D}
\APACinsertmetastar {%
Wonder3D}%
\begin{APACrefauthors}%
Long, X.%
, Guo, Y\BHBI C.%
, Lin, C.%
, Liu, Y.%
, Dou, Z.%
, Liu, L.%
\BDBL {}others%
\end{APACrefauthors}%
\unskip\
\newblock
\APACrefYearMonthDay{2023}{}{}.
\newblock
{\BBOQ}\APACrefatitle {Wonder3D: Single Image to 3D using Cross-Domain Diffusion} {Wonder3d: Single image to 3d using cross-domain diffusion}.{\BBCQ}
\newblock
\APACjournalVolNumPages{arXiv preprint arXiv:2310.15008}{}{}{,}
\newblock

\newblock

\PrintBackRefs{\CurrentBib}

\bibitem [\protect \citeauthoryear {%
Lu%
\ \protect \BOthers {.}}{%
Lu%
\ \protect \BOthers {.}}{%
{\protect \APACyear {2024}}%
}]{%
WildWision-Arena}
\APACinsertmetastar {%
WildWision-Arena}%
\begin{APACrefauthors}%
Lu, Y.%
, Jiang, D.%
, Chen, W.%
, Wang, W.%
, Choi, Y.%
\BCBL {} Lin, B.Y.%
\end{APACrefauthors}%
\unskip\
\newblock
\APACrefYearMonthDay{2024}{February}{}.
\newblock
\APACrefbtitle {WildVision Arena: Benchmarking Multimodal LLMs in the Wild.} {Wildvision arena: Benchmarking multimodal llms in the wild.}
\newblock
\begin{APACrefURL} {https://huggingface.co/spaces/WildVision/vision-arena/} \end{APACrefURL}
\PrintBackRefs{\CurrentBib}

\bibitem [\protect \citeauthoryear {%
Luo%
, Johnson%
\BCBL {}\ \BBA {} Lee%
}{%
Luo%
\ \protect \BOthers {.}}{%
{\protect \APACyear {2024}}%
}]{%
cap3d}
\APACinsertmetastar {%
cap3d}%
\begin{APACrefauthors}%
Luo, T.%
, Johnson, J.%
\BCBL {} Lee, H.%
\end{APACrefauthors}%
\unskip\
\newblock
\APACrefYearMonthDay{2024}{}{}.
\newblock
{\BBOQ}\APACrefatitle {View Selection for 3D Captioning via Diffusion Ranking} {View selection for 3d captioning via diffusion ranking}.{\BBCQ}
\newblock
\APACjournalVolNumPages{arXiv preprint arXiv:2404.07984}{}{}{,}
\newblock

\newblock

\PrintBackRefs{\CurrentBib}

\bibitem [\protect \citeauthoryear {%
Ma%
\ \protect \BOthers {.}}{%
Ma%
\ \protect \BOthers {.}}{%
{\protect \APACyear {2023}}%
}]{%
GeoDream}
\APACinsertmetastar {%
GeoDream}%
\begin{APACrefauthors}%
Ma, B.%
, Deng, H.%
, Zhou, J.%
, Liu, Y\BHBI S.%
, Huang, T.%
\BCBL {} Wang, X.%
\end{APACrefauthors}%
\unskip\
\newblock
\APACrefYearMonthDay{2023}{}{}.
\newblock
{\BBOQ}\APACrefatitle {Geodream: Disentangling 2d and geometric priors for high-fidelity and consistent 3d generation} {Geodream: Disentangling 2d and geometric priors for high-fidelity and consistent 3d generation}.{\BBCQ}
\newblock
\APACjournalVolNumPages{arXiv preprint arXiv:2311.17971}{}{}{,}
\newblock

\newblock

\PrintBackRefs{\CurrentBib}

\bibitem [\protect \citeauthoryear {%
Melas{-}Kyriazi%
\ \protect \BOthers {.}}{%
Melas{-}Kyriazi%
\ \protect \BOthers {.}}{%
{\protect \APACyear {2024}}%
}]{%
IM-3D}
\APACinsertmetastar {%
IM-3D}%
\begin{APACrefauthors}%
Melas{-}Kyriazi, L.%
, Laina, I.%
, Rupprecht, C.%
, Neverova, N.%
, Vedaldi, A.%
, Gafni, O.%
\BCBL {} Kokkinos, F.%
\end{APACrefauthors}%
\unskip\
\newblock
\APACrefYearMonthDay{2024}{}{}.
\newblock
{\BBOQ}\APACrefatitle {{IM-3D:} Iterative Multiview Diffusion and Reconstruction for High-Quality 3D Generation} {{IM-3D:} iterative multiview diffusion and reconstruction for high-quality 3d generation}.{\BBCQ}
\newblock
\APACjournalVolNumPages{CoRR}{abs/2402.08682}{}{,}
\newblock

\newblock

\PrintBackRefs{\CurrentBib}

\bibitem [\protect \citeauthoryear {%
Metzer%
, Richardson%
, Patashnik%
, Giryes%
\BCBL {}\ \BBA {} Cohen-Or%
}{%
Metzer%
\ \protect \BOthers {.}}{%
{\protect \APACyear {2023}}%
}]{%
Latent-NeRF}
\APACinsertmetastar {%
Latent-NeRF}%
\begin{APACrefauthors}%
Metzer, G.%
, Richardson, E.%
, Patashnik, O.%
, Giryes, R.%
\BCBL {} Cohen-Or, D.%
\end{APACrefauthors}%
\unskip\
\newblock
\APACrefYearMonthDay{2023}{}{}.
\newblock
{\BBOQ}\APACrefatitle {Latent-nerf for shape-guided generation of 3d shapes and textures} {Latent-nerf for shape-guided generation of 3d shapes and textures}.{\BBCQ}
\newblock
 \APACrefbtitle {Proceedings of the IEEE/CVF Conference on Computer Vision and Pattern Recognition} {Proceedings of the ieee/cvf conference on computer vision and pattern recognition}\ (\BPGS\ 12663--12673).
\PrintBackRefs{\CurrentBib}

\bibitem [\protect \citeauthoryear {%
mrfakename%
\ \protect \BOthers {.}}{%
mrfakename%
\ \protect \BOthers {.}}{%
{\protect \APACyear {{\protect \bibnodate {}}}}%
}]{%
tts-arena}
\APACinsertmetastar {%
tts-arena}%
\begin{APACrefauthors}%
mrfakename%
, Srivastav, V.%
, Fourrier, C.%
, Pouget, L.%
, Lacombe, Y.%
, main%
\BCBL {} Gandhi, S.%
\end{APACrefauthors}%
\unskip\
\newblock
\APACrefYearMonthDay{{\protect \bibnodate {}}}{}{}.
\newblock
\APACrefbtitle {Text to Speech Arena.} {Text to speech arena.}
\newblock
\APAChowpublished {\url{https://huggingface.co/spaces/TTS-AGI/TTS-Arena}}.
\newblock
\APACaddressPublisher{}{Hugging Face}.
\PrintBackRefs{\CurrentBib}

\bibitem [\protect \citeauthoryear {%
Nakano%
\ \protect \BOthers {.}}{%
Nakano%
\ \protect \BOthers {.}}{%
{\protect \APACyear {2021}}%
}]{%
Webgpt}
\APACinsertmetastar {%
Webgpt}%
\begin{APACrefauthors}%
Nakano, R.%
, Hilton, J.%
, Balaji, S.%
, Wu, J.%
, Ouyang, L.%
, Kim, C.%
\BDBL {}Schulman, J.%
\end{APACrefauthors}%
\unskip\
\newblock
\APACrefYearMonthDay{2021}{}{}.
\newblock
{\BBOQ}\APACrefatitle {WebGPT: Browser-assisted question-answering with human feedback} {Webgpt: Browser-assisted question-answering with human feedback}.{\BBCQ}
\newblock
 \APACrefbtitle {arXiv.} {arxiv.}
\PrintBackRefs{\CurrentBib}

\bibitem [\protect \citeauthoryear {%
Nichol%
, Jun%
, Dhariwal%
, Mishkin%
\BCBL {}\ \BBA {} Chen%
}{%
Nichol%
\ \protect \BOthers {.}}{%
{\protect \APACyear {2022}}%
}]{%
Ponit-e}
\APACinsertmetastar {%
Ponit-e}%
\begin{APACrefauthors}%
Nichol, A.%
, Jun, H.%
, Dhariwal, P.%
, Mishkin, P.%
\BCBL {} Chen, M.%
\end{APACrefauthors}%
\unskip\
\newblock
\APACrefYearMonthDay{2022}{}{}.
\newblock
{\BBOQ}\APACrefatitle {Point-e: A system for generating 3d point clouds from complex prompts} {Point-e: A system for generating 3d point clouds from complex prompts}.{\BBCQ}
\newblock
\APACjournalVolNumPages{arXiv preprint arXiv:2212.08751}{}{}{,}
\newblock

\newblock

\PrintBackRefs{\CurrentBib}

\bibitem [\protect \citeauthoryear {%
Ouyang%
\ \protect \BOthers {.}}{%
Ouyang%
\ \protect \BOthers {.}}{%
{\protect \APACyear {2022}}%
}]{%
InstructGPT}
\APACinsertmetastar {%
InstructGPT}%
\begin{APACrefauthors}%
Ouyang, L.%
, Wu, J.%
, Jiang, X.%
, Almeida, D.%
, Wainwright, C.%
, Mishkin, P.%
\BDBL {}others%
\end{APACrefauthors}%
\unskip\
\newblock
\APACrefYearMonthDay{2022}{}{}.
\newblock
{\BBOQ}\APACrefatitle {Training language models to follow instructions with human feedback} {Training language models to follow instructions with human feedback}.{\BBCQ}
\newblock
\APACjournalVolNumPages{Advances in neural information processing systems}{35}{}{27730--27744,}
\newblock

\newblock

\PrintBackRefs{\CurrentBib}

\bibitem [\protect \citeauthoryear {%
Poole%
, Jain%
, Barron%
\BCBL {}\ \BBA {} Mildenhall%
}{%
Poole%
\ \protect \BOthers {.}}{%
{\protect \APACyear {2022}}%
}]{%
Dreamfusion}
\APACinsertmetastar {%
Dreamfusion}%
\begin{APACrefauthors}%
Poole, B.%
, Jain, A.%
, Barron, J.T.%
\BCBL {} Mildenhall, B.%
\end{APACrefauthors}%
\unskip\
\newblock
\APACrefYearMonthDay{2022}{}{}.
\newblock
{\BBOQ}\APACrefatitle {Dreamfusion: Text-to-3d using 2d diffusion} {Dreamfusion: Text-to-3d using 2d diffusion}.{\BBCQ}
\newblock
\APACjournalVolNumPages{arXiv preprint arXiv:2209.14988}{}{}{,}
\newblock

\newblock

\PrintBackRefs{\CurrentBib}

\bibitem [\protect \citeauthoryear {%
Qi%
\ \protect \BOthers {.}}{%
Qi%
\ \protect \BOthers {.}}{%
{\protect \APACyear {2024}}%
}]{%
gpt4point}
\APACinsertmetastar {%
gpt4point}%
\begin{APACrefauthors}%
Qi, Z.%
, Fang, Y.%
, Sun, Z.%
, Wu, X.%
, Wu, T.%
, Wang, J.%
\BDBL {}Zhao, H.%
\end{APACrefauthors}%
\unskip\
\newblock
\APACrefYearMonthDay{2024}{}{}.
\newblock
{\BBOQ}\APACrefatitle {Gpt4point: A unified framework for point-language understanding and generation} {Gpt4point: A unified framework for point-language understanding and generation}.{\BBCQ}
\newblock
 \APACrefbtitle {Proceedings of the IEEE/CVF Conference on Computer Vision and Pattern Recognition} {Proceedings of the ieee/cvf conference on computer vision and pattern recognition}\ (\BPGS\ 26417--26427).
\PrintBackRefs{\CurrentBib}

\bibitem [\protect \citeauthoryear {%
Qian%
\ \protect \BOthers {.}}{%
Qian%
\ \protect \BOthers {.}}{%
{\protect \APACyear {2024}}%
}]{%
Magic123}
\APACinsertmetastar {%
Magic123}%
\begin{APACrefauthors}%
Qian, G.%
, Mai, J.%
, Hamdi, A.%
, Ren, J.%
, Siarohin, A.%
, Li, B.%
\BDBL {}Ghanem, B.%
\end{APACrefauthors}%
\unskip\
\newblock
\APACrefYearMonthDay{2024}{}{}.
\newblock
{\BBOQ}\APACrefatitle {Magic123: One Image to High-Quality 3D Object Generation Using Both 2D and 3D Diffusion Priors} {Magic123: One image to high-quality 3d object generation using both 2d and 3d diffusion priors}.{\BBCQ}
\newblock
 \APACrefbtitle {The Twelfth International Conference on Learning Representations (ICLR).} {The twelfth international conference on learning representations (iclr).}
\newblock
\begin{APACrefURL} {https://openreview.net/forum?id=0jHkUDyEO9} \end{APACrefURL}
\PrintBackRefs{\CurrentBib}

\bibitem [\protect \citeauthoryear {%
Qin%
\ \protect \BOthers {.}}{%
Qin%
\ \protect \BOthers {.}}{%
{\protect \APACyear {2020}}%
}]{%
rembg}
\APACinsertmetastar {%
rembg}%
\begin{APACrefauthors}%
Qin, X.%
, Zhang, Z.%
, Huang, C.%
, Dehghan, M.%
, Za{\"{\i}}ane, O.R.%
\BCBL {} J{\"{a}}gersand, M.%
\end{APACrefauthors}%
\unskip\
\newblock
\APACrefYearMonthDay{2020}{}{}.
\newblock
{\BBOQ}\APACrefatitle {U\({}^{\mbox{2}}\)-Net: Going deeper with nested U-structure for salient object detection} {U\({}^{\mbox{2}}\)-net: Going deeper with nested u-structure for salient object detection}.{\BBCQ}
\newblock
\APACjournalVolNumPages{Pattern Recognit.}{106}{}{107404,}
\newblock

\newblock

\PrintBackRefs{\CurrentBib}

\bibitem [\protect \citeauthoryear {%
Radford%
\ \protect \BOthers {.}}{%
Radford%
\ \protect \BOthers {.}}{%
{\protect \APACyear {2021}}%
}]{%
Clip}
\APACinsertmetastar {%
Clip}%
\begin{APACrefauthors}%
Radford, A.%
, Kim, J.%
, Hallacy, C.%
, Ramesh, A.%
, Goh, G.%
, Agarwal, S.%
\BDBL {}Sutskever, I.%
\end{APACrefauthors}%
\unskip\
\newblock
\APACrefYearMonthDay{2021}{Feb}{}.
\newblock
{\BBOQ}\APACrefatitle {Learning Transferable Visual Models From Natural Language Supervision} {Learning transferable visual models from natural language supervision}.{\BBCQ}
\newblock
\APACjournalVolNumPages{Cornell University - arXiv,Cornell University - arXiv}{}{}{,}
\newblock

\newblock

\PrintBackRefs{\CurrentBib}

\bibitem [\protect \citeauthoryear {%
Raj%
\ \protect \BOthers {.}}{%
Raj%
\ \protect \BOthers {.}}{%
{\protect \APACyear {2023}}%
}]{%
raj2023dreambooth3d}
\APACinsertmetastar {%
raj2023dreambooth3d}%
\begin{APACrefauthors}%
Raj, A.%
, Kaza, S.%
, Poole, B.%
, Niemeyer, M.%
, Ruiz, N.%
, Mildenhall, B.%
\BDBL {}others%
\end{APACrefauthors}%
\unskip\
\newblock
\APACrefYearMonthDay{2023}{}{}.
\newblock
{\BBOQ}\APACrefatitle {Dreambooth3d: Subject-driven text-to-3d generation} {Dreambooth3d: Subject-driven text-to-3d generation}.{\BBCQ}
\newblock
 \APACrefbtitle {Proceedings of the IEEE/CVF International Conference on Computer Vision} {Proceedings of the ieee/cvf international conference on computer vision}\ (\BPGS\ 2349--2359).
\PrintBackRefs{\CurrentBib}

\bibitem [\protect \citeauthoryear {%
Rombach%
, Blattmann%
, Lorenz%
, Esser%
\BCBL {}\ \BBA {} Ommer%
}{%
Rombach%
\ \protect \BOthers {.}}{%
{\protect \APACyear {2022}}%
}]{%
Stable_diffusion}
\APACinsertmetastar {%
Stable_diffusion}%
\begin{APACrefauthors}%
Rombach, R.%
, Blattmann, A.%
, Lorenz, D.%
, Esser, P.%
\BCBL {} Ommer, B.%
\end{APACrefauthors}%
\unskip\
\newblock
\APACrefYearMonthDay{2022}{Jun}{}.
\newblock
{\BBOQ}\APACrefatitle {High-Resolution Image Synthesis with Latent Diffusion Models} {High-resolution image synthesis with latent diffusion models}.{\BBCQ}
\newblock
 \APACrefbtitle {2022 IEEE/CVF Conference on Computer Vision and Pattern Recognition (CVPR).} {2022 ieee/cvf conference on computer vision and pattern recognition (cvpr).}
\newblock
\begin{APACrefURL} {http://dx.doi.org/10.1109/cvpr52688.2022.01042} \end{APACrefURL}
\PrintBackRefs{\CurrentBib}

\bibitem [\protect \citeauthoryear {%
Saharia%
\ \protect \BOthers {.}}{%
Saharia%
\ \protect \BOthers {.}}{%
{\protect \APACyear {2022}}%
}]{%
Imagen}
\APACinsertmetastar {%
Imagen}%
\begin{APACrefauthors}%
Saharia, C.%
, Chan, W.%
, Saxena, S.%
, Li, L.%
, Whang, J.%
, Denton, E.L.%
\BDBL {}others%
\end{APACrefauthors}%
\unskip\
\newblock
\APACrefYearMonthDay{2022}{}{}.
\newblock
{\BBOQ}\APACrefatitle {Photorealistic text-to-image diffusion models with deep language understanding} {Photorealistic text-to-image diffusion models with deep language understanding}.{\BBCQ}
\newblock
\APACjournalVolNumPages{Advances in neural information processing systems}{35}{}{36479--36494,}
\newblock

\newblock

\PrintBackRefs{\CurrentBib}

\bibitem [\protect \citeauthoryear {%
Schuhmann%
\ \protect \BOthers {.}}{%
Schuhmann%
\ \protect \BOthers {.}}{%
{\protect \APACyear {2022}}%
}]{%
Laion-5B}
\APACinsertmetastar {%
Laion-5B}%
\begin{APACrefauthors}%
Schuhmann, C.%
, Beaumont, R.%
, Vencu, R.%
, Gordon, C.W.%
, Wightman, R.%
, Cherti, M.%
\BDBL {}Jitsev, J.%
\end{APACrefauthors}%
\unskip\
\newblock
\APACrefYearMonthDay{2022}{}{}.
\newblock
{\BBOQ}\APACrefatitle {{LAION}-5B: An open large-scale dataset for training next generation image-text models} {{LAION}-5b: An open large-scale dataset for training next generation image-text models}.{\BBCQ}
\newblock
 \APACrefbtitle {Thirty-sixth Conference on Neural Information Processing Systems Datasets and Benchmarks Track.} {Thirty-sixth conference on neural information processing systems datasets and benchmarks track.}
\newblock
\begin{APACrefURL} {https://openreview.net/forum?id=M3Y74vmsMcY} \end{APACrefURL}
\PrintBackRefs{\CurrentBib}

\bibitem [\protect \citeauthoryear {%
R.~Shi%
\ \protect \BOthers {.}}{%
R.~Shi%
\ \protect \BOthers {.}}{%
{\protect \APACyear {2023}}%
}]{%
Zero123++}
\APACinsertmetastar {%
Zero123++}%
\begin{APACrefauthors}%
Shi, R.%
, Chen, H.%
, Zhang, Z.%
, Liu, M.%
, Xu, C.%
, Wei, X.%
\BDBL {}Su, H.%
\end{APACrefauthors}%
\unskip\
\newblock
\APACrefYearMonthDay{2023}{}{}.
\newblock
{\BBOQ}\APACrefatitle {Zero123++: a Single Image to Consistent Multi-view Diffusion Base Model} {Zero123++: a single image to consistent multi-view diffusion base model}.{\BBCQ}
\newblock
\APACjournalVolNumPages{CoRR}{abs/2310.15110}{}{,}
\newblock

\newblock

\PrintBackRefs{\CurrentBib}

\bibitem [\protect \citeauthoryear {%
Y.~Shi%
\ \protect \BOthers {.}}{%
Y.~Shi%
\ \protect \BOthers {.}}{%
{\protect \APACyear {2023}}%
}]{%
Mvdream}
\APACinsertmetastar {%
Mvdream}%
\begin{APACrefauthors}%
Shi, Y.%
, Wang, P.%
, Ye, J.%
, Long, M.%
, Li, K.%
\BCBL {} Yang, X.%
\end{APACrefauthors}%
\unskip\
\newblock
\APACrefYearMonthDay{2023}{}{}.
\newblock
{\BBOQ}\APACrefatitle {Mvdream: Multi-view diffusion for 3d generation} {Mvdream: Multi-view diffusion for 3d generation}.{\BBCQ}
\newblock
\APACjournalVolNumPages{arXiv preprint arXiv:2308.16512}{}{}{,}
\newblock

\newblock

\PrintBackRefs{\CurrentBib}

\bibitem [\protect \citeauthoryear {%
Stiennon%
\ \protect \BOthers {.}}{%
Stiennon%
\ \protect \BOthers {.}}{%
{\protect \APACyear {2020}}%
}]{%
Openai_RLHF}
\APACinsertmetastar {%
Openai_RLHF}%
\begin{APACrefauthors}%
Stiennon, N.%
, Ouyang, L.%
, Wu, J.%
, Ziegler, D.%
, Lowe, R.%
, Voss, C.%
\BDBL {}Christiano, P.%
\end{APACrefauthors}%
\unskip\
\newblock
\APACrefYearMonthDay{2020}{Sep}{}.
\newblock
{\BBOQ}\APACrefatitle {Learning to summarize from human feedback} {Learning to summarize from human feedback}.{\BBCQ}
\newblock
\APACjournalVolNumPages{arXiv: Computation and Language,arXiv: Computation and Language}{}{}{,}
\newblock

\newblock

\PrintBackRefs{\CurrentBib}

\bibitem [\protect \citeauthoryear {%
J.~Sun%
\ \protect \BOthers {.}}{%
J.~Sun%
\ \protect \BOthers {.}}{%
{\protect \APACyear {2023}}%
}]{%
DreamCraft3D}
\APACinsertmetastar {%
DreamCraft3D}%
\begin{APACrefauthors}%
Sun, J.%
, Zhang, B.%
, Shao, R.%
, Wang, L.%
, Liu, W.%
, Xie, Z.%
\BCBL {} Liu, Y.%
\end{APACrefauthors}%
\unskip\
\newblock
\APACrefYearMonthDay{2023}{}{}.
\newblock
{\BBOQ}\APACrefatitle {Dreamcraft3d: Hierarchical 3d generation with bootstrapped diffusion prior} {Dreamcraft3d: Hierarchical 3d generation with bootstrapped diffusion prior}.{\BBCQ}
\newblock
\APACjournalVolNumPages{arXiv preprint arXiv:2310.16818}{}{}{,}
\newblock

\newblock

\PrintBackRefs{\CurrentBib}

\bibitem [\protect \citeauthoryear {%
Z.~Sun%
\ \protect \BOthers {.}}{%
Z.~Sun%
\ \protect \BOthers {.}}{%
{\protect \APACyear {2024}}%
}]{%
bootstrap3d}
\APACinsertmetastar {%
bootstrap3d}%
\begin{APACrefauthors}%
Sun, Z.%
, Wu, T.%
, Zhang, P.%
, Zang, Y.%
, Dong, X.%
, Xiong, Y.%
\BDBL {}Wang, J.%
\end{APACrefauthors}%
\unskip\
\newblock
\APACrefYearMonthDay{2024}{}{}.
\newblock
{\BBOQ}\APACrefatitle {Bootstrap3D: Improving 3D Content Creation with Synthetic Data} {Bootstrap3d: Improving 3d content creation with synthetic data}.{\BBCQ}
\newblock
\APACjournalVolNumPages{arXiv preprint arXiv:2406.00093}{}{}{,}
\newblock

\newblock

\PrintBackRefs{\CurrentBib}

\bibitem [\protect \citeauthoryear {%
Tang%
\ \protect \BOthers {.}}{%
Tang%
\ \protect \BOthers {.}}{%
{\protect \APACyear {2024}}%
}]{%
LGM}
\APACinsertmetastar {%
LGM}%
\begin{APACrefauthors}%
Tang, J.%
, Chen, Z.%
, Chen, X.%
, Wang, T.%
, Zeng, G.%
\BCBL {} Liu, Z.%
\end{APACrefauthors}%
\unskip\
\newblock
\APACrefYearMonthDay{2024}{}{}.
\newblock
{\BBOQ}\APACrefatitle {LGM: Large Multi-View Gaussian Model for High-Resolution 3D Content Creation} {Lgm: Large multi-view gaussian model for high-resolution 3d content creation}.{\BBCQ}
\newblock
\APACjournalVolNumPages{arXiv preprint arXiv:2402.05054}{}{}{,}
\newblock

\newblock

\PrintBackRefs{\CurrentBib}

\bibitem [\protect \citeauthoryear {%
Tang%
, Ren%
, Zhou%
, Liu%
\BCBL {}\ \BBA {} Zeng%
}{%
Tang%
, Ren%
\BCBL {}\ \protect \BOthers {.}}{%
{\protect \APACyear {2023}}%
}]{%
dreamgaussian}
\APACinsertmetastar {%
dreamgaussian}%
\begin{APACrefauthors}%
Tang, J.%
, Ren, J.%
, Zhou, H.%
, Liu, Z.%
\BCBL {} Zeng, G.%
\end{APACrefauthors}%
\unskip\
\newblock
\APACrefYearMonthDay{2023}{}{}.
\newblock
{\BBOQ}\APACrefatitle {DreamGaussian: Generative Gaussian Splatting for Efficient 3D Content Creation} {Dreamgaussian: Generative gaussian splatting for efficient 3d content creation}.{\BBCQ}
\newblock
\APACjournalVolNumPages{arXiv preprint arXiv:2309.16653}{}{}{,}
\newblock

\newblock

\PrintBackRefs{\CurrentBib}

\bibitem [\protect \citeauthoryear {%
Tang%
, Wang%
\BCBL {}\ \protect \BOthers {.}}{%
Tang%
, Wang%
\BCBL {}\ \protect \BOthers {.}}{%
{\protect \APACyear {2023}}%
}]{%
tang2023_makeit3d}
\APACinsertmetastar {%
tang2023_makeit3d}%
\begin{APACrefauthors}%
Tang, J.%
, Wang, T.%
, Zhang, B.%
, Zhang, T.%
, Yi, R.%
, Ma, L.%
\BCBL {} Chen, D.%
\end{APACrefauthors}%
\unskip\
\newblock
\APACrefYearMonthDay{2023}{}{}.
\newblock
{\BBOQ}\APACrefatitle {Make-It-3D: High-Fidelity 3D Creation from A Single Image with Diffusion Prior} {Make-it-3d: High-fidelity 3d creation from a single image with diffusion prior}.{\BBCQ}
\newblock
 \APACrefbtitle {International Conference on Computer Vision {ICCV}.} {International conference on computer vision {ICCV}.}
\PrintBackRefs{\CurrentBib}

\bibitem [\protect \citeauthoryear {%
Taori%
\ \protect \BOthers {.}}{%
Taori%
\ \protect \BOthers {.}}{%
{\protect \APACyear {2023}}%
}]{%
alpaca}
\APACinsertmetastar {%
alpaca}%
\begin{APACrefauthors}%
Taori, R.%
, Gulrajani, I.%
, Zhang, T.%
, Dubois, Y.%
, Li, X.%
, Guestrin, C.%
\BDBL {}Hashimoto, T.B.%
\end{APACrefauthors}%
\unskip\
\newblock
\APACrefYearMonthDay{2023}{}{}.
\newblock
\APACrefbtitle {Stanford Alpaca: An Instruction-following LLaMA model.} {Stanford alpaca: An instruction-following llama model.}
\newblock
\APAChowpublished {\url{https://github.com/tatsu-lab/stanford_alpaca}}.
\newblock
\APACaddressPublisher{}{GitHub}.
\PrintBackRefs{\CurrentBib}

\bibitem [\protect \citeauthoryear {%
Tremblay%
\ \protect \BOthers {.}}{%
Tremblay%
\ \protect \BOthers {.}}{%
{\protect \APACyear {2022}}%
}]{%
RTMV}
\APACinsertmetastar {%
RTMV}%
\begin{APACrefauthors}%
Tremblay, J.%
, Meshry, M.%
, Evans, A.%
, Kautz, J.%
, Keller, A.%
, Khamis, S.%
\BDBL {}Birchfield, S.%
\end{APACrefauthors}%
\unskip\
\newblock
\APACrefYearMonthDay{2022}{}{}.
\newblock
{\BBOQ}\APACrefatitle {RTMV: A Ray-Traced Multi-View Synthetic Dataset for Novel View Synthesis} {Rtmv: A ray-traced multi-view synthetic dataset for novel view synthesis}.{\BBCQ}
\newblock
\APACjournalVolNumPages{IEEE/CVF European Conference on Computer Vision Workshop (Learn3DG ECCVW), 2022}{}{}{,}
\newblock

\newblock

\PrintBackRefs{\CurrentBib}

\bibitem [\protect \citeauthoryear {%
Voleti%
\ \protect \BOthers {.}}{%
Voleti%
\ \protect \BOthers {.}}{%
{\protect \APACyear {2024}}%
}]{%
SV3D}
\APACinsertmetastar {%
SV3D}%
\begin{APACrefauthors}%
Voleti, V.%
, Yao, C.%
, Boss, M.%
, Letts, A.%
, Pankratz, D.%
, Tochilkin, D.%
\BDBL {}Jampani, V.%
\end{APACrefauthors}%
\unskip\
\newblock
\APACrefYearMonthDay{2024}{}{}.
\newblock
{\BBOQ}\APACrefatitle {{SV3D:} Novel Multi-view Synthesis and 3D Generation from a Single Image using Latent Video Diffusion} {{SV3D:} novel multi-view synthesis and 3d generation from a single image using latent video diffusion}.{\BBCQ}
\newblock
\APACjournalVolNumPages{CoRR}{abs/2403.12008}{}{,}
\newblock

\newblock

\PrintBackRefs{\CurrentBib}

\bibitem [\protect \citeauthoryear {%
H.~Wang%
, Du%
, Li%
, Yeh%
\BCBL {}\ \BBA {} Shakhnarovich%
}{%
H.~Wang%
\ \protect \BOthers {.}}{%
{\protect \APACyear {2023}}%
}]{%
Score_Jacobian_Chaining}
\APACinsertmetastar {%
Score_Jacobian_Chaining}%
\begin{APACrefauthors}%
Wang, H.%
, Du, X.%
, Li, J.%
, Yeh, R.A.%
\BCBL {} Shakhnarovich, G.%
\end{APACrefauthors}%
\unskip\
\newblock
\APACrefYearMonthDay{2023}{}{}.
\newblock
{\BBOQ}\APACrefatitle {Score jacobian chaining: Lifting pretrained 2d diffusion models for 3d generation} {Score jacobian chaining: Lifting pretrained 2d diffusion models for 3d generation}.{\BBCQ}
\newblock
 \APACrefbtitle {Proceedings of the IEEE/CVF Conference on Computer Vision and Pattern Recognition} {Proceedings of the ieee/cvf conference on computer vision and pattern recognition}\ (\BPGS\ 12619--12629).
\PrintBackRefs{\CurrentBib}

\bibitem [\protect \citeauthoryear {%
T.~Wang%
\ \protect \BOthers {.}}{%
T.~Wang%
\ \protect \BOthers {.}}{%
{\protect \APACyear {2023}}%
}]{%
wang2022_rodin}
\APACinsertmetastar {%
wang2022_rodin}%
\begin{APACrefauthors}%
Wang, T.%
, Zhang, B.%
, Zhang, T.%
, Gu, S.%
, Bao, J.%
, Baltrusaitis, T.%
\BDBL {}Guo, B.%
\end{APACrefauthors}%
\unskip\
\newblock
\APACrefYearMonthDay{2023}{}{}.
\newblock
{\BBOQ}\APACrefatitle {Rodin: A Generative Model for Sculpting 3D Digital Avatars Using Diffusion} {Rodin: A generative model for sculpting 3d digital avatars using diffusion}.{\BBCQ}
\newblock
\APACjournalVolNumPages{CVPR}{}{}{,}
\newblock

\newblock

\PrintBackRefs{\CurrentBib}

\bibitem [\protect \citeauthoryear {%
Y.~Wang%
\ \protect \BOthers {.}}{%
Y.~Wang%
\ \protect \BOthers {.}}{%
{\protect \APACyear {2022}}%
}]{%
Self_Instruct}
\APACinsertmetastar {%
Self_Instruct}%
\begin{APACrefauthors}%
Wang, Y.%
, Kordi, Y.%
, Mishra, S.%
, Liu, A.%
, Smith, N.A.%
, Khashabi, D.%
\BCBL {} Hajishirzi, H.%
\end{APACrefauthors}%
\unskip\
\newblock
\APACrefYearMonthDay{2022}{}{}.
\newblock
{\BBOQ}\APACrefatitle {Self-instruct: Aligning language models with self-generated instructions} {Self-instruct: Aligning language models with self-generated instructions}.{\BBCQ}
\newblock
\APACjournalVolNumPages{arXiv preprint arXiv:2212.10560}{}{}{,}
\newblock

\newblock

\PrintBackRefs{\CurrentBib}

\bibitem [\protect \citeauthoryear {%
Z.~Wang%
, Lu%
\BCBL {}\ \protect \BOthers {.}}{%
Z.~Wang%
, Lu%
\BCBL {}\ \protect \BOthers {.}}{%
{\protect \APACyear {2024}}%
}]{%
ProlificDreamer}
\APACinsertmetastar {%
ProlificDreamer}%
\begin{APACrefauthors}%
Wang, Z.%
, Lu, C.%
, Wang, Y.%
, Bao, F.%
, Li, C.%
, Su, H.%
\BCBL {} Zhu, J.%
\end{APACrefauthors}%
\unskip\
\newblock
\APACrefYearMonthDay{2024}{}{}.
\newblock
{\BBOQ}\APACrefatitle {Prolificdreamer: High-fidelity and diverse text-to-3d generation with variational score distillation} {Prolificdreamer: High-fidelity and diverse text-to-3d generation with variational score distillation}.{\BBCQ}
\newblock
\APACjournalVolNumPages{Advances in Neural Information Processing Systems}{36}{}{,}
\newblock

\newblock

\PrintBackRefs{\CurrentBib}

\bibitem [\protect \citeauthoryear {%
Z.~Wang%
, Wang%
, Hancke%
, Liu%
\BCBL {}\ \BBA {} Lau%
}{%
Z.~Wang%
, Wang%
\BCBL {}\ \protect \BOthers {.}}{%
{\protect \APACyear {2024}}%
}]{%
wang2024themestation}
\APACinsertmetastar {%
wang2024themestation}%
\begin{APACrefauthors}%
Wang, Z.%
, Wang, T.%
, Hancke, G.%
, Liu, Z.%
\BCBL {} Lau, R.W.%
\end{APACrefauthors}%
\unskip\
\newblock
\APACrefYearMonthDay{2024}{}{}.
\newblock
{\BBOQ}\APACrefatitle {Themestation: Generating theme-aware 3d assets from few exemplars} {Themestation: Generating theme-aware 3d assets from few exemplars}.{\BBCQ}
\newblock
 \APACrefbtitle {ACM SIGGRAPH 2024 Conference Papers} {Acm siggraph 2024 conference papers}\ (\BPGS\ 1--12).
\PrintBackRefs{\CurrentBib}

\bibitem [\protect \citeauthoryear {%
T.~Wu%
, Wang%
\BCBL {}\ \protect \BOthers {.}}{%
T.~Wu%
, Wang%
\BCBL {}\ \protect \BOthers {.}}{%
{\protect \APACyear {2023}}%
}]{%
wu2022voxurf}
\APACinsertmetastar {%
wu2022voxurf}%
\begin{APACrefauthors}%
Wu, T.%
, Wang, J.%
, Pan, X.%
, Xu, X.%
, Theobalt, C.%
, Liu, Z.%
\BCBL {} Lin, D.%
\end{APACrefauthors}%
\unskip\
\newblock
\APACrefYearMonthDay{2023}{}{}.
\newblock
{\BBOQ}\APACrefatitle {Voxurf: Voxel-based Efficient and Accurate Neural Surface Reconstruction} {Voxurf: Voxel-based efficient and accurate neural surface reconstruction}.{\BBCQ}
\newblock
 \APACrefbtitle {International Conference on Learning Representations (ICLR).} {International conference on learning representations (iclr).}
\PrintBackRefs{\CurrentBib}

\bibitem [\protect \citeauthoryear {%
T.~Wu%
\ \protect \BOthers {.}}{%
T.~Wu%
\ \protect \BOthers {.}}{%
{\protect \APACyear {2024}}%
}]{%
Gpt4v_evaluation}
\APACinsertmetastar {%
Gpt4v_evaluation}%
\begin{APACrefauthors}%
Wu, T.%
, Yang, G.%
, Li, Z.%
, Zhang, K.%
, Liu, Z.%
, Guibas, L.%
\BDBL {}Wetzstein, G.%
\end{APACrefauthors}%
\unskip\
\newblock
\APACrefYearMonthDay{2024}{}{}.
\newblock
{\BBOQ}\APACrefatitle {Gpt-4v (ision) is a human-aligned evaluator for text-to-3d generation} {Gpt-4v (ision) is a human-aligned evaluator for text-to-3d generation}.{\BBCQ}
\newblock
 \APACrefbtitle {Proceedings of the IEEE/CVF Conference on Computer Vision and Pattern Recognition} {Proceedings of the ieee/cvf conference on computer vision and pattern recognition}\ (\BPGS\ 22227--22238).
\PrintBackRefs{\CurrentBib}

\bibitem [\protect \citeauthoryear {%
T.~Wu%
, Zhang%
\BCBL {}\ \protect \BOthers {.}}{%
T.~Wu%
, Zhang%
\BCBL {}\ \protect \BOthers {.}}{%
{\protect \APACyear {2023}}%
}]{%
omniobject3d}
\APACinsertmetastar {%
omniobject3d}%
\begin{APACrefauthors}%
Wu, T.%
, Zhang, J.%
, Fu, X.%
, Wang, Y.%
, Jiawei~Ren, L.P.%
, Wu, W.%
\BDBL {}Liu, Z.%
\end{APACrefauthors}%
\unskip\
\newblock
\APACrefYearMonthDay{2023}{}{}.
\newblock
{\BBOQ}\APACrefatitle {OmniObject3D: Large-Vocabulary 3D Object Dataset for Realistic Perception, Reconstruction and Generation} {Omniobject3d: Large-vocabulary 3d object dataset for realistic perception, reconstruction and generation}.{\BBCQ}
\newblock
 \APACrefbtitle {IEEE/CVF Conference on Computer Vision and Pattern Recognition (CVPR).} {Ieee/cvf conference on computer vision and pattern recognition (cvpr).}
\PrintBackRefs{\CurrentBib}

\bibitem [\protect \citeauthoryear {%
X.~Wu%
\ \protect \BOthers {.}}{%
X.~Wu%
\ \protect \BOthers {.}}{%
{\protect \APACyear {2023}}%
}]{%
HPSv2}
\APACinsertmetastar {%
HPSv2}%
\begin{APACrefauthors}%
Wu, X.%
, Hao, Y.%
, Sun, K.%
, Chen, Y.%
, Zhu, F.%
, Zhao, R.%
\BCBL {} Li, H.%
\end{APACrefauthors}%
\unskip\
\newblock
\APACrefYearMonthDay{2023}{}{}.
\newblock
{\BBOQ}\APACrefatitle {Human Preference Score v2: A Solid Benchmark for Evaluating Human Preferences of Text-to-Image Synthesis} {Human preference score v2: A solid benchmark for evaluating human preferences of text-to-image synthesis}.{\BBCQ}
\newblock
\APACjournalVolNumPages{arXiv preprint arXiv:2306.09341}{}{}{,}
\newblock

\newblock

\PrintBackRefs{\CurrentBib}

\bibitem [\protect \citeauthoryear {%
J.~Xu%
, Cheng%
\BCBL {}\ \protect \BOthers {.}}{%
J.~Xu%
, Cheng%
\BCBL {}\ \protect \BOthers {.}}{%
{\protect \APACyear {2024}}%
}]{%
InstantMesh}
\APACinsertmetastar {%
InstantMesh}%
\begin{APACrefauthors}%
Xu, J.%
, Cheng, W.%
, Gao, Y.%
, Wang, X.%
, Gao, S.%
\BCBL {} Shan, Y.%
\end{APACrefauthors}%
\unskip\
\newblock
\APACrefYearMonthDay{2024}{}{}.
\newblock
{\BBOQ}\APACrefatitle {InstantMesh: Efficient 3D Mesh Generation from a Single Image with Sparse-view Large Reconstruction Models} {Instantmesh: Efficient 3d mesh generation from a single image with sparse-view large reconstruction models}.{\BBCQ}
\newblock
\APACjournalVolNumPages{arXiv preprint arXiv:2404.07191}{}{}{,}
\newblock

\newblock

\PrintBackRefs{\CurrentBib}

\bibitem [\protect \citeauthoryear {%
J.~Xu%
, Liu%
\BCBL {}\ \protect \BOthers {.}}{%
J.~Xu%
, Liu%
\BCBL {}\ \protect \BOthers {.}}{%
{\protect \APACyear {2024}}%
}]{%
ImageReward}
\APACinsertmetastar {%
ImageReward}%
\begin{APACrefauthors}%
Xu, J.%
, Liu, X.%
, Wu, Y.%
, Tong, Y.%
, Li, Q.%
, Ding, M.%
\BDBL {}Dong, Y.%
\end{APACrefauthors}%
\unskip\
\newblock
\APACrefYearMonthDay{2024}{}{}.
\newblock
{\BBOQ}\APACrefatitle {Imagereward: Learning and evaluating human preferences for text-to-image generation} {Imagereward: Learning and evaluating human preferences for text-to-image generation}.{\BBCQ}
\newblock
\APACjournalVolNumPages{Advances in Neural Information Processing Systems}{36}{}{,}
\newblock

\newblock

\PrintBackRefs{\CurrentBib}

\bibitem [\protect \citeauthoryear {%
R.~Xu%
\ \protect \BOthers {.}}{%
R.~Xu%
\ \protect \BOthers {.}}{%
{\protect \APACyear {2023}}%
}]{%
pointllm}
\APACinsertmetastar {%
pointllm}%
\begin{APACrefauthors}%
Xu, R.%
, Wang, X.%
, Wang, T.%
, Chen, Y.%
, Pang, J.%
\BCBL {} Lin, D.%
\end{APACrefauthors}%
\unskip\
\newblock
\APACrefYearMonthDay{2023}{}{}.
\newblock
{\BBOQ}\APACrefatitle {PointLLM: Empowering Large Language Models to Understand Point Clouds} {Pointllm: Empowering large language models to understand point clouds}.{\BBCQ}
\newblock
\APACjournalVolNumPages{arXiv preprint arXiv:2308.16911}{}{}{,}
\newblock

\newblock

\PrintBackRefs{\CurrentBib}

\bibitem [\protect \citeauthoryear {%
Y.~Xu%
\ \protect \BOthers {.}}{%
Y.~Xu%
\ \protect \BOthers {.}}{%
{\protect \APACyear {2024}}%
}]{%
GRM}
\APACinsertmetastar {%
GRM}%
\begin{APACrefauthors}%
Xu, Y.%
, Shi, Z.%
, Yifan, W.%
, Peng, S.%
, Yang, C.%
, Shen, Y.%
\BCBL {} Gordon, W.%
\end{APACrefauthors}%
\unskip\
\newblock
\APACrefYearMonthDay{2024}{}{}.
\newblock
{\BBOQ}\APACrefatitle {GRM: Large Gaussian Reconstruction Model for Efficient 3D Reconstruction and Generation} {Grm: Large gaussian reconstruction model for efficient 3d reconstruction and generation}.{\BBCQ}
\newblock
\APACjournalVolNumPages{arxiv: 2403.14621}{}{}{,}
\newblock

\newblock

\PrintBackRefs{\CurrentBib}

\bibitem [\protect \citeauthoryear {%
Ye%
\ \protect \BOthers {.}}{%
Ye%
\ \protect \BOthers {.}}{%
{\protect \APACyear {2024}}%
}]{%
DreamReward}
\APACinsertmetastar {%
DreamReward}%
\begin{APACrefauthors}%
Ye, J.%
, Liu, F.%
, Li, Q.%
, Wang, Z.%
, Wang, Y.%
, Wang, X.%
\BDBL {}Zhu, J.%
\end{APACrefauthors}%
\unskip\
\newblock
\APACrefYearMonthDay{2024}{}{}.
\newblock
\APACrefbtitle {DreamReward: Text-to-3D Generation with Human Preference.} {Dreamreward: Text-to-3d generation with human preference.}
\PrintBackRefs{\CurrentBib}

\bibitem [\protect \citeauthoryear {%
Yin%
\ \protect \BOthers {.}}{%
Yin%
\ \protect \BOthers {.}}{%
{\protect \APACyear {2023}}%
}]{%
metric3d}
\APACinsertmetastar {%
metric3d}%
\begin{APACrefauthors}%
Yin, W.%
, Zhang, C.%
, Chen, H.%
, Cai, Z.%
, Yu, G.%
, Wang, K.%
\BDBL {}Shen, C.%
\end{APACrefauthors}%
\unskip\
\newblock
\APACrefYearMonthDay{2023}{}{}.
\newblock
{\BBOQ}\APACrefatitle {Metric3d: Towards zero-shot metric 3d prediction from a single image} {Metric3d: Towards zero-shot metric 3d prediction from a single image}.{\BBCQ}
\newblock
 \APACrefbtitle {Proceedings of the IEEE/CVF International Conference on Computer Vision} {Proceedings of the ieee/cvf international conference on computer vision}\ (\BPGS\ 9043--9053).
\PrintBackRefs{\CurrentBib}

\bibitem [\protect \citeauthoryear {%
C.~Zheng%
\ \BBA {} Vedaldi%
}{%
C.~Zheng%
\ \BBA {} Vedaldi%
}{%
{\protect \APACyear {2023}}%
}]{%
Free3D}
\APACinsertmetastar {%
Free3D}%
\begin{APACrefauthors}%
Zheng, C.%
\BCBT {}\ \BBA {} Vedaldi, A.%
\end{APACrefauthors}%
\unskip\
\newblock
\APACrefYearMonthDay{2023}{}{}.
\newblock
{\BBOQ}\APACrefatitle {Free3D: Consistent Novel View Synthesis without 3D Representation} {Free3d: Consistent novel view synthesis without 3d representation}.{\BBCQ}
\newblock
\APACjournalVolNumPages{arXiv}{}{}{,}
\newblock

\newblock

\PrintBackRefs{\CurrentBib}

\bibitem [\protect \citeauthoryear {%
L.~Zheng%
\ \protect \BOthers {.}}{%
L.~Zheng%
\ \protect \BOthers {.}}{%
{\protect \APACyear {2023}}%
}]{%
Chatbot-Arena}
\APACinsertmetastar {%
Chatbot-Arena}%
\begin{APACrefauthors}%
Zheng, L.%
, Chiang, W\BHBI L.%
, Sheng, Y.%
, Zhuang, S.%
, Wu, Z.%
, Zhuang, Y.%
\BDBL {}others%
\end{APACrefauthors}%
\unskip\
\newblock
\APACrefYearMonthDay{2023}{}{}.
\newblock
{\BBOQ}\APACrefatitle {Judging llm-as-a-judge with mt-bench and chatbot arena} {Judging llm-as-a-judge with mt-bench and chatbot arena}.{\BBCQ}
\newblock
\APACjournalVolNumPages{Advances in Neural Information Processing Systems}{36}{}{46595--46623,}
\newblock

\newblock

\PrintBackRefs{\CurrentBib}

\bibitem [\protect \citeauthoryear {%
Zhu%
\ \BBA {} Zhuang%
}{%
Zhu%
\ \BBA {} Zhuang%
}{%
{\protect \APACyear {2023}}%
}]{%
HIFA}
\APACinsertmetastar {%
HIFA}%
\begin{APACrefauthors}%
Zhu, J.%
\BCBT {}\ \BBA {} Zhuang, P.%
\end{APACrefauthors}%
\unskip\
\newblock
\APACrefYearMonthDay{2023}{}{}.
\newblock
\APACrefbtitle {HiFA: High-fidelity Text-to-3D Generation with Advanced Diffusion Guidance.} {Hifa: High-fidelity text-to-3d generation with advanced diffusion guidance.}
\PrintBackRefs{\CurrentBib}

\bibitem [\protect \citeauthoryear {%
Zou%
\ \protect \BOthers {.}}{%
Zou%
\ \protect \BOthers {.}}{%
{\protect \APACyear {2023}}%
}]{%
Triplane}
\APACinsertmetastar {%
Triplane}%
\begin{APACrefauthors}%
Zou, Z\BHBI X.%
, Yu, Z.%
, Guo, Y\BHBI C.%
, Li, Y.%
, Liang, D.%
, Cao, Y\BHBI P.%
\BCBL {} Zhang, S\BHBI H.%
\end{APACrefauthors}%
\unskip\
\newblock
\APACrefYearMonthDay{2023}{}{}.
\newblock
{\BBOQ}\APACrefatitle {Triplane Meets Gaussian Splatting: Fast and Generalizable Single-View 3D Reconstruction with Transformers} {Triplane meets gaussian splatting: Fast and generalizable single-view 3d reconstruction with transformers}.{\BBCQ}
\newblock
\APACjournalVolNumPages{arXiv preprint arXiv:2312.09147}{}{}{,}
\newblock

\newblock

\PrintBackRefs{\CurrentBib}

\end{thebibliography}

\end{document}